\documentclass[journal]{IEEEtran}
\usepackage{multirow}
\usepackage{amsmath,epsfig}
\usepackage{amsmath,graphicx}
\usepackage{amsfonts,amssymb}
\usepackage{bm}
\usepackage[noend]{algpseudocode}
\usepackage{algorithmicx,algorithm}
\usepackage{epstopdf}
\usepackage{amsmath}
\usepackage{cases}
\usepackage{color}
\usepackage{tabularx}
\usepackage[table,xcdraw]{xcolor}
\usepackage{pifont}
\usepackage{subfigure}
\usepackage{array}
\usepackage{hyperref}
\usepackage{ulem}

\ifCLASSINFOpdf
\else
\fi
\begin{document}
\title{Contrastive Learning Based Recursive Dynamic Multi-Scale Network for Image Deraining}

\author{ Zhiying~Jiang,~Risheng~Liu,~\IEEEmembership{Member,~IEEE,}~Shuzhou~Yang, Zengxi~Zhang~and~Xin~Fan,~\IEEEmembership{Senior~Member,~IEEE}% <-this % stops a space
        \thanks{This work is partially supported by the National Natural Science Foundation of China under Grant~(Nos. 61922019), and the Fundamental Research Funds for the Central Universities.}
\thanks{Zhiying Jiang, Shuzhou Yang and Zengxi Zhang are with the School of Software Technology, Dalian University of Technology, Dalian 116024, China. (e-mail: zyjiang0630@gmail.com; yszdyx@gmail.com; cyouzoukyuu@gmail.com).}% <-this % stops a space
% <-this % stops a space
\thanks{Risheng Liu is with DUT-RU International School of Information Science
	\& Engineering and the Key Laboratory for Ubiquitous Network and Service Software of Liaoning Province, Dalian University of Technology, Dalian 116024, China. He is also with the Peng Cheng Laboratory, Shenzhen 518066, China and the Pazhou Laboratory~(Huangpu), Guangzhou 510715, China. (Corresponding author, e-mail: rsliu@dlut.edu.cn).}
\thanks{Xin Fan is with DUT-RU International School of Information Science \& Engineering and the Key Laboratory for Ubiquitous Network and Service Software of Liaoning Province, Dalian University of Technology, Dalian 116024, China. He is also with Peng Cheng Laboratory, Shenzhen 518066, China. (e-mail: xin.fan@dlut.edu.cn).}
\thanks{Manuscript received April 19, 2005; revised August 26, 2015.}}


\markboth{Journal of \LaTeX\ Class Files,~Vol.~14, No.~8, August~2015}%
{Shell \MakeLowercase{\textit{et al.}}: Bare Demo of IEEEtran.cls for IEEE Journals}



\maketitle

\begin{abstract}
Rain streaks significantly decrease the visibility of captured images and are also a stumbling block that restricts the performance of subsequent computer vision applications. The existing deep learning-based image deraining methods employ manually crafted networks and learn a straightforward projection from rainy images to clear images. In pursuit of better deraining performance, they focus on elaborating a more complicated architecture rather than exploiting the intrinsic properties of the positive and negative information.  
In this paper, we propose a contrastive learning-based image deraining method that investigates the correlation between rainy and clear images and leverages a contrastive prior to optimize the mutual information of the rainy and restored counterparts.
Given the complex and varied real-world rain patterns, we develop a recursive mechanism. It involves multi-scale feature extraction and dynamic cross-level information recruitment modules. The former advances the portrayal of diverse rain patterns more precisely, while the latter can selectively compensate high-level features for shallow-level information. We term the proposed recursive dynamic multi-scale network with a contrastive prior, RDMC. 
Extensive experiments on synthetic benchmarks and real-world images demonstrate that the proposed RDMC delivers strong performance on the depiction of rain streaks and outperforms the state-of-the-art methods. Moreover, a practical evaluation of object detection and semantic segmentation shows the effectiveness of the proposed method.
 
\end{abstract}


% \fi 
\IEEEpeerreviewmaketitle

\section{Introduction}
\IEEEPARstart{C}{onsidering} the low visibility and adverse interference caused by rain, single image deraining has attracted extensive research attention in recent years. It has become a significant processing step for computer vision applications, e.g., object detection~\cite{liu2016ssd,liu2022target,liu2022twin}, recognition~\cite{Chen_2019_CVPR,he2016deep} and semantic segmentation~\cite{islam2020semantic,jiang2022target}.
To eliminate rain interference and restore the rain-free background, numerous deraining methods have been investigated~\cite{li2016rain,liu2020Knowledge,risheng2018aggre}. Most of these methods are aimed at video data where they benefit from adjacent frames to determine rain locations. In contrast, only spatial information can be utilized in a single image. Therefore, single image deraining is highly ill-posed and imperative for a variety of real-world tasks.

For single image deraining, existing methods can be roughly classified into conventional model priors based and deep learning based methods. Conventional model-based methods rely on image patches to predict rain streaks, such as frequency decomposition~\cite{kang2012automatic}, discriminative sparse coding~\cite{luo2015removing}, and Gaussian mixture models~\cite{li2016rain}. However, the spatial contextual information of the global image is ignored. In addition, the prior assumptions of these methods are designed for a specific situation and are not applicable to all scenarios. Therefore, the rain streaks in various directions and shapes cannot be removed thoroughly by conventional methods.

With the development of deep learning, researchers have drawn support from its powerful feature representation ability to estimate irregular rain and have achieved remarkable deraining performance~\cite{jiang2020multi,liu2020investigating,zamir2021multi}. Nevertheless, there are still several issues that need to be considered.
First, the comprehensive background details and various rain patterns are distributed in the high-frequency components simultaneously, so it is challenging for the network to separate undesirable rain from the high-frequency layer.
Second, the powerful feature representation of deep learning relies on a large amount of training data. However, the available quantity of supervised data can barely cover all types of rain appearances, which hinders the performance of these deep learning-based methods in dealing with real-world rain streaks.
Previous works elaborated deeper structures to enhance the translation of real-world images. Nevertheless, they focus only on the similarity between the restored images and the ground truth, where a one-way reduction of the difference is prone to overfitting.

In this paper, we develop an effective image deraining network to address the issues mentioned above.
First, we denote the observed rainy, nonrain and restored rain-free images as negative, positive and anchor, respectively. Drawing on the success of contrastive learning in computer vision tasks~\cite{chen2020simple}, we expect the generated image to be close to the positive sample and far from the negative sample.
In practice, we employ a Contrastive Prior~({CP}) with perceptual information, which not only improves the performance of rain removal but also strengthens the perception of a given scene.

\begin{figure}[tp]
	\centering
	\setlength{\tabcolsep}{1pt}
	\begin{tabular}{cccccccccccc}
		\includegraphics[width=0.46\textwidth,height=0.19\textwidth]{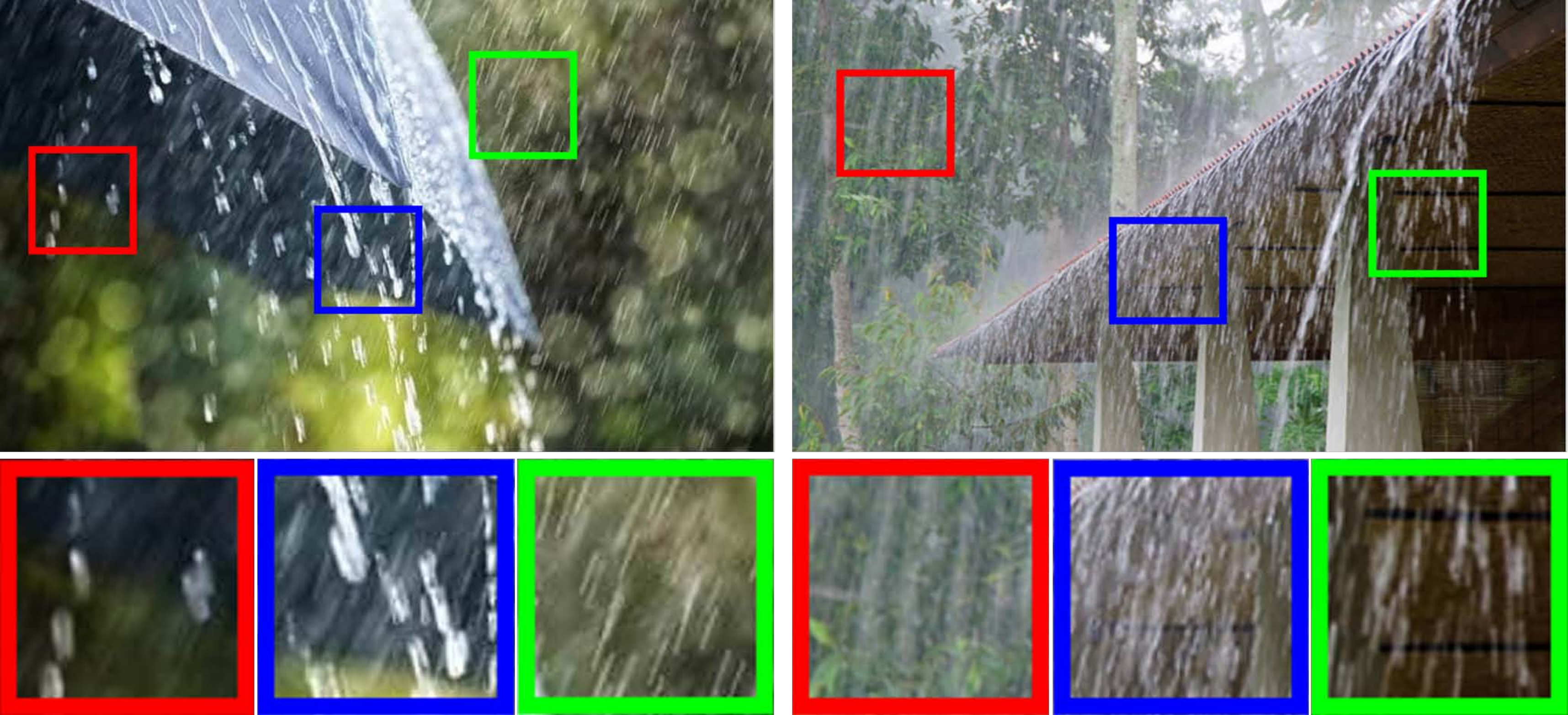}
	\end{tabular}
	\caption{Two samples of real-world rainy images. In both the left and right images, there are different types of rain streaks in terms of shape, density, illumination, and orientation. }
	\label{fig:two_samples}
\end{figure}

Taking into account the complexity and intertwining of rain interference shown in Fig.~\ref{fig:two_samples}, we propose a recursive mechanism that decomposes rain removal into multiple stages to iteratively eliminate the visible rain. Throughout the process, the different stages act in concert, with the later stage benefiting from the intermediate results of the previous stage. In addition, to improve the tolerance to rain streak scales, we further propose a contextualized Multi-scale Feature Extraction~(MFE) network that abstracts the context feature from a larger range of perceptions.
As we know, feature compensation enables information preservation in the deep layers. To establish a reasonable trade-off between efficiency and performance, we adopt a Neural Architecture Search~\cite{liu2019auto,liu2021learning,liu2021smoa} to determine the network with the best performance and to search for valuable connections between the encoder and decoder to alleviate the information loss.

To summarize, our contributions are as follows:

\begin{itemize}
	
%	\item \textcolor{red}{We propose a contrastive learning based recursive dynamic multi-scale network termed as RDMC for single image deraining. To our best knowledge, this is the first paper to introduce contrastive learning into image deraining. We consider the opposite relationship between the restored image with the rainy and the rain-free image. 
%%		By employing the contrastive prior, 
%	The proposed network is more effective in restoring the images with interlaced rain streaks and improving the perceptual comprehension of scene content.}
	\item This work introduces contrastive learning to image deraining and consider the opposite relationship between the restored image and the rainy and rain-free images to improve the perceptual comprehension of the scene, making it more effective for complicated rainy scenarios.
%	 and improving the perceptual comprehension of scene.

%	\item \textcolor{red}{Unlike the previous deep learning methods that focus on elaborating network architecture, we adopt the recursive mechanism to perform the removal stage by stage and treat the interlaced rain dependently. In this manner, we overcome the limitation of the receptive field and ensure the robustness for the arbitrary pattern that emerged in real-world scenarios.}
	\item We adopt a recursive mechanism to perform the removal stage by stage. Since the latter stage benefits from the latent results, the interlaced rain can be processed independently and progressively, advancing the generalization of heavy rain occasions.
	
	\item We propose the Dynamic Cross-level Recruitment~(DCR) to perform feature compensation across different levels, resolving the information diminishing that occurs in the deep layers. In addition, optimal connection searching enables a reasonable trade-off between effectiveness and efficiency.

	\item Extensive experiments demonstrate that the proposed RDMC achieves state-of-the-art performance on the various synthetic and real-world datasets. We further verify the effectiveness on the object detection and semantic segmentation tasks for a comprehensive evaluation.

\end{itemize}

\section{Related work}
\normalem
In this section, we will give a brief review of single image deraining and introduce supplementary works related to the proposed method, including contrastive learning and neural architecture search.
\subsection{Single Image Deraining}
The objective of single image deraining is to eliminate the interference of visible rain to reconstruct the corresponding high-quality images. Recently, a dozen single image deraining methods have been proposed.
Most of them utilized model priors to simulate the distribution of rain. For example, Kang~\emph{et al.}~\cite{kang2012automatic} employed morphological component analysis on the frequency domain. 
Built upon the nonlinear generative model, Luo~\emph{et al.}~\cite{luo2015removing} proposed mutual property guided discriminative sparse coding, and Li~\emph{et al.}~\cite{li2016rain} adopted Gaussian mixture models for rain extraction. In addition, Zhu~\emph{et al.}~\cite{zhu2017unpaired} developed a novel angular deviation prior to locating the rain-corrupted region within the direction information.
Although these conventional model-based methods manipulate the handcrafted priors for rain streaks, they tend to retain much of the visible rain and suffer from information loss in the background.

With the rapid development of deep learning, researchers have begun to turn to the development of data-driven algorithms for image detaining~\cite{fu2016weighted,yang2016joint,fu2017removing,fu2021rain}. 
To remove the veiling effect caused by rain accumulation, Li~\emph{et al.}~\cite{li2017single} processed rain at different scales individually. Zhang~\emph{et al.}~\cite{zhang2018density} constructed a density-aware deraining network. Hu~\emph{et al.}~\cite{hu2019depth} first analysed the visual effect at different depths and formulated rain corruption with the visible rain and haze simultaneously. To exploit the effect of the paired operation, a collaborative representation across different scales is developed in~\cite{jiang2020multi}. Liu~\emph{et al.}~\cite{liu2019dual} introduced a dual residual connection to the general framework. Then, Ren~\emph{et al.}~\cite{ren2020Progressive} developed a progressive paradigm. Wang~\emph{et al.}~\cite{wang2020spatial} presented a semiautomatic deraining method that integrated the temporal priors and supervision information to constrain the recovered results from global to local. 

Due to the gap between synthetic training data and real-world testing images, the performance of deep learning-based methods on real-world images is not very credible. Existing works~\cite{wei2019semi,yasarla2020syn2real} exploited transfer learning to adapt pretrained models to the real world. We need to note that visible rain under natural conditions is complex and diverse. There may be multiple forms of rain shading in a specific scene, which inspires us to treat the different patterns independently and reveal them progressively.
\subsection{Contrastive Learning}
Based on a large amount of manually synthetic data, supervised learning has achieved promising advances in computer vision. But in practice, hand labelling is challenging and time-consuming, and a model obtained from supervised learning cannot be transferred to other tasks. Therefore, self-supervised representation learning, which adapts itself as supervised information to learn the feature representations, has become an emerging research focus. More recently, contrastive learning has been the principal framework within the self-supervised approach. The critical idea of contrastive learning is to draw the anchor closer to the positive samples and to move away from the negative ones in specific representation spaces. To date, there have been several attempts to introduce contrastive learning to high-level vision tasks.
For example, Dai~\emph{et al.}~\cite{dai2017contrastive} proposed a novel image captioning method, which leverages the generic framework of contrastive learning.  It ensures an improvement against the corresponding baseline and gets superior performance in generalizing to other models. After that, a linear classifier trained by~\cite{chen2020simple} also improved remarkably over the existing methods. 

In low-level vision, Park~\emph{et al.}~\cite{park2020contrastive} first utilized contrastive learning in the unpaired image translation task.
Wu~\emph{et al.}~\cite{wu2021contrastive} proposed a compact dehazing network that demonstrated the improvement of contrastive learning performed on the autoencoder-like framework.
For single image deraining, there is still a gap in the attempt to introduce contrast learning. We note that the existing deraining methods employ the learnable parameters and models to simulate the mapping from the observed rainy image to the rain-free image or the rain streaks, ignoring the potential effect of the rainy image. We assume that the rainy image is able to shift the restored image away from a rainy appearance, while the rain-free image forces them closer to a clearer background. Therefore, these opposite principles  promote the deraining performance, especially in complicated real-world scenarios. 
\subsection{Neural Architecture Search}
The design of handcrafted networks with empirical observations is time-consuming and error prone. To this end, Neural Architecture Search~(NAS), which enables the procedure of automating architecture engineering, is in high demand and has been applied to a series of tasks~\cite{liu2019auto,liu2022learn}. Existing NAS algorithms can be categorized into three dimensions, search space, search strategy, and performance estimation strategy. 

The search space is the primary factor affecting network performance which defines a set of fundamental operations of the network. Many elaborate elements have been introduced into the search space, such as skip connections and multiple branches. In addition, repeatable cells~\cite{zoph2018learning} and dense connections~\cite{cai2018path} are also employed as candidate operations.
As mentioned above, existing search spaces are discrete and cannot be differentiated. To relax them into a continuous space and use gradient descent to optimize them, differentiable methods~(i.e., DARTS)~\cite{liu2018darts} have been proposed. Unlike conventional methods, which search for a special structure in a continuous space, DARTS searches for a complete high-performance framework with complex graph topology in a rich search space.

To reduce the network training consumption and improve the efficiency of performance estimation, a series of reduction strategies have been developed, e.g., reducing training epochs~\cite{zela2018towards, rawal2018nodes}, employing a subset of datasets~\cite{chrabaszcz2017downsampled}, and pruning supergraphs~\cite{xie2018snas}. These strategies make it possible for us to build high-performance deraining networks via an automatic search mechanism, releasing the requirement for a handcrafted and empirical design.

\begin{figure*}[tp]
	\centering
	\setlength{\tabcolsep}{1pt}
	\begin{tabular}{cccccccccccc}
		\includegraphics[width=\textwidth]{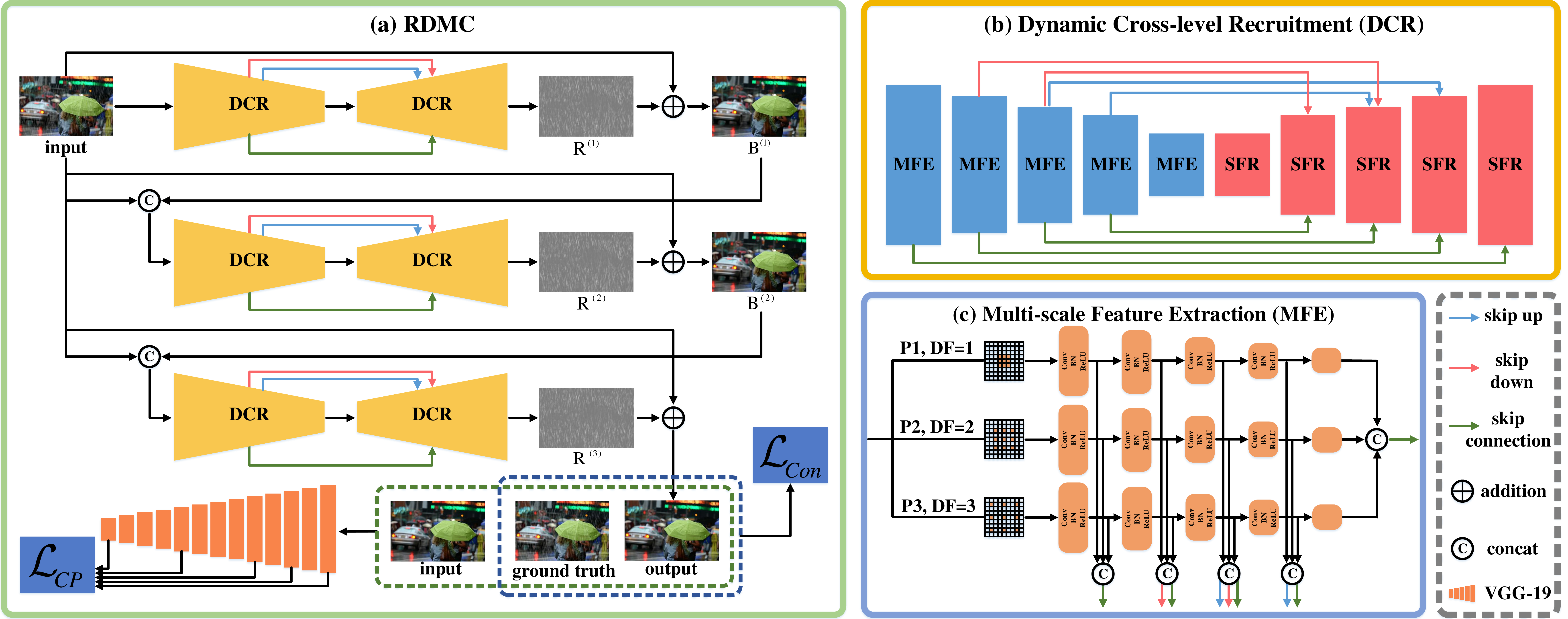}
		
	\end{tabular}
	\caption{(a)~Workflow of the proposed RDMC. We can see that the proposed method consists of three recursive stages, to progressively perform rain removal. (b)~Illustration of the Dynamic Cross-level Recruitment~(DCR) module, where the optimal architecture has been established with NAS. (c)~Multi-scale Feature Extraction~(MFE) used in DCR. }
	\label{fig:workflow}
\end{figure*}

\section{The proposed method}
We suppose that the observed rainy image~$\bold{I}$~can be decomposed into a linear combination of rain map~$\bold{R}$~and rain-free background~$\bold{B}$, expressed as:
\begin{equation}
	\bold{I}=\bold{B}+\bold{R}.
\end{equation}

Single image deraining refers to eliminating the interference of visible rain~$\bold{R}$ and reconstructing the rain-free scenery~$\bold{B}$. Fig.~\ref{fig:workflow} shows the framework of the proposed RDMC, which executes the removal process progressively. In this manner, the most evident and coherent rain streaks can be removed first, and the remaining rain streaks that are interfused in the complicated background are gradually discerned. Rain streaks under realistic conditions exhibit a variety of patterns that vary in size, intensity, orientation and luminance. To break the restriction of the specific rain patterns, we employ a Multi-scale Feature Extraction~(MFE), which is able to accommodate rain with diverse patterns and remove them in one unified framework. After that, we elaborate a Dynamic Cross-level Recruitment~(DCR) module to compensate deep layers for the shallow feature information. It instigates a neural architecture search to automatically construct the cross-level connections, which not only reduces the calculations and storage but also ensures the performance of detail preservation. To explore the similarities and differences between the clear image and the corresponding degraded counterpart, we introduce contrastive learning, where the reconstructed result is forced closer to the ground truth and away from the degraded rainy image. In practice, the contrastive relationship is employed as the prior constraint in optimizing the whole framework. To this end, the proposed method tends to be more robust to real-world cases and eliminates the reliance on the quality of the training data.

In the following, we illustrate the design of MFE for characterizing rain streaks with diverse dimensions in Section A and DCR for exploiting adaptive information recruitment in Section B. Finally, Section C proposes a contrastive learning-based prior constraint in regard to both the degraded images and their counterparts.

\subsection{Multi-scale Feature Extraction}
Rain patterns present differently in diverse dimensions,~i.e., illumination, intensity, and orientation, making it difficult for the model to comprehensively characterize rain corruption. As analysed, contextual information would advance rain identification to benefit variable rain removal more effectively. Since single-scale convolution cannot pick up the contextual information as expected~\cite{liu2022attention,liu2021learning}, we introduce a multi-scale module via dilated convolution to enlarge the receptive fields, and we aggregate the context feature in multiple scales to learn a rain portrait without an emerging resolution decrease.

Specifically, for the first stage, we treat the observed rainy image as input directly, while for the second and third stages, the result of the previous stage is concatenated with the observation as the input of the current stage. As shown in Fig.~\ref{fig:workflow}~(c), we feed the input into three parallel paths, where each path consists of five convolutions configured with a certain Dilated Factor~(DF). In our method, the DFs of paths P1, P2 and P3 are set as~$1, 2$ and $3$, respectively. Different paths possess different receptive fields, which increase along with the DF. The convolution kernel sizes of all three paths are~$3\times3$. Therefore, the resolution of the output features after each convolution layer is of the same size. 

To formalize this process, we denote~$\bold{B}^{(i)}$ as the reconstructed rain-free result from the~$i$-th stage, and $\bold{I}$ denotes the observation. The reconstructed latent result is progressively updated as follows:
\begin{equation}
	\bold{f}_{\rm in,p}^{(i+1)}={\rm concat}(\bold{B}^{(i)},\bold{I}),\label{fin}
\end{equation} 
\begin{equation}
	\bold{f}_{\rm mid,p}^{(i+1)}=\sigma({\rm BN}(\bold{W}^{(i+1)}_{\rm in,p}*{\bold{f}^{(i+1)}_{\rm in,p}}+{\bold b^{(i+1)}_{\rm in,p}})),\label{fmid}
\end{equation}
\begin{equation}
	\bold{f}_{\rm out,p}^{(i+1)}=\sigma({\rm BN}(\bold{W}^{(i+1)}_{\rm mid,p}*{\rm max}(0,{\bold{f}^{(i+1)}_{\rm mid,p}})+{\bold b^{(i+1)}_{\rm mid,p}})),\label{fout}
\end{equation}
where~$\rm p$ denotes the dilation path, and~$\bold{f}_{\rm in}^{(i+1)}$ in Eq.~\eqref{fin} produces the actual input of the~$(i+1)$-th stage. Note that for the first stage,~$\bold{B}^{(0)}=\bold{I}$. In Eq.~\eqref{fmid} and Eq.~\eqref{fout}, $*$ denotes the convolution operator,~$\bold{W}^{(i+1)}_{\rm in,p},\bold{W}^{(i+1)}_{\rm mid,p}$ and~${\bold b^{(i+1)}_{\rm in,p},\bold b^{(i+1)}_{\rm mid,p}}$ denote the filter weight and bias of the convolution layer, respectively, BN denotes batch normalization, and~$\sigma$ presents nonlinear activation. After the first two layers, we make the variable ${\bold{f}_{\rm mid,p}^{(i+1)}}={\bold{f}_{\rm out,p}^{(i+1)}}$ in Eq.~\eqref{fout} to perform the forward feature update.

As shown in the bottom region of Fig.~\ref{fig:workflow}~(c), after each convolution layer, the results from the three paths are concatenated together as expressed in Eq.~\eqref{fcat}.
\begin{equation}
	{\bold F}_{l}^{(i+1)}={\rm concat}({\bold f}_{ l,1}^{(i+1)},{\bold f}_{ l,2}^{(i+1)},{\bold f}_{ l,3}^{(i+1)}),\label{fcat}
\end{equation}
where~${\bold F}_{l}^{(i+1)}$ indicates the concatenated features of the~$l$-th layer at the~$(i+1)$-th stage.~${\bold f}_{ l,1}^{(i+1)}, {\bold f}_{ l,2}^{(i+1)}$ and~${\bold f}_{ l,3}^{(i+1)}$ present the features obtained from the three paths of the~$l$-th layer. Then,~${\bold F}_{l}^{(i+1)}$ is transported to the corresponding decoder layers as well as the cross-level target layers selected by the automatic searching mechanism proposed in Section B.

\subsection{Dynamic Cross-level Recruitment Module }
\begin{figure}[tp]
	\centering
	\setlength{\tabcolsep}{1pt}
	\begin{tabular}{cccccccccccc}
		\includegraphics[width=0.45\textwidth]{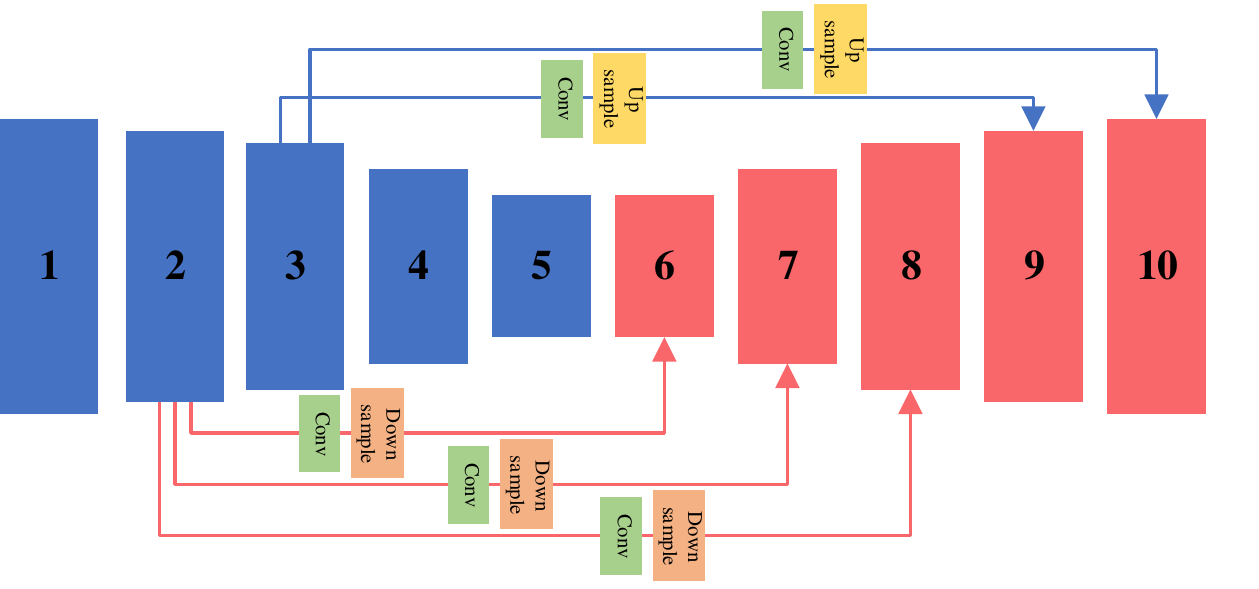}		
	\end{tabular}
	\caption{ Illustration of the search space for a cross-level skip connection. The red lines show the skip-down space of the second layer, while the blue lines present the skip-up space of the third layer. }
	\label{fig:skip}
\end{figure}

With the deepening of the network architecture, the feature information of the shallow layers is gradually lost. Existing works~\cite{ronneberger2015u,Dong2020Multi} attempt to merge the latent features from the same level of both the encoder and decoder, which motivates the network to aggregate the features from different levels to achieve information compensation and improves the performance to a certain degree. However, the computing efficiency and storage limitations obstruct the implementation of such merging assumption. 
We know that for a network with~$2n$ layers, both the encoder and decoder have~$n$ layers, and any layer in the encoder can be connected to~$n-1$ layers in the decoder in a cross-layer manner. That is, there are~$n\times(n-1)$ possible connections for the network with~$2n$ layers. Nevertheless, not all of these potential connections are significant. Inspired by Neural Architecture Search~(NAS), which is able to establish the best-performance structure with the given search space, we endow a weight parameter,~$\alpha$, to each potential connection and adopt a differentiable-based search algorithm~\cite{liu2018darts} to determine the pivotal skip connection. Specifically, for the search space, we divide the cross-level connections into two categories, i.e., skip-up and skip-down, where skip-up means a link pointing to a resolution higher than itself and skip-down points to a lower resolution layer. As shown in Fig.~\ref{fig:skip}, for the second layer, there are three potential skip-down connections with layers 6, 7 and 8, while for the third layer, layers 9 and 10 are two options to form the skip-up connection with it. At this point, we determine all the optional connections across the different level layers in the encoder. 

The adopted differential-based search algorithm establishes a continuous searching space, and the conventional gradient descent is employed for optimization~\cite{liu2021bilevel}. We choose the two connections with the highest weight coefficient,~$\alpha$, in skip-up and skip-down separately to obtain the optimal architecture. Ultimately, the skip-up connection patterns from layer 3 to 9, layer 4 to 8 and the skip-down connection patterns from layer 2 to 8, and layer 3 to 7 are chosen to establish the best-performance network. Note that before skip-up and skip-down, we employ an additional convolution and upsample or downsample operation to adjust the feature size and channel number of the encoder to fit the target in the decoder. In addition, the encoder in our network employs the multi-scale feature extraction proposed in Section A, while the decoder is a conventional network with a Single-scale Feature Reconstruction~(SFR).

\begin{figure*}[h]
	\centering
	\begin{minipage}{0.35\textwidth}
		
		\subfigure{
			\begin{minipage}{1\textwidth}
				\includegraphics[width=1\textwidth,height=0.240\textheight]{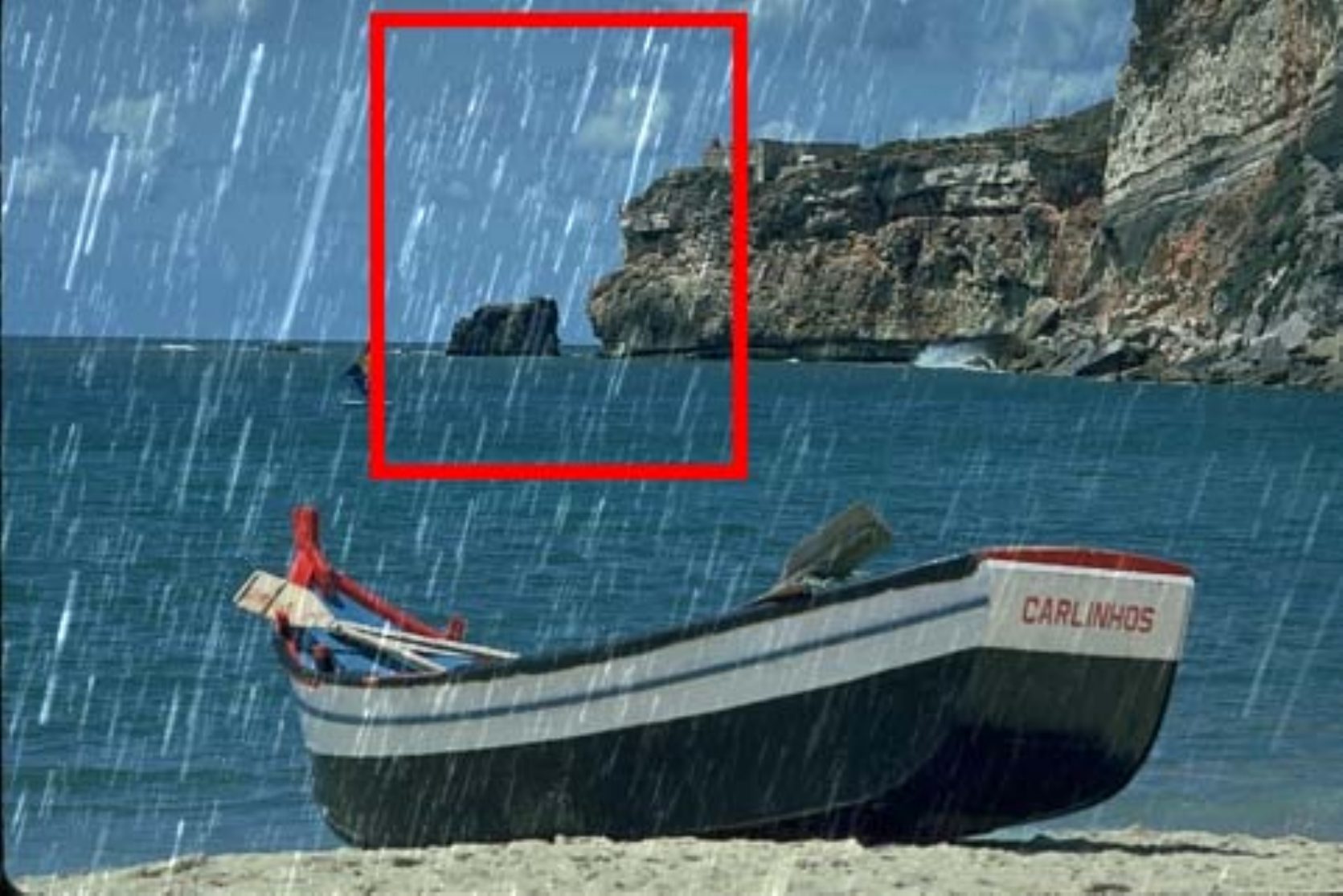}
				\centering  \footnotesize 26.22 / 0.78\\Input\\
			\end{minipage}
		}
	\end{minipage}
	\begin{minipage}{0.1\textwidth}
		\subfigure{
			\begin{minipage}{1\textwidth}
				\includegraphics[width=1\textwidth,height=0.102\textheight]{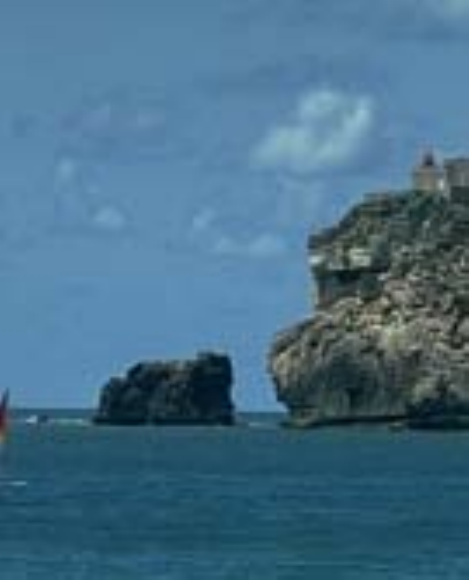}
				\centering \footnotesize- / -\\ Ground truth 	\\
			\end{minipage}
		}
		\subfigure{
			\begin{minipage}{1\textwidth}
				\includegraphics[width=1\textwidth,height=0.101\textheight]{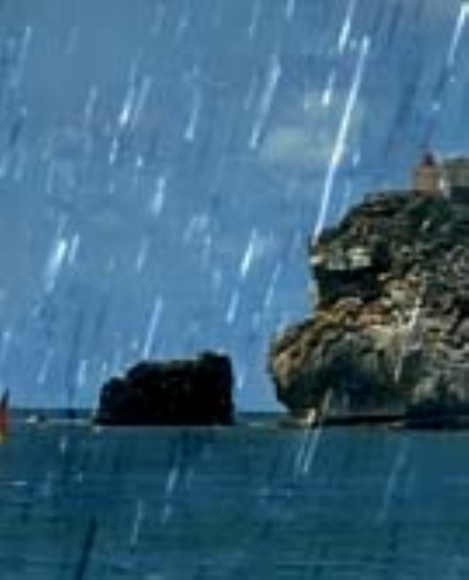}
				\centering \footnotesize 24.03 / 0.87\\SIRR\\
			\end{minipage}
		}
	\end{minipage}
	\begin{minipage}{0.1\textwidth}
		\subfigure{
			\begin{minipage}{1\textwidth}
				\includegraphics[width=1\textwidth,height=0.102\textheight]{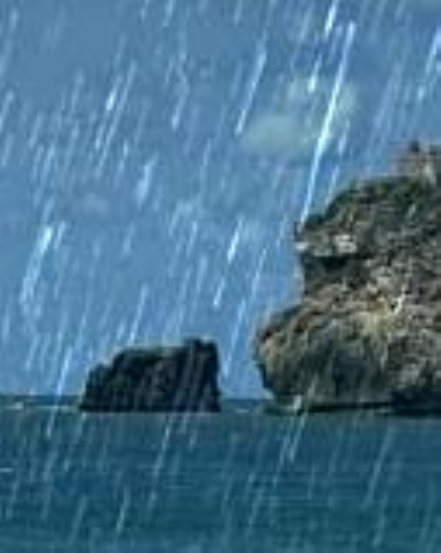}
				\centering \footnotesize 28.05 / 0.79\\DSC\\
			\end{minipage}
		}
		\subfigure{
			\begin{minipage}{1\textwidth}
				\includegraphics[width=1\textwidth,height=0.101\textheight]{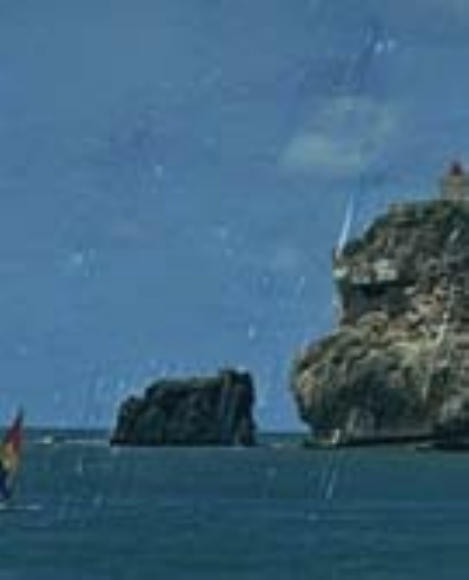}
				\centering \footnotesize 26.40 / 0.89\\ \vspace{-0.2em}Syn2Real\\
			\end{minipage}
		}
	\end{minipage}
	\begin{minipage}{0.1\textwidth}
		\subfigure{
			\begin{minipage}{1\textwidth}
				\includegraphics[width=1\textwidth,height=0.102\textheight]{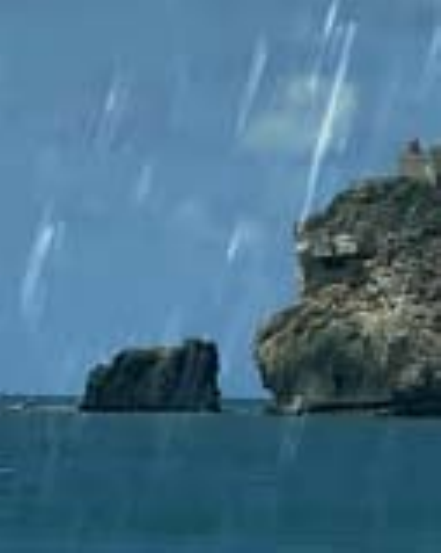}
				\centering \footnotesize29.69 / 0.85\\ LP\\
			\end{minipage}
		}
		\subfigure{
			\begin{minipage}{1\textwidth}
				\includegraphics[width=1\textwidth,height=0.101\textheight]{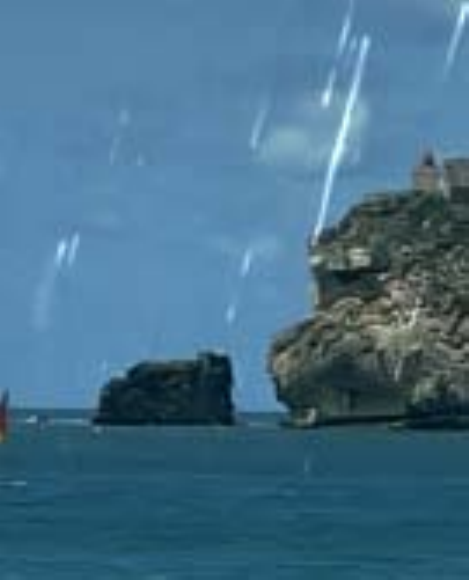}
				\centering \footnotesize 32.35 / 0.96\\MSPFN\\
			\end{minipage}
		}
	\end{minipage}
	\begin{minipage}{0.1\textwidth}
		\subfigure{
			\begin{minipage}{1\textwidth}
				\includegraphics[width=1\textwidth,height=0.102\textheight]{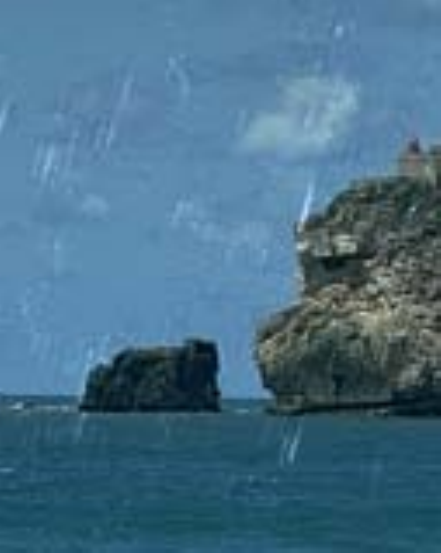}
				\centering \footnotesize 36.36 / 0.95\\ JORDER \\
			\end{minipage}
		}
		\subfigure{
			\begin{minipage}{1\textwidth}
				\includegraphics[width=1\textwidth,height=0.101\textheight]{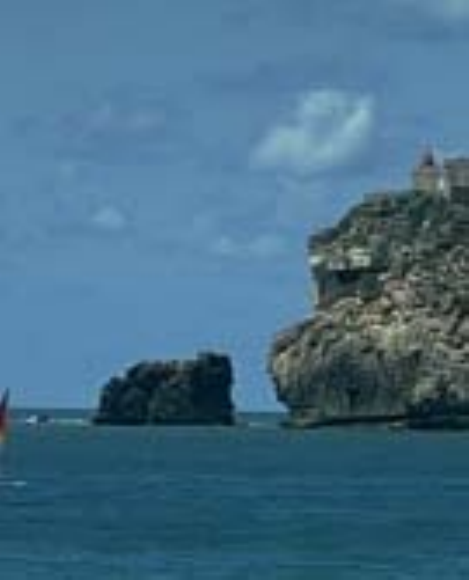}
				\centering \footnotesize 37.95 / 0.97\\DualGCN\\
			\end{minipage}
		}
	\end{minipage}
	\begin{minipage}{0.1\textwidth}
		\subfigure{
			\begin{minipage}{1\textwidth}
				\includegraphics[width=1\textwidth,height=0.102\textheight]{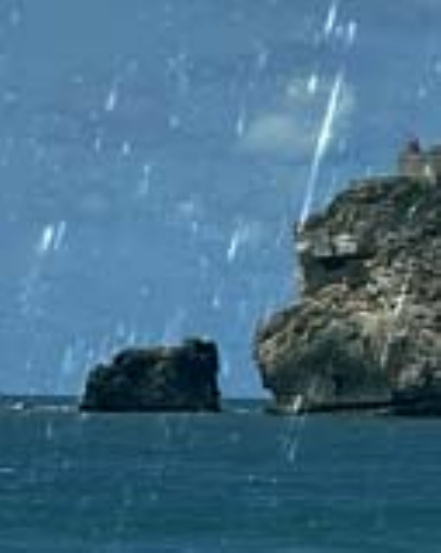}
				\centering \footnotesize30.80 / 0.95\\ DDN \\
			\end{minipage}
		}
		\subfigure{
			\begin{minipage}{1\textwidth}
				\includegraphics[width=1\textwidth,height=0.101\textheight]{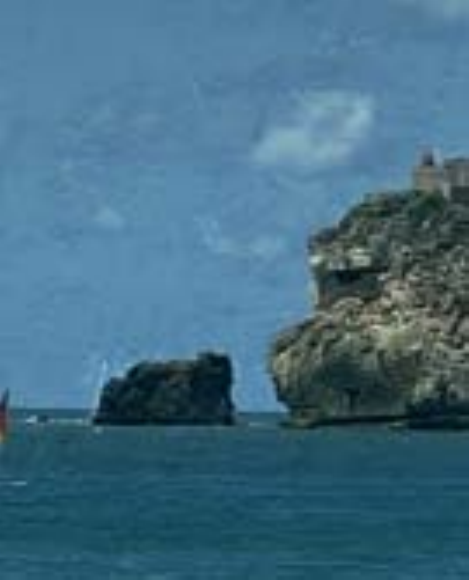}
				\centering \footnotesize 36.97 / 0.98\\ MPRNet\\
			\end{minipage}
		}
	\end{minipage}
	\begin{minipage}{0.1\textwidth}
		\subfigure{
			\begin{minipage}{1\textwidth}
				\includegraphics[width=1\textwidth,height=0.102\textheight]{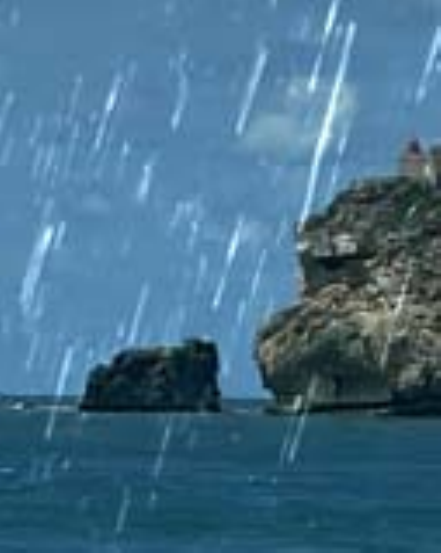}
				\centering \footnotesize 28.37 / 0.92\\DualRes \\
			\end{minipage}
		}
		\subfigure{
			\begin{minipage}{1\textwidth}
				\includegraphics[width=1\textwidth,height=0.101\textheight]{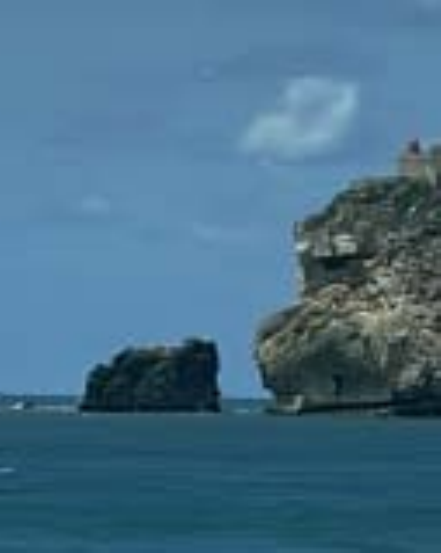}
				\centering \footnotesize 38.71 / 0.98\\Ours\\
			\end{minipage}
		}
	\end{minipage}
	\caption{Visual comparisons on the Rain12 dataset. The zoomed-in regions in the results show that our method is superior in removing visible rain streaks without introducing additional noise interference. The corresponding metric values PSNR / SSIM are reported under the images. }
	\label{fig:syn12com}
\end{figure*}
\begin{figure*}[h]
	\centering
	\begin{minipage}{0.35\textwidth}
		
		\subfigure{
			\begin{minipage}{1\textwidth}
				\includegraphics[width=1\textwidth,height=0.240\textheight]{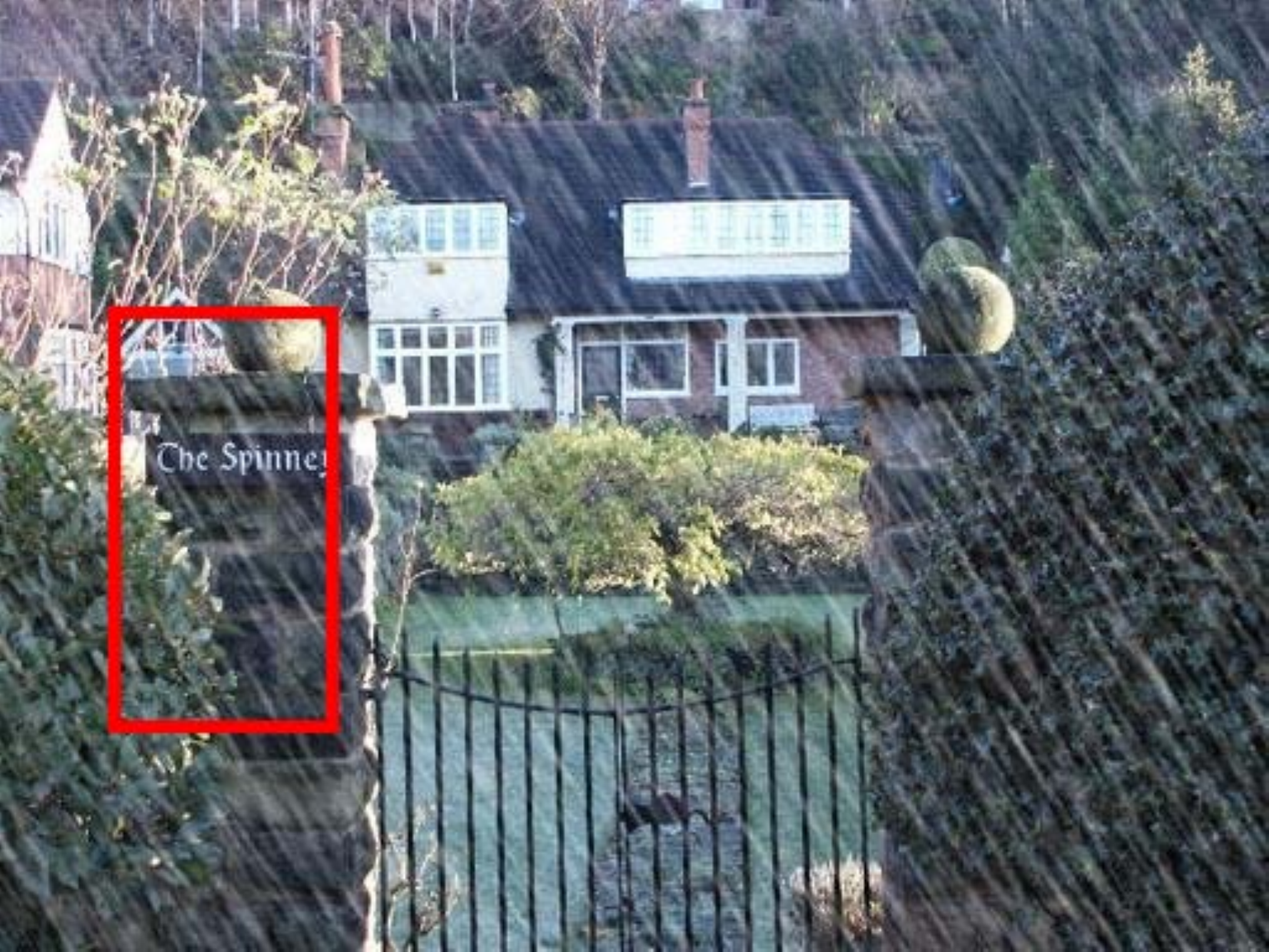}
				\centering  \footnotesize 19.65 / 0.77\\Input\\
			\end{minipage}
		}
	\end{minipage}
	\begin{minipage}{0.1\textwidth}
		\subfigure{
			\begin{minipage}{1\textwidth}
				\includegraphics[width=1\textwidth,height=0.102\textheight]{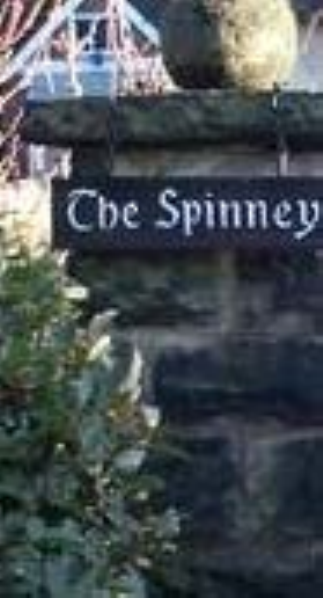}
				\centering \footnotesize- / -\\ Ground truth 	\\
			\end{minipage}
		}
		\subfigure{
			\begin{minipage}{1\textwidth}
				\includegraphics[width=1\textwidth,height=0.101\textheight]{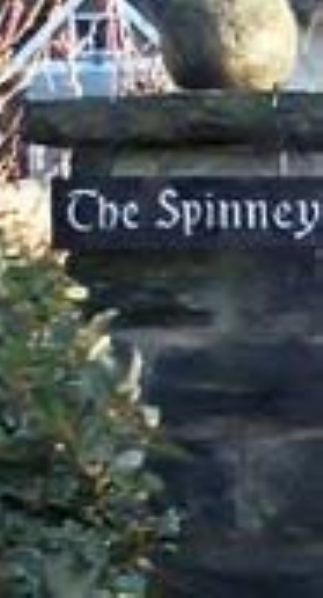}
				\centering \footnotesize 25.76 / 0.89\\SIRR\\
			\end{minipage}
		}
	\end{minipage}
	\begin{minipage}{0.1\textwidth}
		\subfigure{
			\begin{minipage}{1\textwidth}
				\includegraphics[width=1\textwidth,height=0.102\textheight]{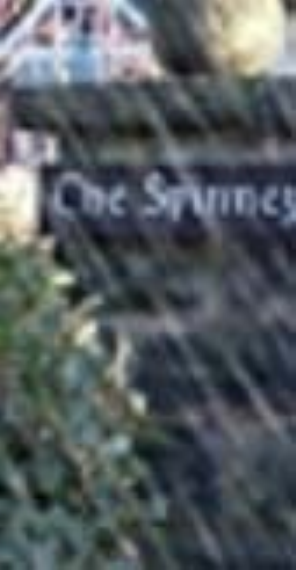}
				\centering \footnotesize 21.51 / 0.77\\DSC\\
			\end{minipage}
		}
		\subfigure{
			\begin{minipage}{1\textwidth}
				\includegraphics[width=1\textwidth,height=0.101\textheight]{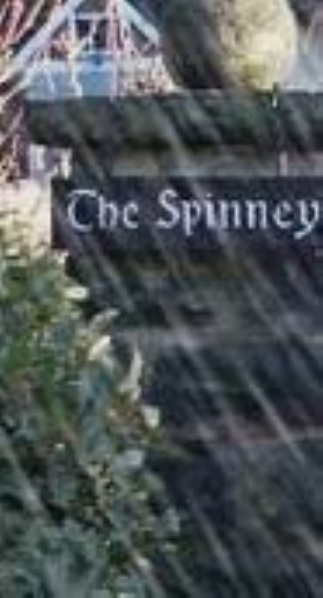}
				\centering \footnotesize 22.23 / 0.80\\ \vspace{-0.2em}Syn2Real\\
			\end{minipage}
		}
	\end{minipage}
	\begin{minipage}{0.1\textwidth}
		\subfigure{
			\begin{minipage}{1\textwidth}
				\includegraphics[width=1\textwidth,height=0.102\textheight]{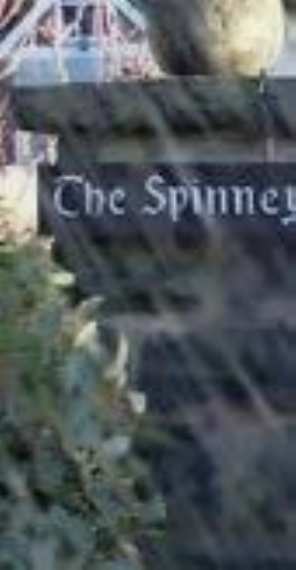}
				\centering \footnotesize22.05 / 0.78\\ LP\\
			\end{minipage}
		}
		\subfigure{
			\begin{minipage}{1\textwidth}
				\includegraphics[width=1\textwidth,height=0.101\textheight]{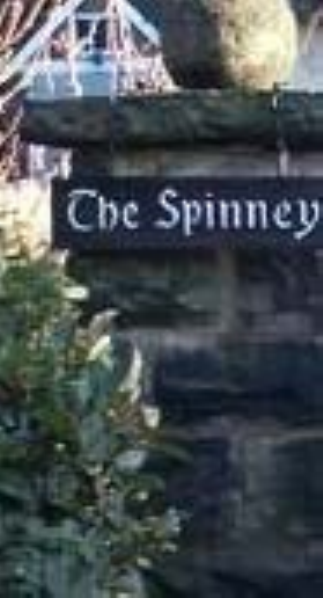}
				\centering \footnotesize 30.26 / 0.95\\MSPFN\\
			\end{minipage}
		}
	\end{minipage}
	\begin{minipage}{0.1\textwidth}
		\subfigure{
			\begin{minipage}{1\textwidth}
				\includegraphics[width=1\textwidth,height=0.102\textheight]{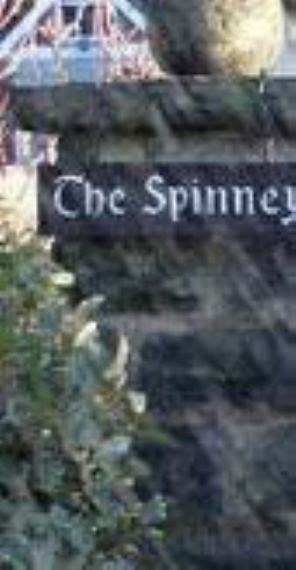}
				\centering \footnotesize 21.17 / 0.81\\ JORDER \\
			\end{minipage}
		}
		\subfigure{
			\begin{minipage}{1\textwidth}
				\includegraphics[width=1\textwidth,height=0.101\textheight]{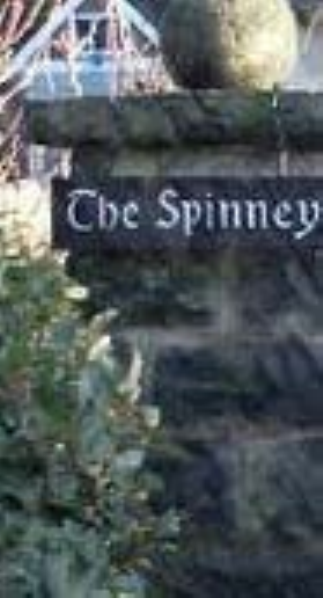}
				\centering \footnotesize 23.48 / 0.86\\DualGCN\\
			\end{minipage}
		}
	\end{minipage}
	\begin{minipage}{0.1\textwidth}
		\subfigure{
			\begin{minipage}{1\textwidth}
				\includegraphics[width=1\textwidth,height=0.102\textheight]{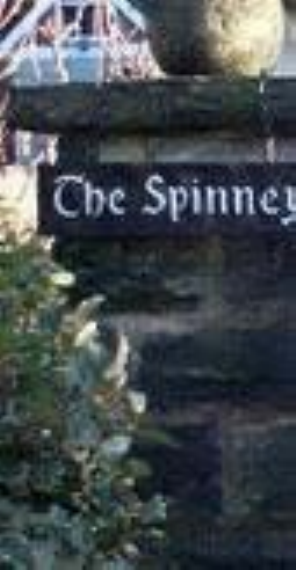}
				\centering \footnotesize25.92 / 0.89\\ DDN \\
			\end{minipage}
		}
		\subfigure{
			\begin{minipage}{1\textwidth}
				\includegraphics[width=1\textwidth,height=0.101\textheight]{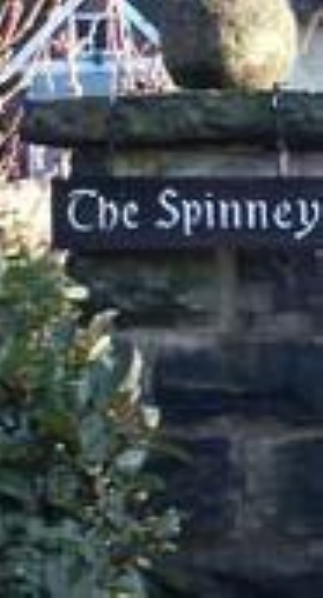}
				\centering \footnotesize 30.65 / 0.95\\ MPRNet\\
			\end{minipage}
		}
	\end{minipage}
	\begin{minipage}{0.1\textwidth}
		\subfigure{
			\begin{minipage}{1\textwidth}
				\includegraphics[width=1\textwidth,height=0.102\textheight]{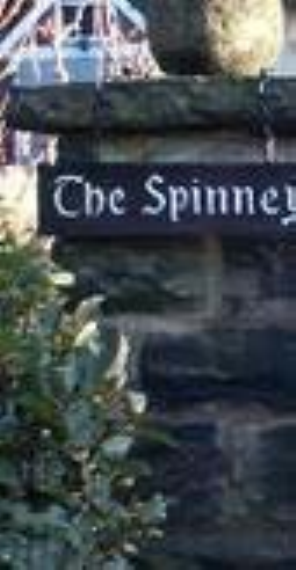}
				\centering \footnotesize29.80 / 0.95\\DualRes \\
			\end{minipage}
		}
		\subfigure{
			\begin{minipage}{1\textwidth}
				\includegraphics[width=1\textwidth,height=0.101\textheight]{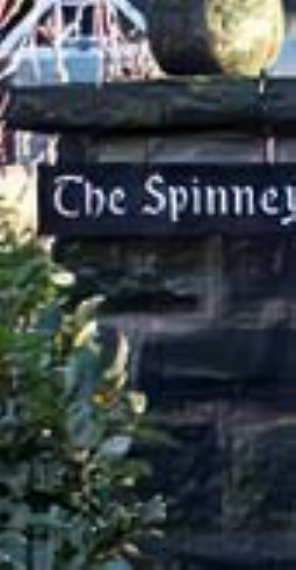}
				\centering \footnotesize 32.96 / 0.96\\Ours\\
			\end{minipage}
		}
	\end{minipage}
	
	\caption{Visual comparisons on the Rain1400 dataset. Our method effectively removes rain interference, which can be seen in the zoomed-in region. Compared with other methods, the result of our method restores the images with sharper context and is visually closer to the ground truth.  }
	\label{fig:syn1400com}
\end{figure*}
\subsection{Contrastive Prior}
Due to the particular distribution of rain, learning rain-free images via a deep network directly cannot generate favorable results. Motivated by contrastive learning~\cite{chen2020simple, dai2017contrastive}, it divides the existing data into two categories, i.e., positives and negatives, and learns a representation space in which the results are enforced closer to the positive pairs and away from the negative ones. In single image deraining, we take the observed rainy image~$\bold{I}$ as a negative case and the rain-free clear background~$\bold{B}$ as positive. The restored images are treated as the anchor. Recently, intermediate features of the pretrained networks have been proven useful in perceptual space~\cite{li2020single,wei2019single}. Therefore, we adopt VGG-19~\cite{simonyan2014very} as our latent representation space generator, denoted as~$\mathcal{G}$, and the proposed contrastive prior can be expressed as:
\begin{equation}
	\mathcal{L}_{CP}=\Psi(\mathcal{G}(\bold{I}), \mathcal{G}(\bold{B}), \mathcal{G}(\mathcal{N}(\bold{I},\omega))),\label{Lcp1}
\end{equation}
where~$\bold{B}$ and~$\bold{I}$ denote the positive and negative samples, respectively.~$\mathcal{N}(\bold{I},\omega)$ is the restored results. To make the most of the fixed pretrained model, we extract the intermediate features from multiple hidden layers and employ~$l_1$ regularization to strengthen the opposing constraint. Eq.~\eqref{Lcp1} can be further formulated as:
\begin{equation}
	\mathcal{L}_{CP}=\sum_{i=1}^n\rho_i\cdot\frac{\lVert\mathcal{G}_i (\bold{B})-\mathcal{G}_i (\mathcal{N}(\bold{I},\omega))\rVert_1}{\lVert\mathcal{G}_ i(\bold{I})-\mathcal{G}_i (\mathcal{N}(\bold{I},\omega))\rVert_1},
\end{equation}
where~$i$ represents the latent feature from the ~$i$-th layer and~$\rho_i$ is the weight parameter. In this paper, we set~$i=1,3,5,9,13$ and the corresponding weight~$\rho_i=\frac{1}{32},\frac{1}{16},\frac{1}{8},\frac{1}{4},1$, since the feature from a deeper layer covers more complicated perceptual information than the shallower layer. 

Consistency loss is also used to ensure the congruence between the reconstructed image and ground truth; it regularizes the reconstructed result from the two sides: one is to be accordant with the ground truth in the data field, and the other is to be aligned with the ground truth in the structure. Specifically, we adopt MSE and SSIM to achieve them. The MSE loss calculates the average squared distance between the predicted result~$\mathcal{N}(\bold{I},\omega)$ and label~$\bold{B}$. SSIM measures the similarity from three points, i.e., brightness, contrast and structure. Consistency loss can be reformulated as:
\begin{equation}
	\mathcal{L}_{Con}=\mathcal{L}_{SSIM}(\mathcal{N}(\bold{I},\omega),\bold{B})+\gamma_1\mathcal{L}_{MSE}(\mathcal{N}(\bold{I},\omega),\bold{B}),
\end{equation}
\begin{equation}
	\mathcal{L}_{SSIM}=1-SSIM(\mathcal{N}(\bold{I},\omega),\bold{B}),
\end{equation}
where~$\gamma_1$ is a hyperparameter to balance the two terms. Therefore, the full loss function of the proposed RDMC can be further defined as:
\begin{equation}
	\mathcal{L}=\mathcal{L}_{SSIM}+\gamma_1\mathcal{L}_{MSE}+\gamma_2\mathcal{L}_{CP}.
\end{equation}
In the above formula, $\gamma_2$ is another hyperparameter. The consistency loss, composed of the first and second items, can be treated as a positive-oriented constraint, trending to make the reconstructed result incline to the positives. In contrast, the contrastive prior in the last item achieves a further negative-oriented regularization, where the opposing principle between the restored results and the rainy images advances the generalization in complicated scenarios. We demonstrate the effectiveness  of the contrastive prior in experiments.

\begin{figure*}[h]
	\centering
	\begin{minipage}{0.35\textwidth}
		
		\subfigure{
			\begin{minipage}{1\textwidth}
				\includegraphics[width=1\textwidth,height=0.240\textheight]{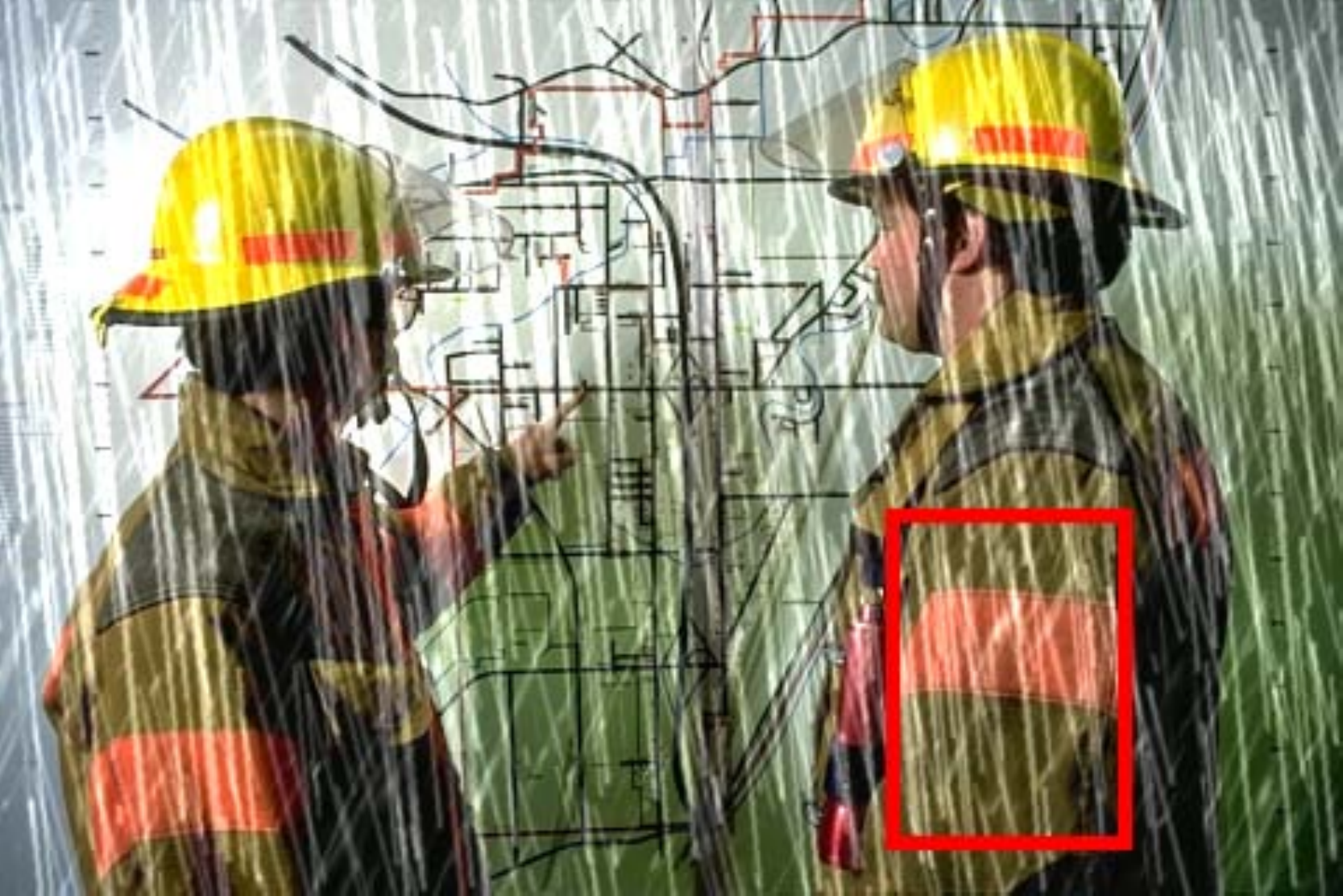}
				\centering  \footnotesize 14.13 / 0.54\\Input\\
			\end{minipage}
		}
	\end{minipage}
	\begin{minipage}{0.1\textwidth}
		\subfigure{
			\begin{minipage}{1\textwidth}
				\includegraphics[width=1\textwidth,height=0.102\textheight]{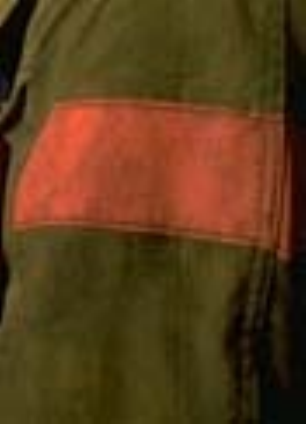}
				\centering \footnotesize- / -\\ Ground truth 	\\
			\end{minipage}
		}
		\subfigure{
			\begin{minipage}{1\textwidth}
				\includegraphics[width=1\textwidth,height=0.101\textheight]{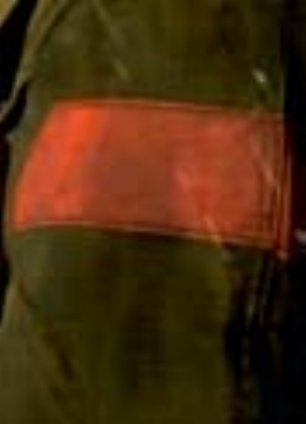}
				\centering \footnotesize 23.68 / 0.86\\SIRR\\
			\end{minipage}
		}
	\end{minipage}
	\begin{minipage}{0.1\textwidth}
		\subfigure{
			\begin{minipage}{1\textwidth}
				\includegraphics[width=1\textwidth,height=0.102\textheight]{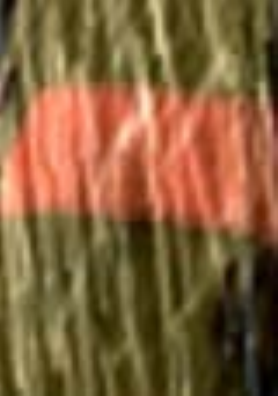}
				\centering \footnotesize 15.79 / 0.44\\DSC\\
			\end{minipage}
		}
		\subfigure{
			\begin{minipage}{1\textwidth}
				\includegraphics[width=1\textwidth,height=0.101\textheight]{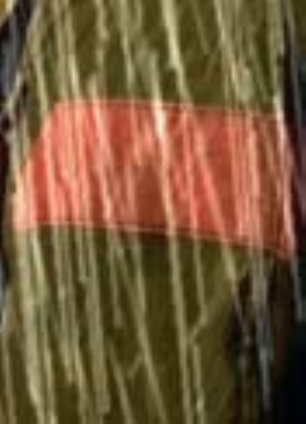}
				\centering \footnotesize 16.83 / 0.55\\ \vspace{-0.2em}Syn2Real\\
			\end{minipage}
		}
	\end{minipage}
	\begin{minipage}{0.1\textwidth}
		\subfigure{
			\begin{minipage}{1\textwidth}
				\includegraphics[width=1\textwidth,height=0.102\textheight]{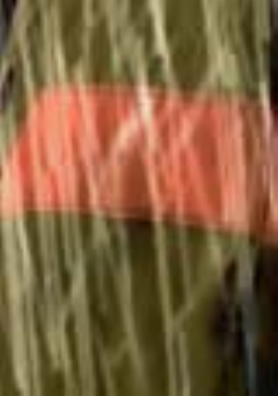}
				\centering \footnotesize16.29 / 0.51\\ LP\\
			\end{minipage}
		}
		\subfigure{
			\begin{minipage}{1\textwidth}
				\includegraphics[width=1\textwidth,height=0.101\textheight]{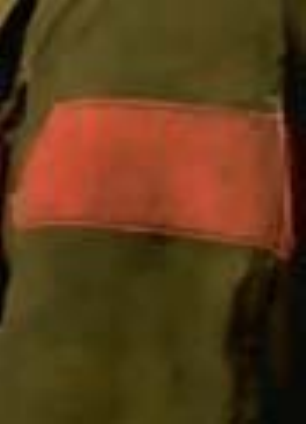}
				\centering \footnotesize 28.23 / 0.94\\MSPFN\\
			\end{minipage}
		}
	\end{minipage}
	\begin{minipage}{0.1\textwidth}
		\subfigure{
			\begin{minipage}{1\textwidth}
				\includegraphics[width=1\textwidth,height=0.102\textheight]{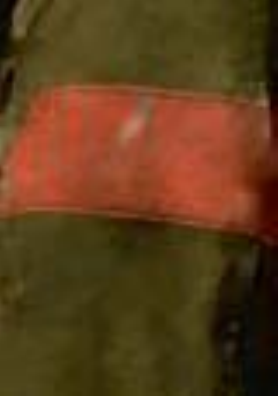}
				\centering \footnotesize 26.49 / 0.91\\ JORDER \\
			\end{minipage}
		}
		\subfigure{
			\begin{minipage}{1\textwidth}
				\includegraphics[width=1\textwidth,height=0.101\textheight]{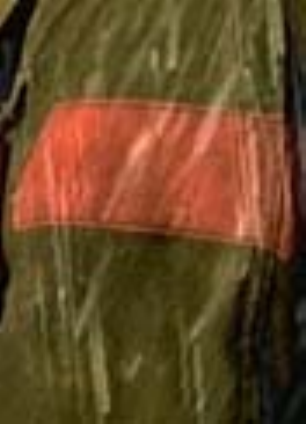}
				\centering \footnotesize 22.59 / 0.82\\DualGCN\\
			\end{minipage}
		}
	\end{minipage}
	\begin{minipage}{0.1\textwidth}
		\subfigure{
			\begin{minipage}{1\textwidth}
				\includegraphics[width=1\textwidth,height=0.102\textheight]{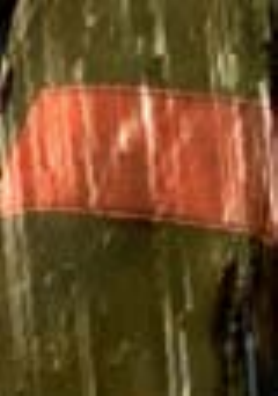}
				\centering \footnotesize19.30 / 0.72\\ DDN \\
			\end{minipage}
		}
		\subfigure{
			\begin{minipage}{1\textwidth}
				\includegraphics[width=1\textwidth,height=0.101\textheight]{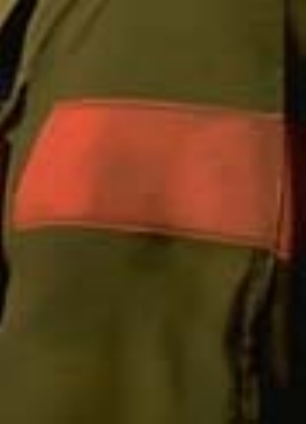}
				\centering \footnotesize 31.65 / 0.96\\ MPRNet\\
			\end{minipage}
		}
	\end{minipage}
	\begin{minipage}{0.1\textwidth}
		\subfigure{
			\begin{minipage}{1\textwidth}
				\includegraphics[width=1\textwidth,height=0.102\textheight]{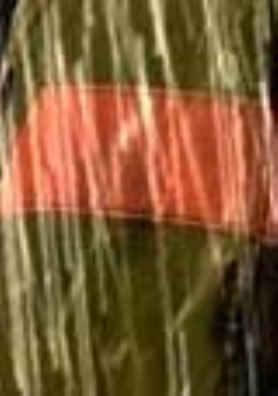}
				\centering \footnotesize16.12 / 0.57\\DualRes \\
			\end{minipage}
		}
		\subfigure{
			\begin{minipage}{1\textwidth}
				\includegraphics[width=1\textwidth,height=0.101\textheight]{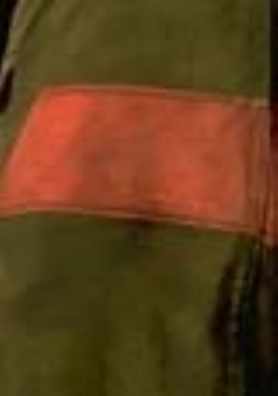}
				\centering \footnotesize 32.97 / 0.97\\Ours\\
			\end{minipage}
		}
	\end{minipage}	
	\caption{Image deraining comparisons on the Rain100H dataset, where the heavy and intensive streaks seriously degrade the background. Our method preserves more details without oversmoothing and additional artifacts, outperforming the others in terms of both the visual results and metric values reported below. }
	\label{fig:synHcom}
\end{figure*}
\begin{table*}[h]
	\begin{center}
		\centering
		\caption{Quantitative comparison (i.e., SSIM, PSNR) between our method and the state-of-the-art rain removal methods on five benchmarks. The best and the second results are highlighted in \textbf{bold} and \underline{underlined}, respectively.}
		\label{tab:Derain}
		\begin{tabular}{c|cc|cc|cc|cc|cc}
			\hline
			{Dataset}&\multicolumn{2}{c|}{Rain12~\cite{li2016rain}}  &\multicolumn{2}{c|}{Rain100L~\cite{yang2016joint}}&\multicolumn{2}{c|}{Rain100H~\cite{yang2016joint}} &\multicolumn{2}{c|}{Rain1200~\cite{zhang2018density}} &\multicolumn{2}{c}{Rain1400~\cite{fu2017removing}}\\
			\hline
			{ Measure} &{PSNR}~$\uparrow$    &{SSIM}~$\uparrow$  &{PSNR}~$\uparrow$    &{SSIM}~$\uparrow$  &{PSNR}~$\uparrow$    &{SSIM}~$\uparrow$  &{PSNR}~$\uparrow$   &{SSIM}~$\uparrow$  &{PSNR}~$\uparrow$&  {SSIM}~$\uparrow$    \\
			\hline
			DSC~\cite{luo2015removing}&29.1389 &0.8207 &27.4178 &0.8086&15.6636 &0.4225 &31.4905&0.9146&31.3033&0.9207\\
			LP\cite{li2016rain}&32.3302&0.9054&26.5103&0.8517& 14.2671&0.5444 &27.5523&0.8479&25.8068&0.8344\\
			JORDER~\cite{yang2016joint}&32.4107&0.8279&36.1142&0.9733&25.3598&0.8804&\underline{32.0062} & 0.9321&\underline{33.9800}&0.9502 \\
			DDN~\cite{fu2017removing}&33.4106&0.9475&28.4199&0.9154&16.6076&0.6240&30.0023&0.9041&30.9721&0.9116\\
			DualRes~\cite{liu2019dual}&30.9270&0.9482&27.3160&0.9111&13.6072&0.4695&29.7376&0.9132&30.6061&0.9420\\
			SIRR~\cite{wei2019semi}&24.2892&0.8662&23.5825&0.8749&15.1231&0.5637&24.6693&0.8621&25.9621&0.8956\\
			Syn2Real~\cite{yasarla2020syn2real}&28.4570&0.9413&24.2665&0.9145&15.2492&0.4905&23.6463&0.8564&24.8149&0.8943\\
			MSPFN~\cite{jiang2020multi}&34.2599&0.9703&30.8781&0.9469&26.8753&0.8963&30.6765&0.9333&31.4019&0.9529\\
			DualGCN~\cite{fu2021rain}&\underline{35.7646}& 0.9715&\textbf{41.1077}&\textbf{0.9940}&18.2625&0.6812&24.7635&0.8404&27.5807&0.9077\\
			MPRNet~\cite{zamir2021multi}&35.5590 &\underline{0.9809}&35.0235&0.9735&\textbf{29.0971}&\textbf{0.9259}&31.6222&\textbf{0.9414}&32.5380&\underline{0.9619}\\
			Ours&\textbf{35.9006}&\textbf{0.9813}&\underline{36.5683}&\underline{0.9760}&\underline{27.5944}&\underline{0.9211}&\textbf{34.3410}&\underline{0.9336}&\textbf{36.4820}&\textbf{0.9873}\\
			\hline
		\end{tabular}
	\end{center}
%	\footnotesize
\end{table*}
\newcommand{\tabincell}[2]{\begin{tabular}{@{}#1@{}}#2\end{tabular}}
\begin{table*}[h]
	\centering
	\caption{Quantitative comparison~(i.e., NIQE and PI) on the real-world dataset. The best and the second best results are highlighted in \textbf{bold} and \underline{underlined}, respectively. }
	\begin{tabular}{>{\centering}p{1.3cm}|>{\centering}p{1.cm}|>{\centering}p{1.cm}|>{\centering}p{1.cm}|>{\centering}p{1.cm}|>{\centering}p{1.cm}|>{\centering}p{1.cm}|>{\centering}p{1.cm}|>{\centering}p{1.cm}|>{\centering}p{1.cm}|>{\centering}p{1.cm}|>{\centering}p{1.cm}  }
		\hline
		Methods&\tabincell{c}{DSC\tabularnewline\cite{luo2015removing}}&\tabincell{c}{LP\tabularnewline\cite{li2016rain}}&\tabincell{c}{JORDER\tabularnewline\cite{yang2016joint}}&\tabincell{c}{DDN\tabularnewline\cite{fu2017removing}}&\tabincell{c}{DualRes\tabularnewline\cite{liu2019dual}}&\tabincell{c}{SIRR\tabularnewline\cite{wei2019semi}} &\tabincell{c}{Syn2Real\tabularnewline\cite{yasarla2020syn2real}}&\tabincell{c}{MSPFN\tabularnewline\cite{jiang2020multi}}&\tabincell{c}{DualGCN\tabularnewline\cite{fu2021rain}}&\tabincell{c}{MPRNet\tabularnewline\cite{zamir2021multi}}&Ours\tabularnewline \hline
		NIQE~$\downarrow$&3.9211 & 3.6523&3.0269&3.6158&3.2889&\underline{2.9551}&3.2997&3.1349&2.9981&3.2740&\textbf{2.9063} \tabularnewline
		PI~$\downarrow$   &3.3572 &3.2797 &2.2449&2.4901&2.3293&\underline{2.1720}&2.2931&2.4049&2.1858&2.4120&\textbf{2.1252} \tabularnewline\hline
	\end{tabular}
	\label{tab:real}
\end{table*}

\section{Experiments}
To evaluate the proposed RDMC, five synthetic datasets and real-world images are adopted. We conduct an extensive comparison with ten state-of-the-art deraining methods, including the conventional model methods, DSC~\cite{luo2015removing} and LP~\cite{li2016rain}, and the deep learning methods, JORDER~\cite{yang2016joint}, DDN~\cite{fu2017removing}, DualRes~\cite{liu2019dual}, SIRR~\cite{wei2019semi}, Syn2Real~\cite{yasarla2020syn2real}, MSPFN~\cite{jiang2020multi}, DualGCN~\cite{fu2021rain} and MPRNet~\cite{zamir2021multi}. All the methods are obtained from their official release. 
For an objective assessment, the widely adopted Peak Signal to Noise Ratio~(PSNR) and Structural Similarity~(SSIM) are employed to quantify the similarity between the results and the rain-free references. For real-world data, the nonreference valuation Natural Image Quality Evaluator~(NIQE) and Perceptual Index~(PI) are adopted. Additionally, we also demonstrate the superiority of RDMC in subsequent applications~(e.g., object detection and semantic segmentation), where the Intersection over Union~(IoU), mean Average Precision~(mAP) and mean Pixel Accuracy~(mPA) are applied for comprehensive verification.

\subsection{Implementation Details}
During training, we cropped the synthetic Rain200~\cite{yang2016joint} into~$128\times128$ patches and obtained a total of~$6000$ image pairs. These image pairs are divided into the training and validation sets, and each of them contains~$3000$ images. The network is implemented using PyTorch and trained on an Nvidia GeForce RTX 2080 Ti GPU.
To search for the optimal network structure, the neural architecture search processing is composed of two alternating phases. For the first phase, we optimize the model on the training set to update the weights of the candidate operation. For the second phase, we optimize the model on the validation set to update the weights of the internal parameters. The ADAM algorithm~\cite{Kingma2014Adam} is adopted as an optimizer. The batch size is set as~$8$. The learning rate is initialized as $1\times10^{-3}$ and reduced to $20\%$ after~$30, 50$ and~$80$ epochs. We train the network for~$150$ epochs with the above setting.
Then, we load the searched model and fine-tune it on the whole dataset for~$100$ epochs, where the learning rate is initialized to $1\times10^{-4}$ and multiplied by~$0.2$ at~$30, 50$ and~$80$ epochs. For the loss function, the hyperparameters of the balance weights,~$\gamma_1$ and~$\gamma_2$, are both set as~$0.1$. 

\begin{figure*}[htb]
	\centering
	\setlength{\tabcolsep}{1pt}
	\begin{tabular}{cccccccccccc}
		\includegraphics[width=0.15\textwidth, height=0.107\textheight]{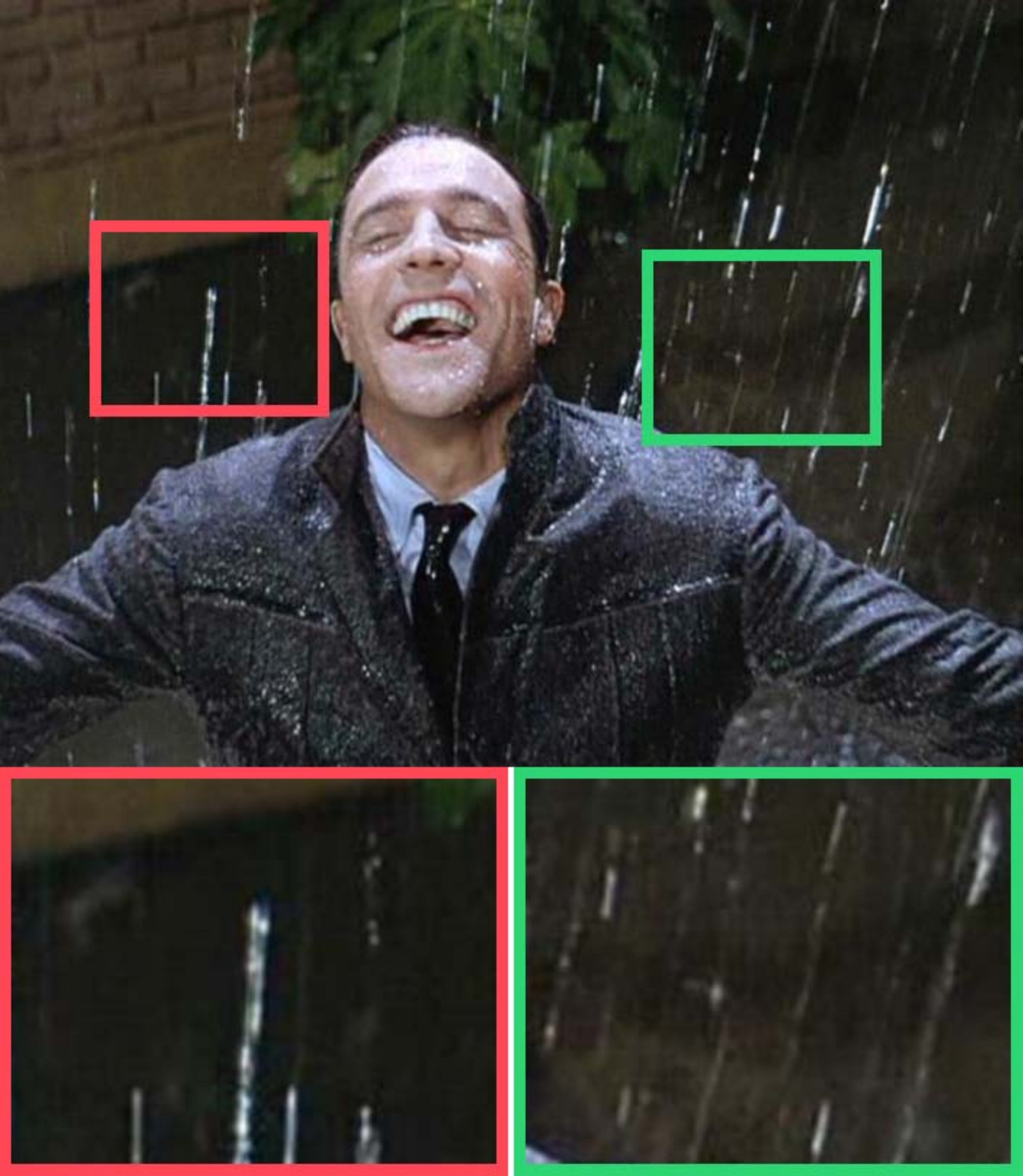}
		&\includegraphics[width=0.15\textwidth, height=0.107\textheight]{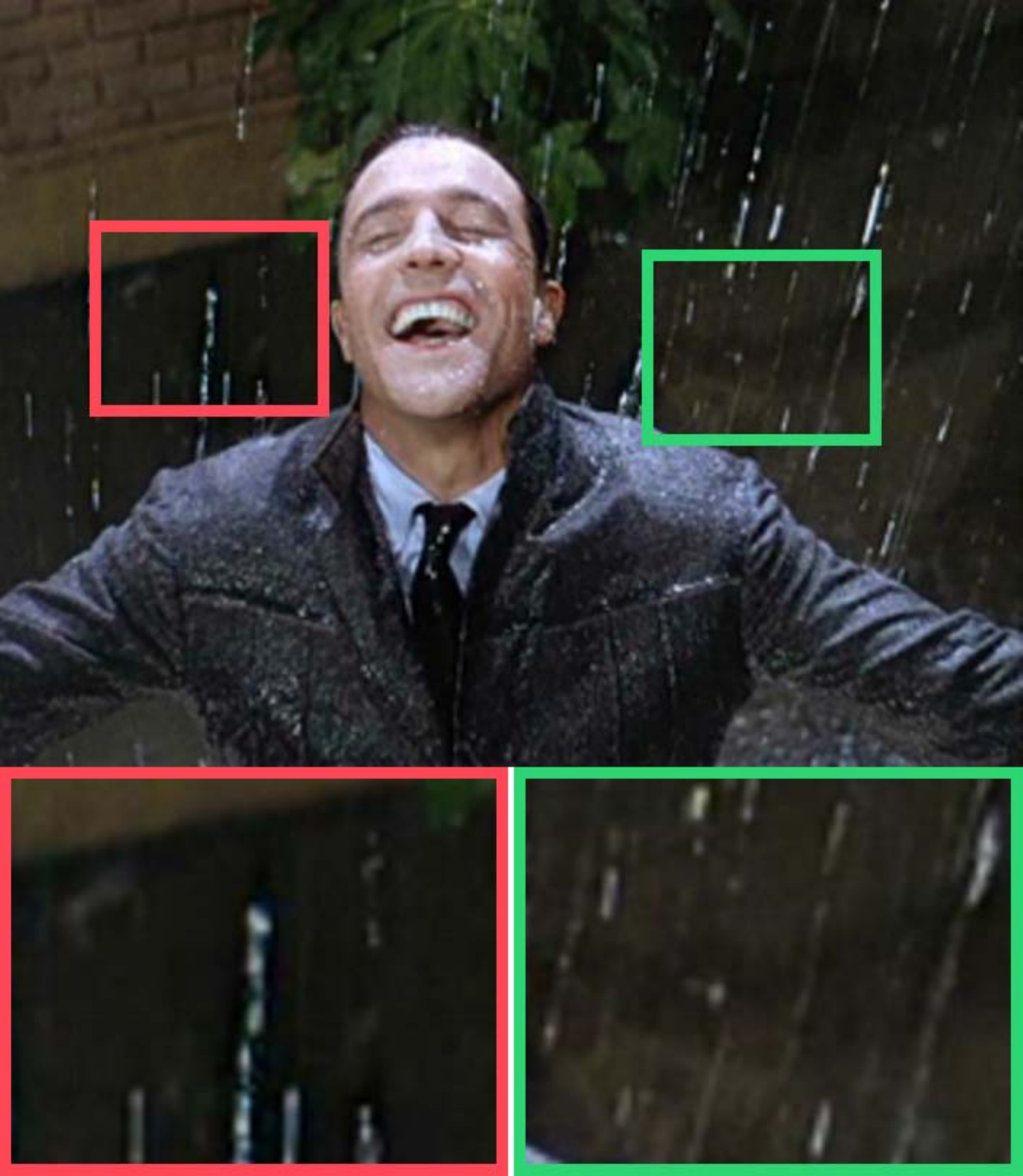}
		&\includegraphics[width=0.15\textwidth, height=0.107\textheight]{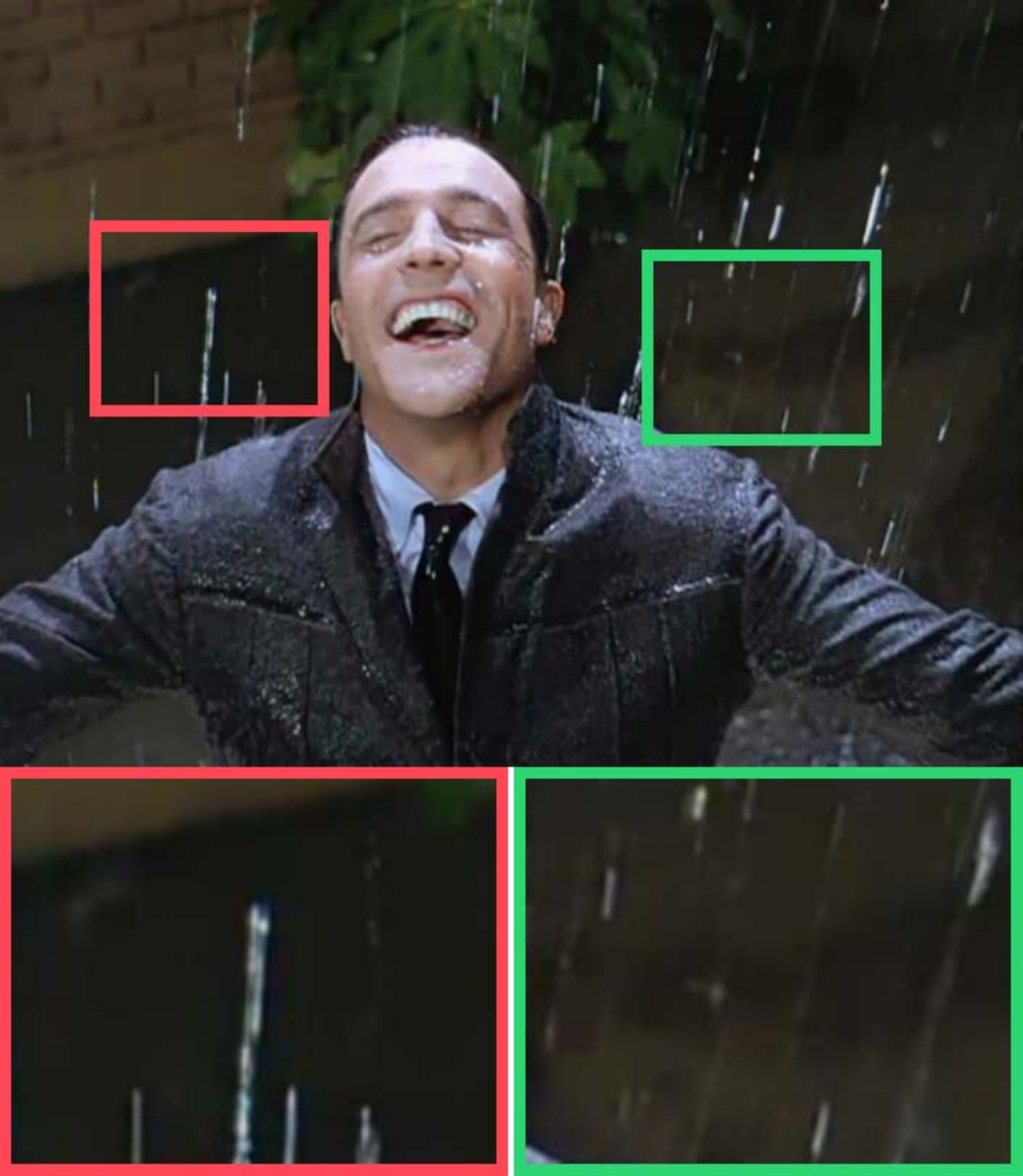}
		&\includegraphics[width=0.15\textwidth, height=0.107\textheight]{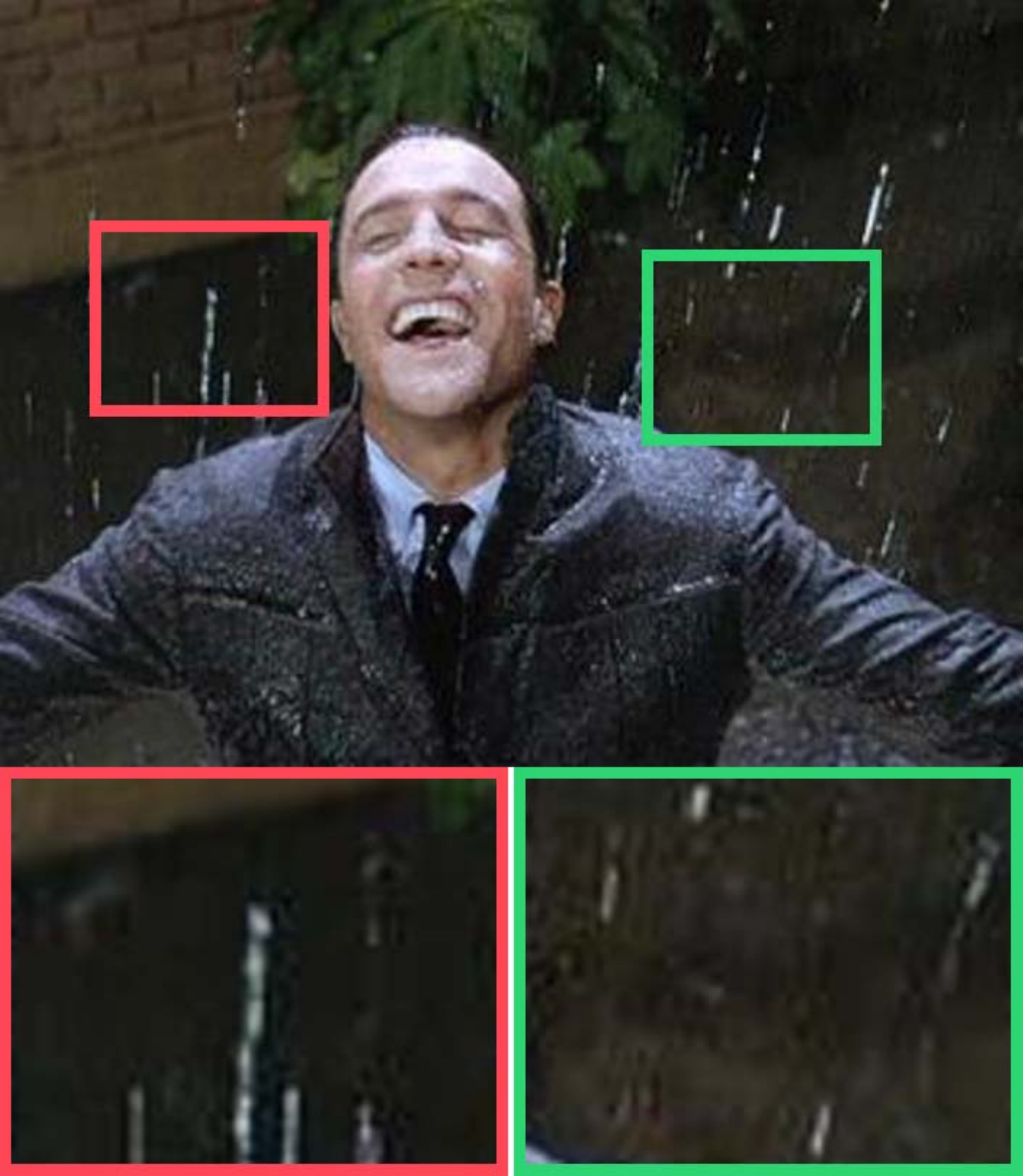}
		&\includegraphics[width=0.15\textwidth, height=0.107\textheight]{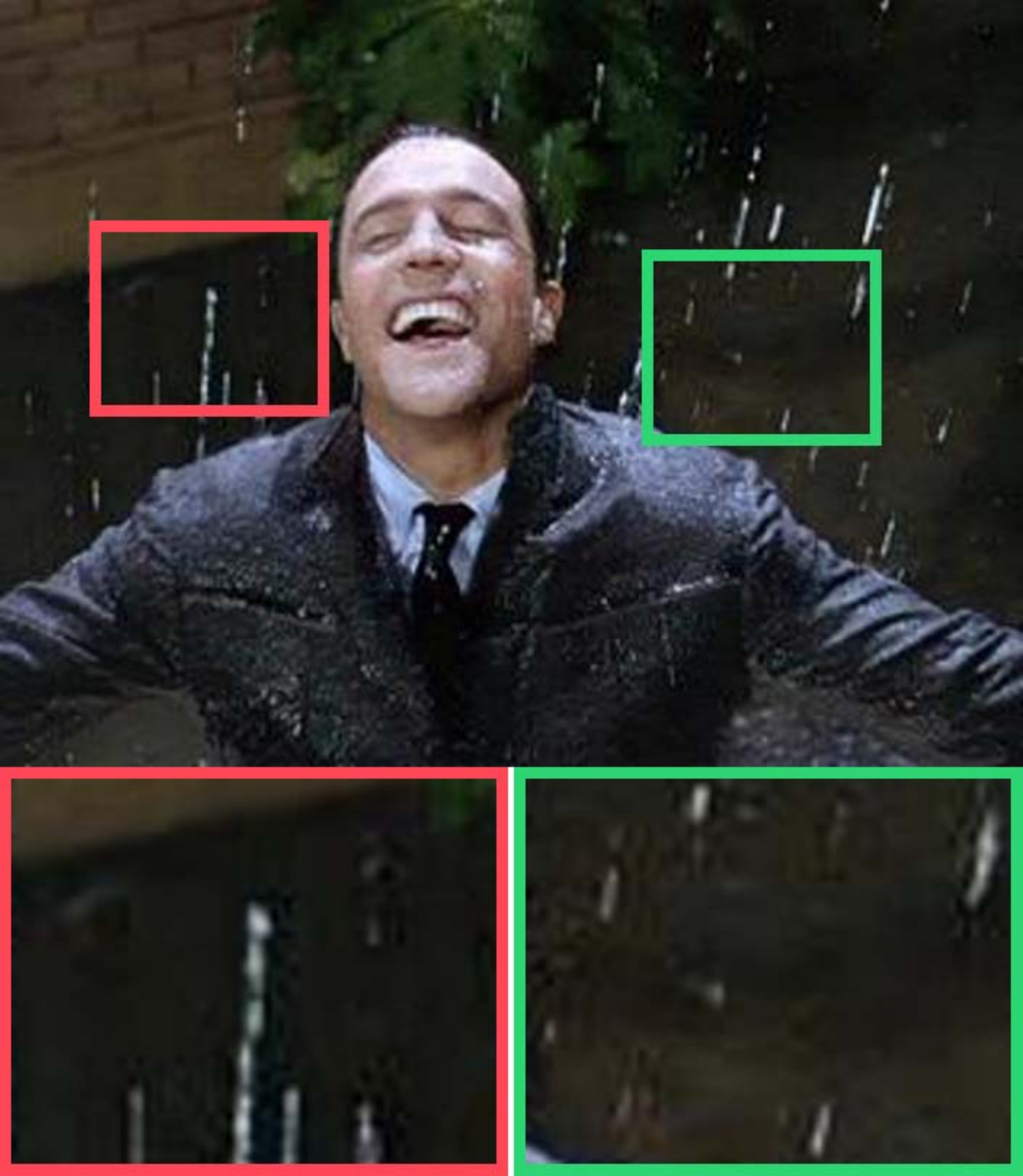}
		&\includegraphics[width=0.15\textwidth, height=0.107\textheight]{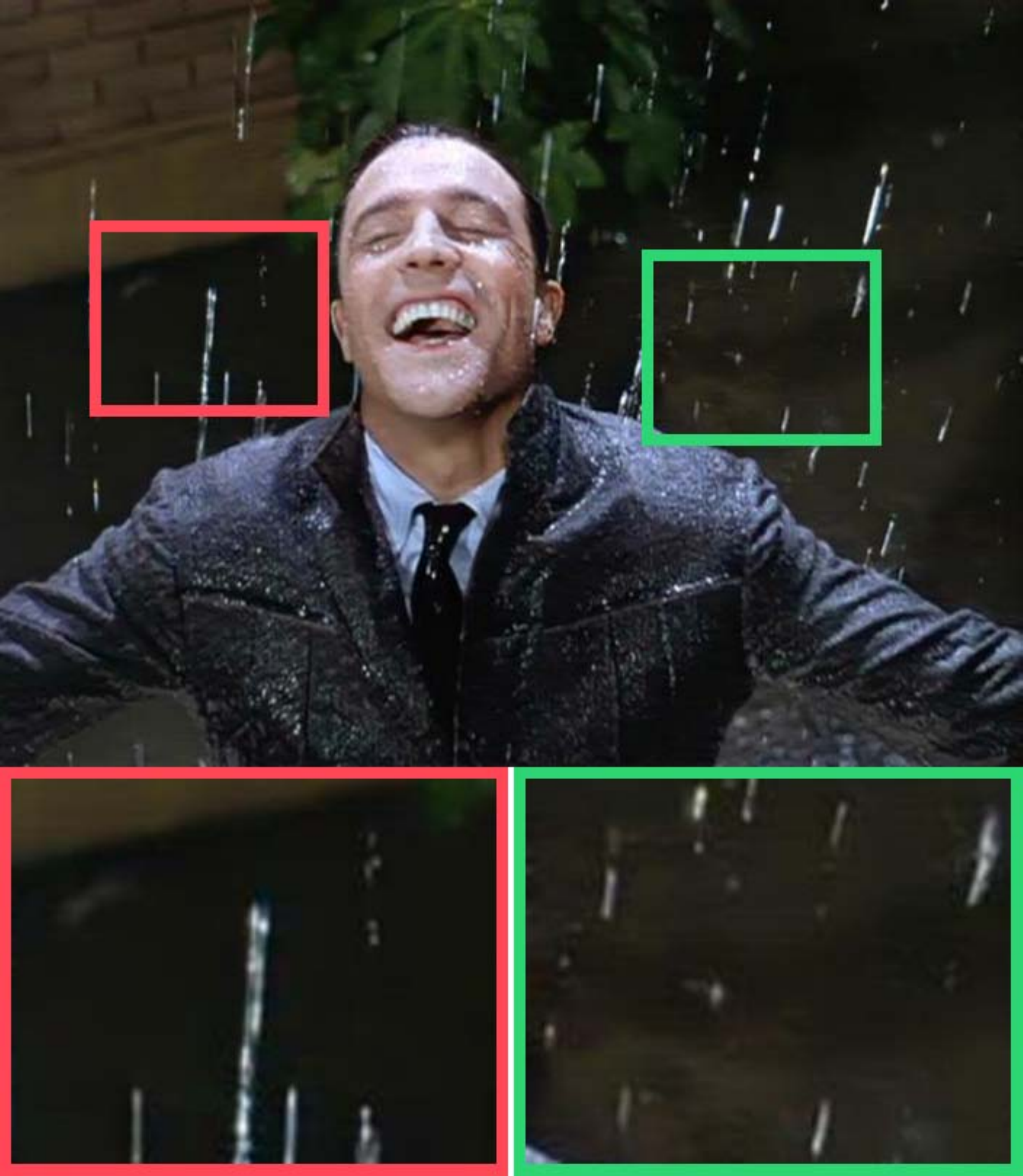}\\
		5.4490 / 4.8421&5.3212 / 4.2889&5.0940 / 3.9293 &4.3629 / 3.1749&4.9519 / 3.7619&4.5472 / 3.3335\\
		Input&DSC&LP&JORDER&DDN&DualRes\\
		\includegraphics[width=0.15\textwidth, height=0.107\textheight]{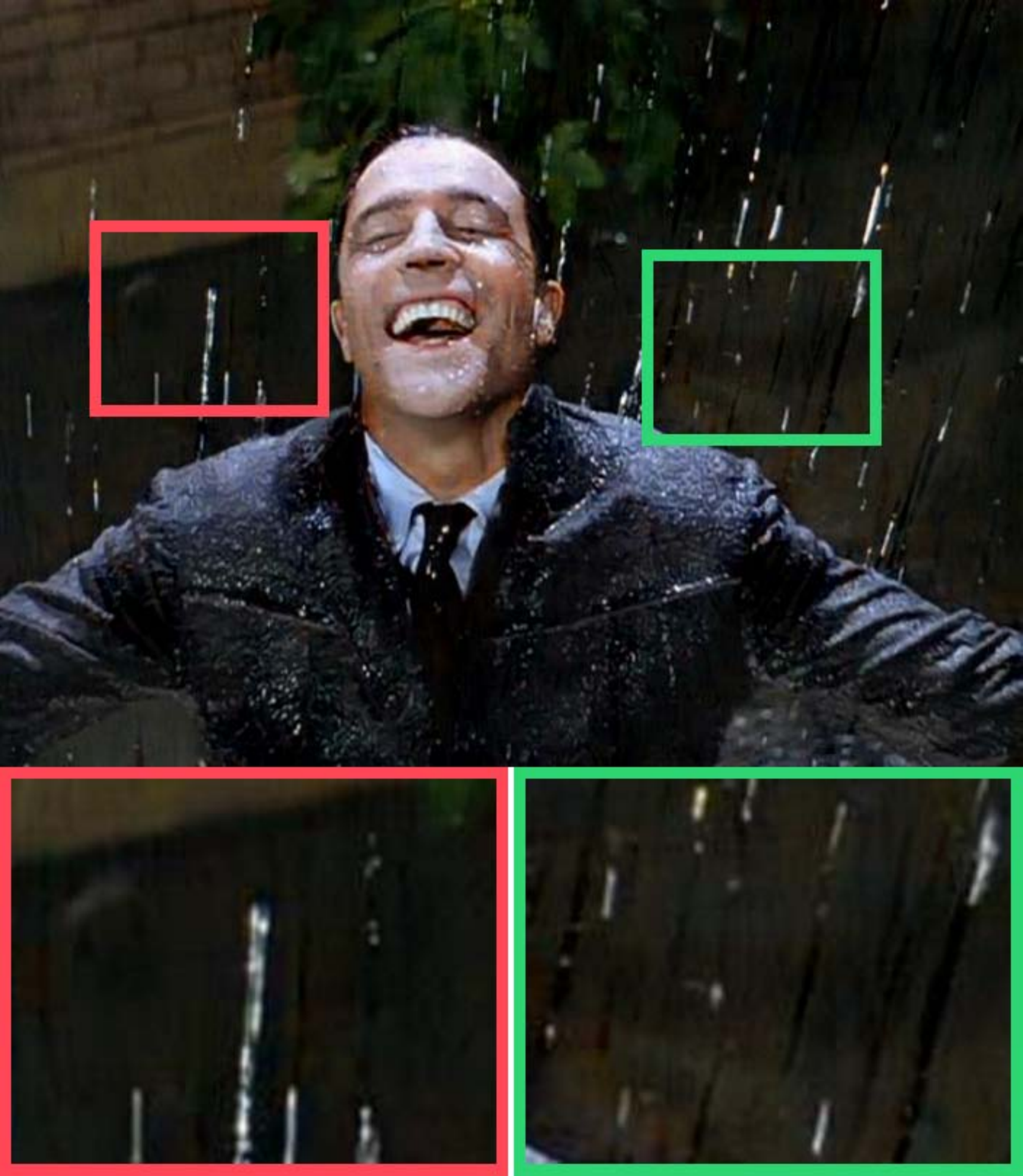}
		&\includegraphics[width=0.15\textwidth, height=0.107\textheight]{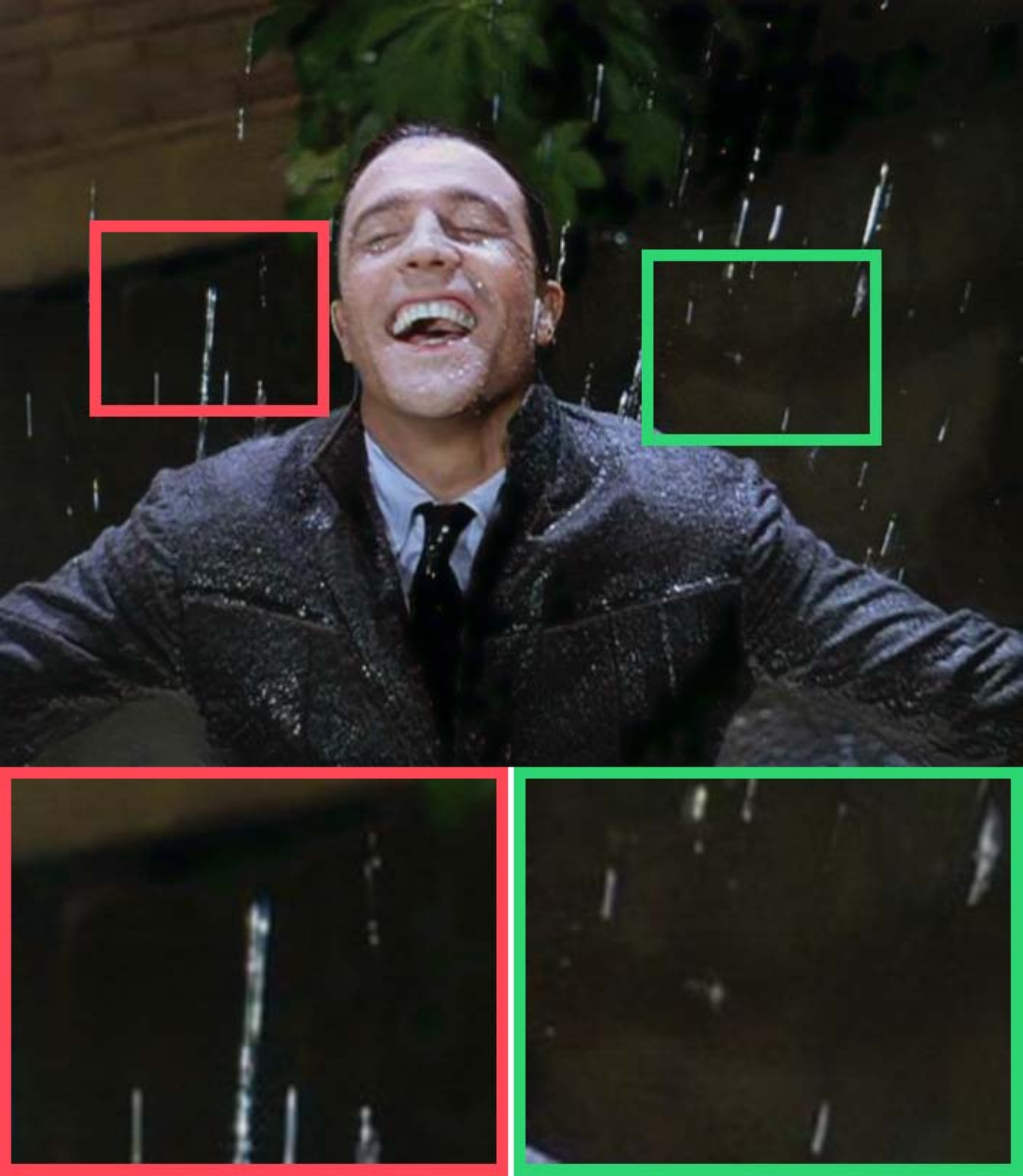}
		&\includegraphics[width=0.15\textwidth, height=0.107\textheight]{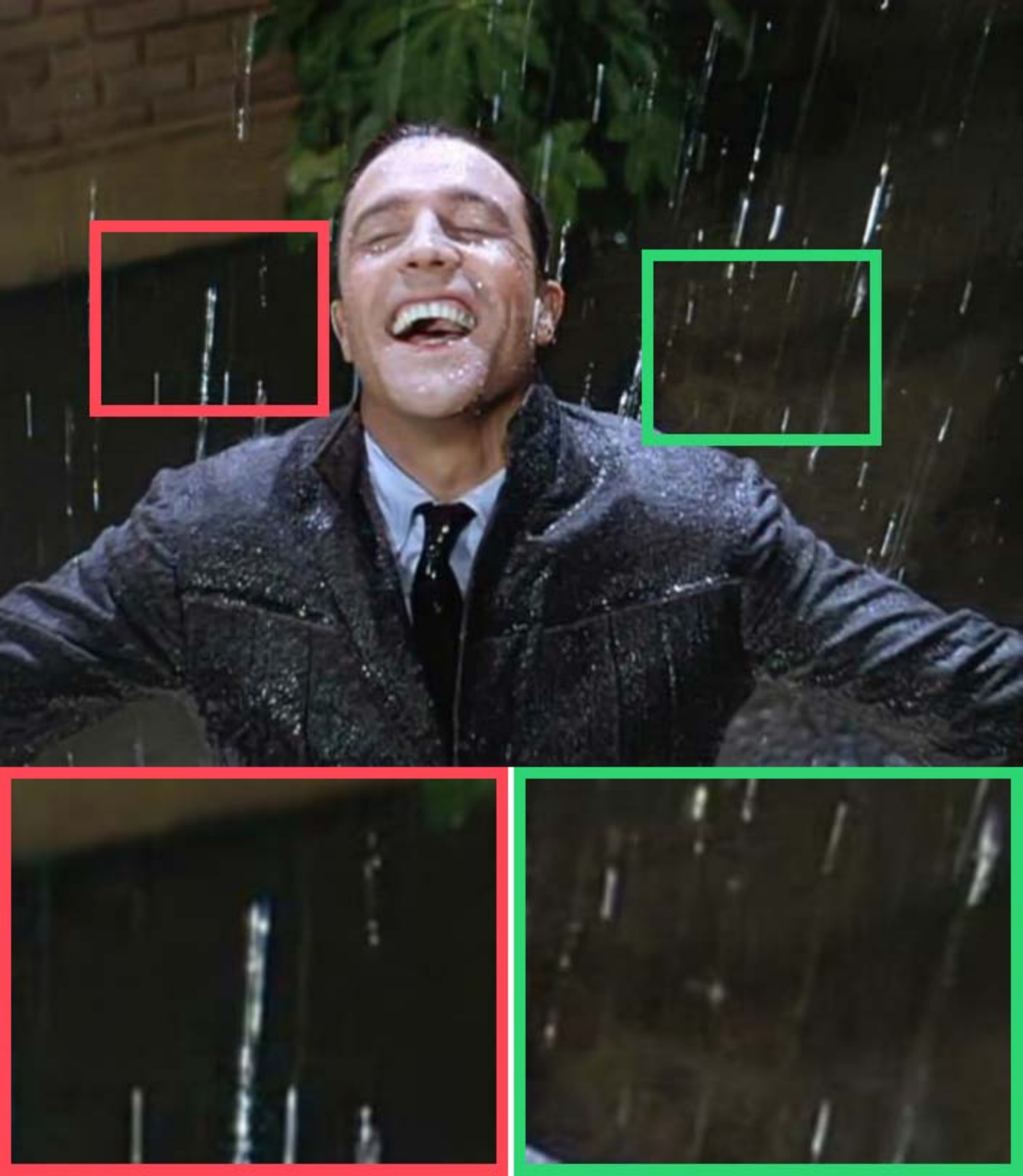}
		&\includegraphics[width=0.15\textwidth, height=0.107\textheight]{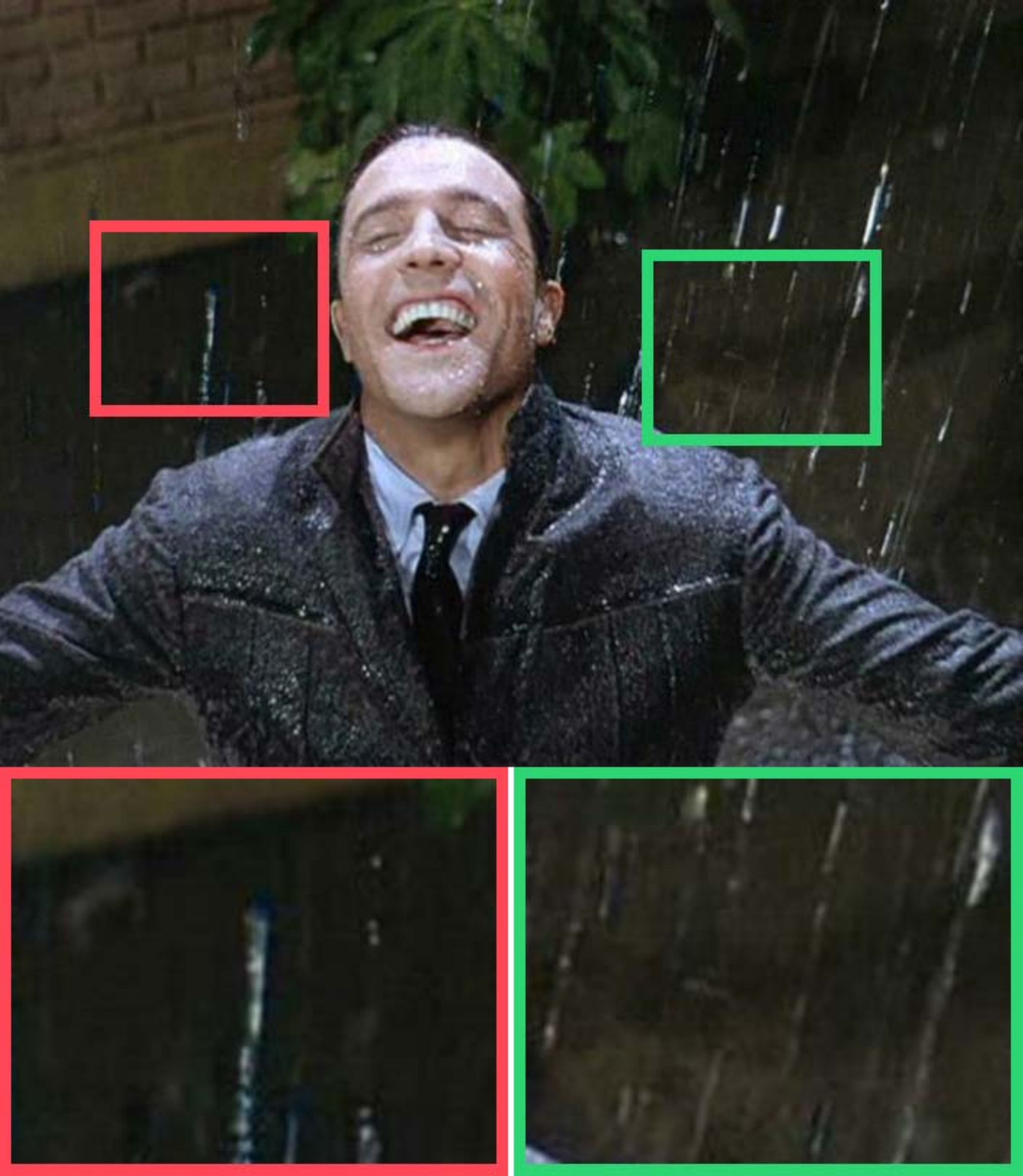}
		&\includegraphics[width=0.15\textwidth, height=0.107\textheight]{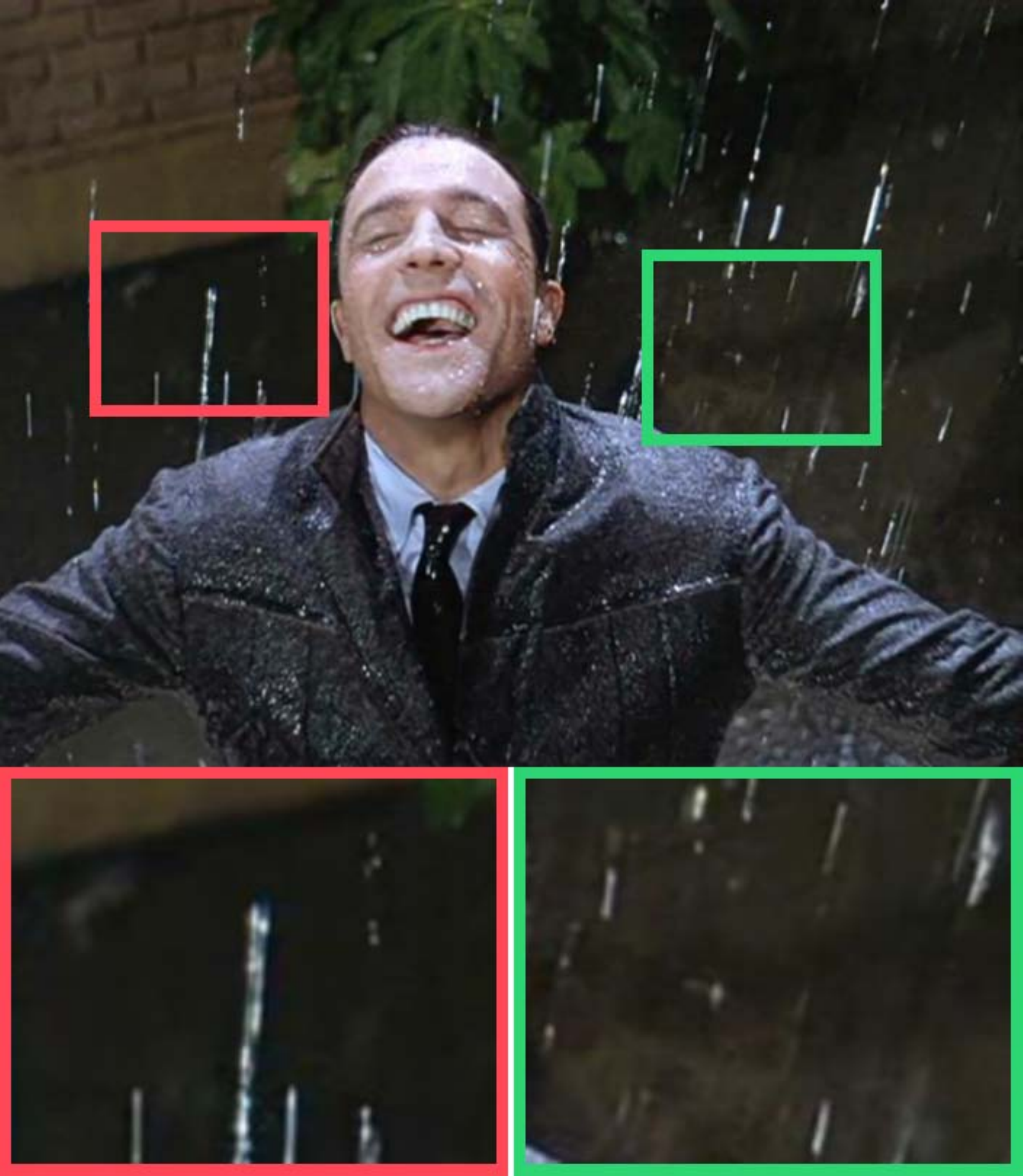}
		&\includegraphics[width=0.15\textwidth, height=0.107\textheight]{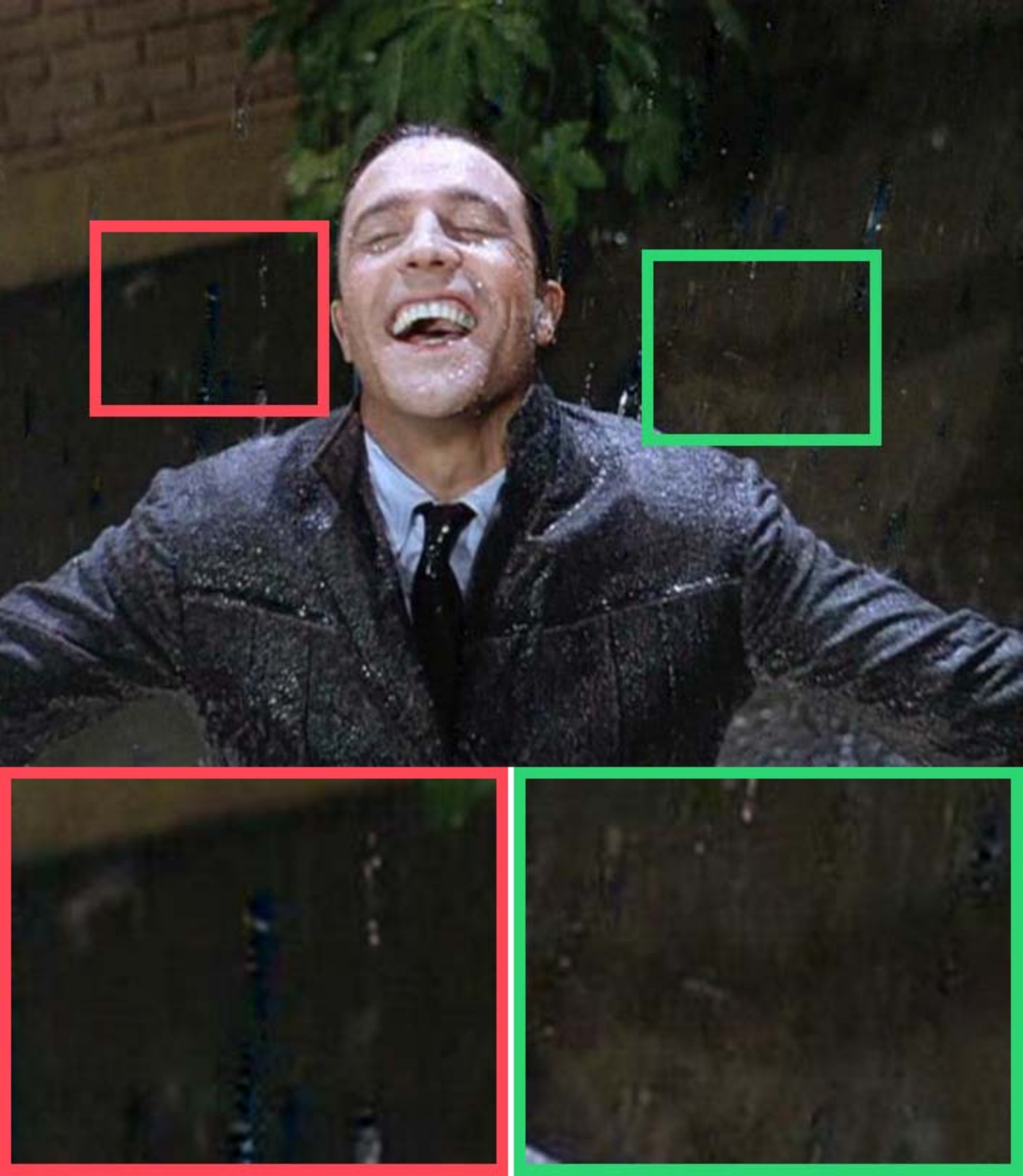}	\\
		4.1595 / 2.9909&4.2058 / 3.0556&4.3164 / 3.1527&4.5585 / 3.4098&4.3357 / 3.1609&4.1114 / 2.8558\\
		SIRR &Syn2Real&MSPFN&DualGCN&MPRNet&Ours\\
		\includegraphics[width=0.15\textwidth, height=0.104\textheight]{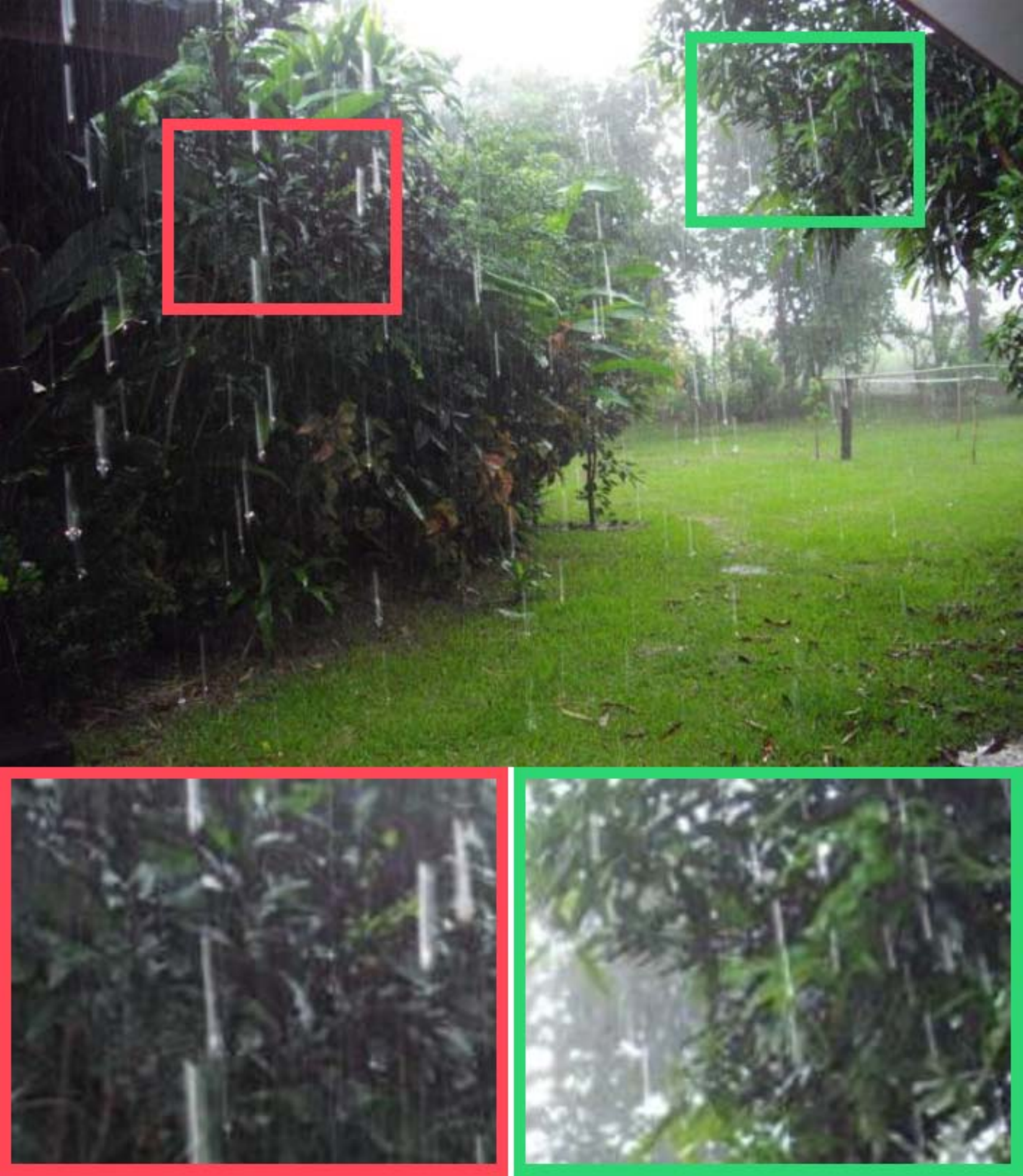}
		&\includegraphics[width=0.15\textwidth, height=0.104\textheight]{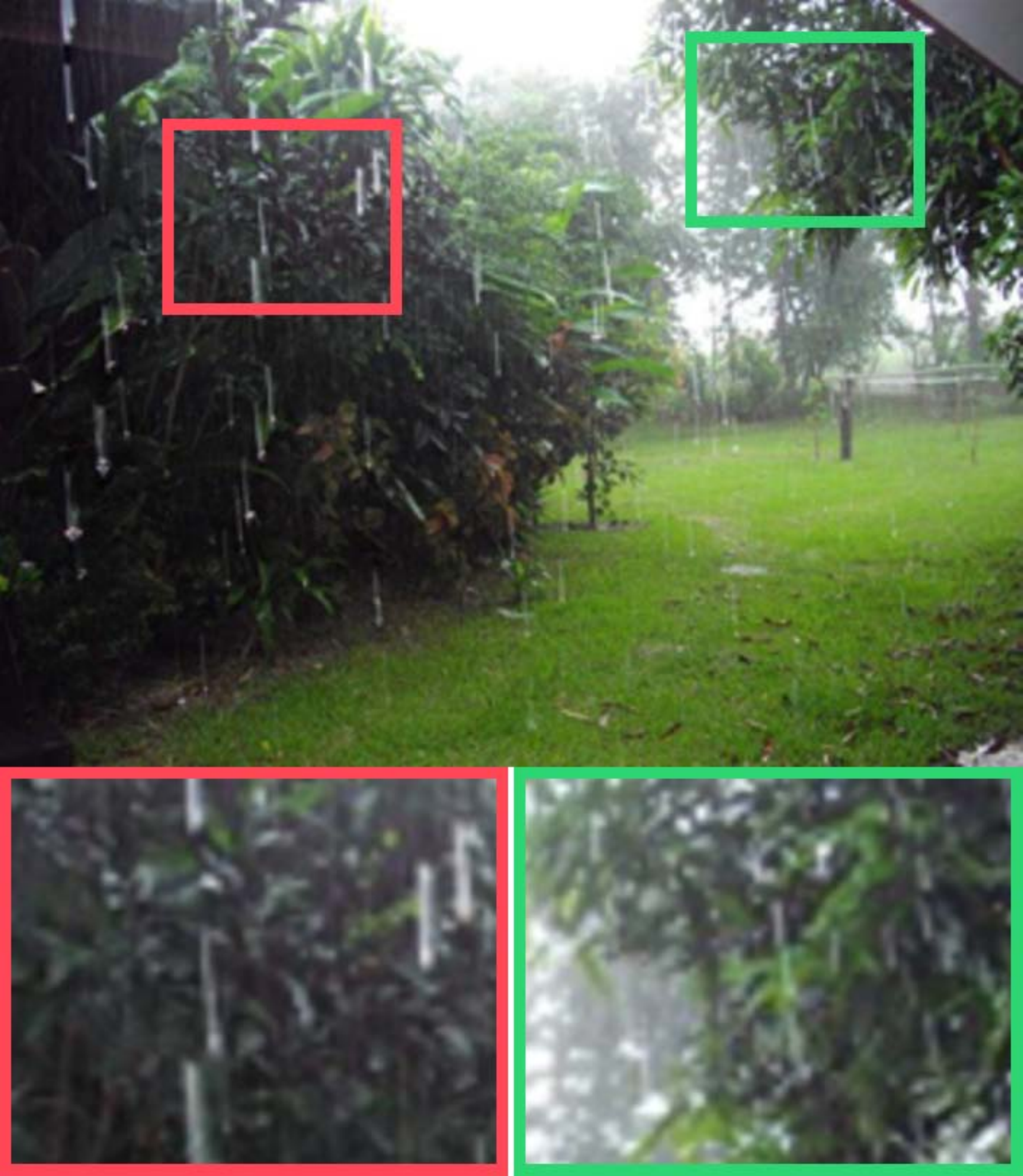}
		&\includegraphics[width=0.15\textwidth, height=0.104\textheight]{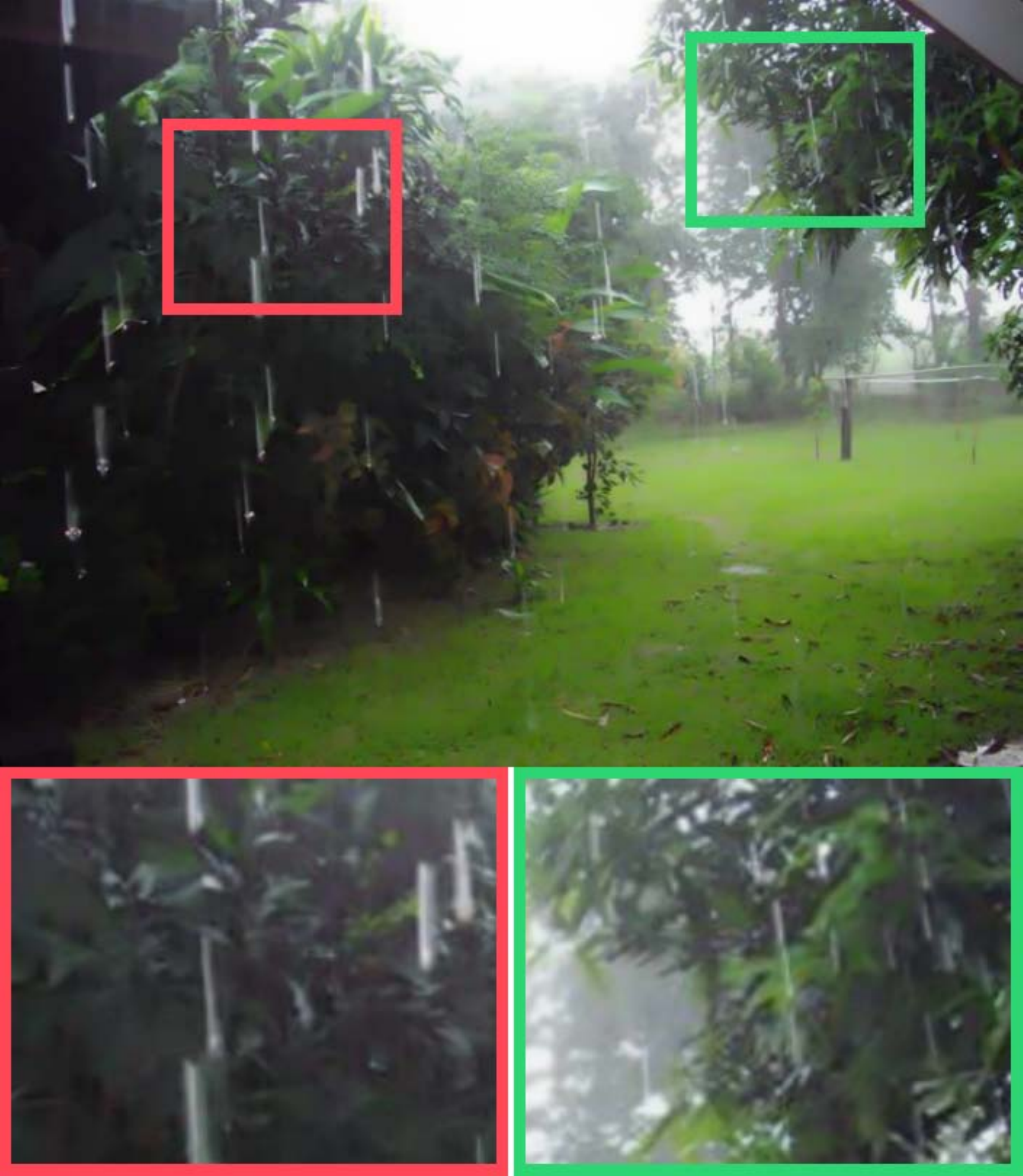}
		&\includegraphics[width=0.15\textwidth,height=0.104\textheight]{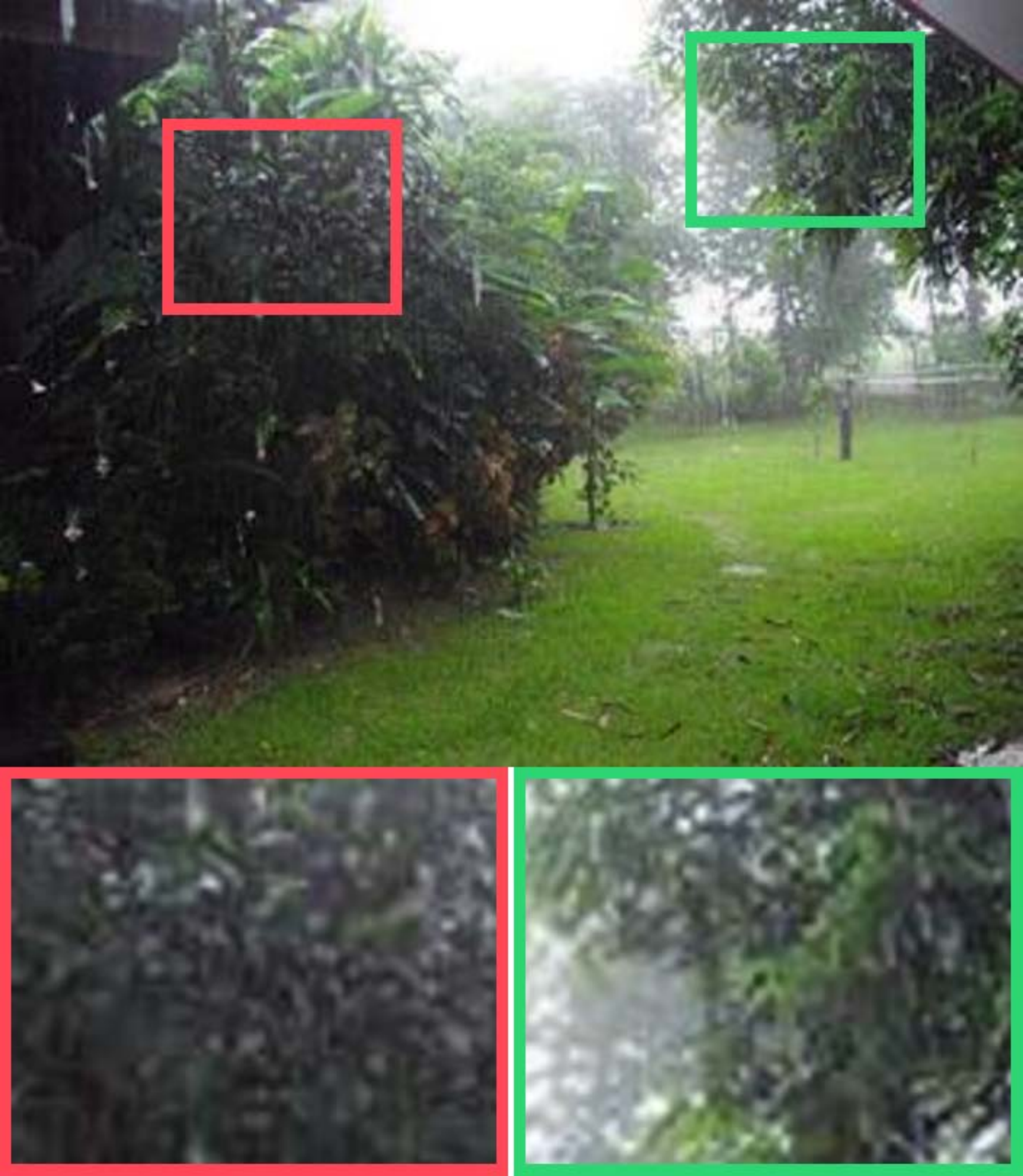}
		&\includegraphics[width=0.15\textwidth, height=0.104\textheight]{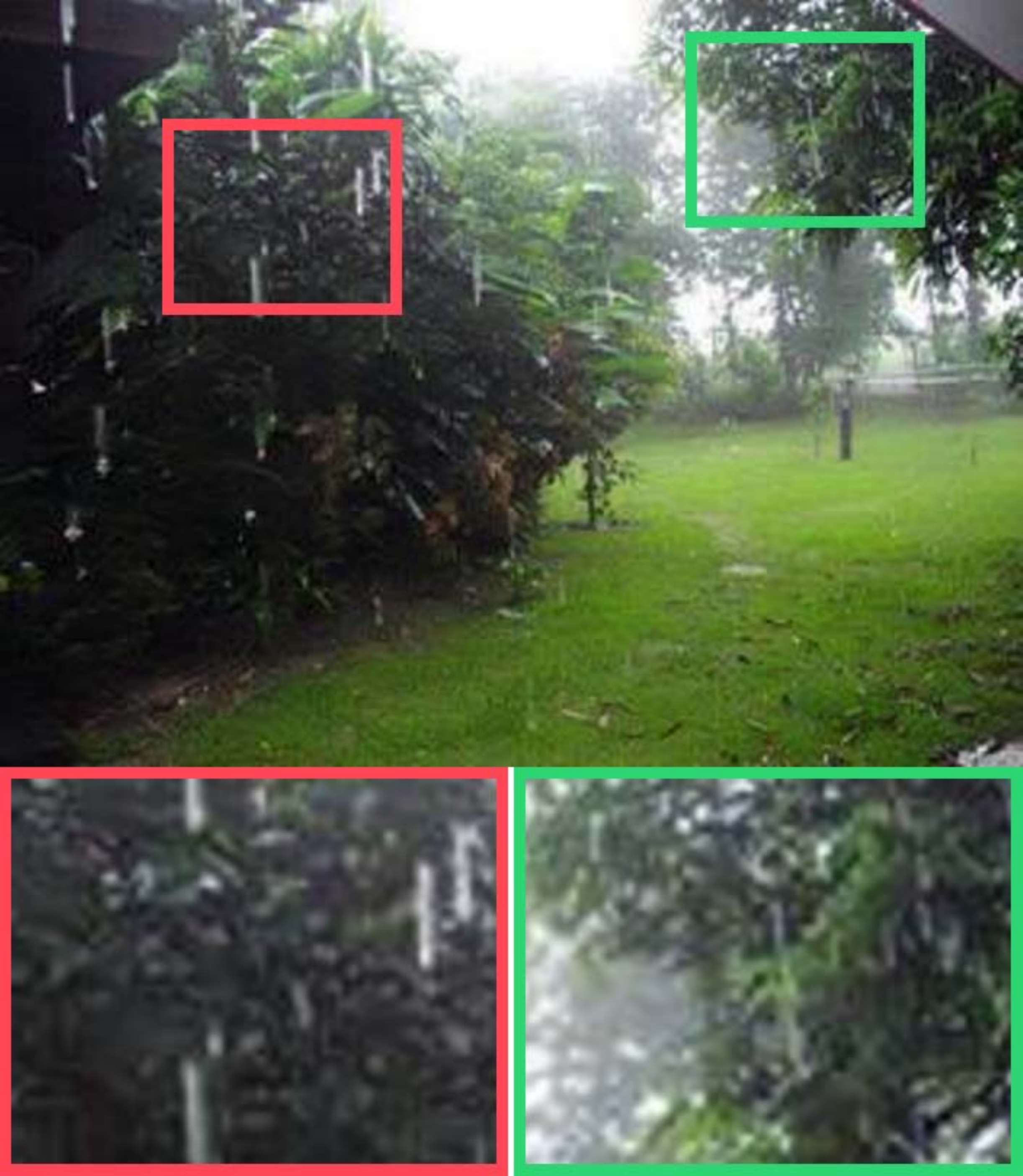}
		&\includegraphics[width=0.15\textwidth, height=0.104\textheight]{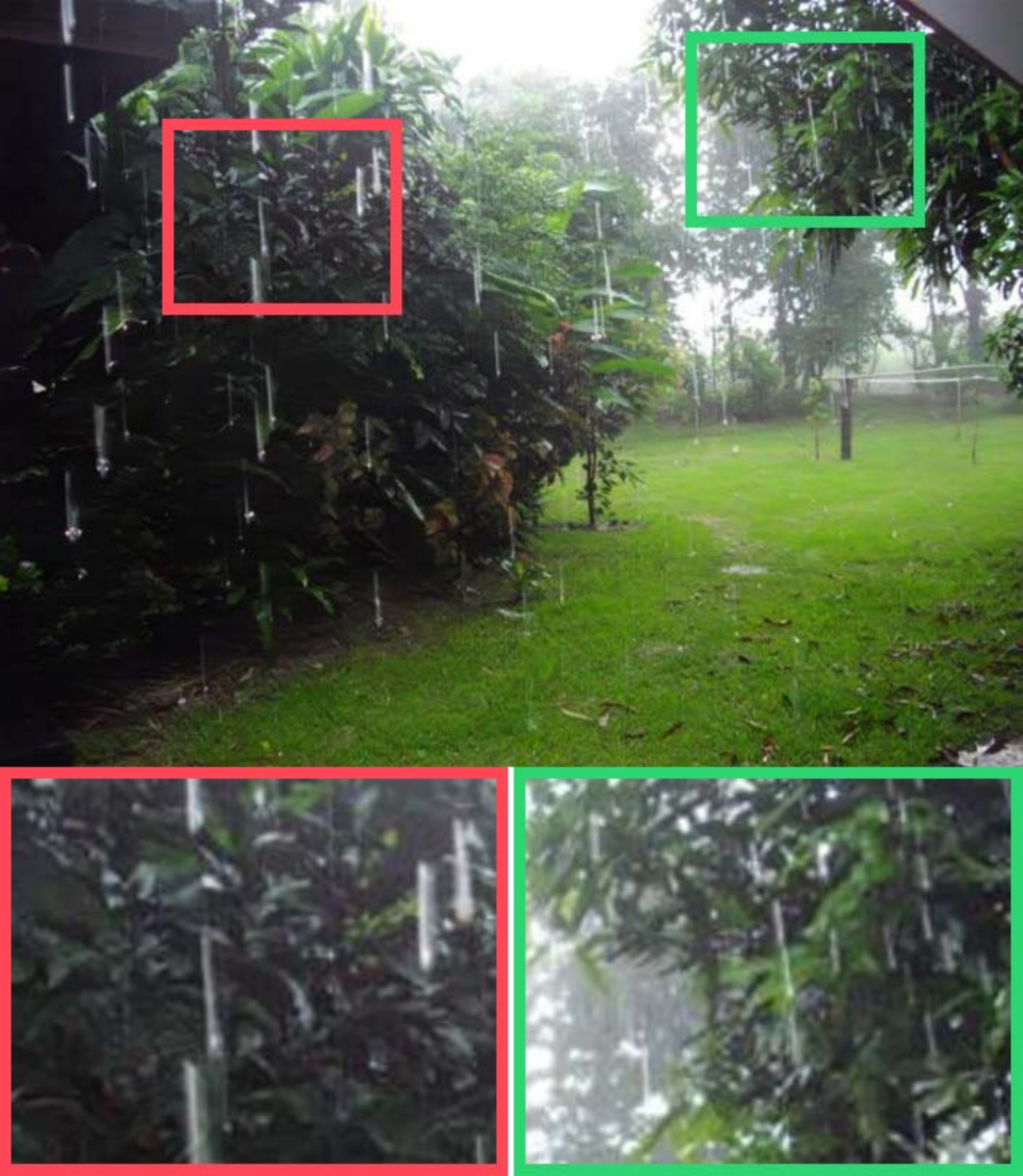}\\
		4.4139 / 3.2297&4.3105 / 3.0256&3.6106 / 2.8301&3.3612 / 2.3443&3.9741 / 2.7590&3.5448 / 2.4597\\
		Input&DSC&LP&JORDER&DDN&DualRes\\
		\includegraphics[width=0.15\textwidth, height=0.104\textheight]{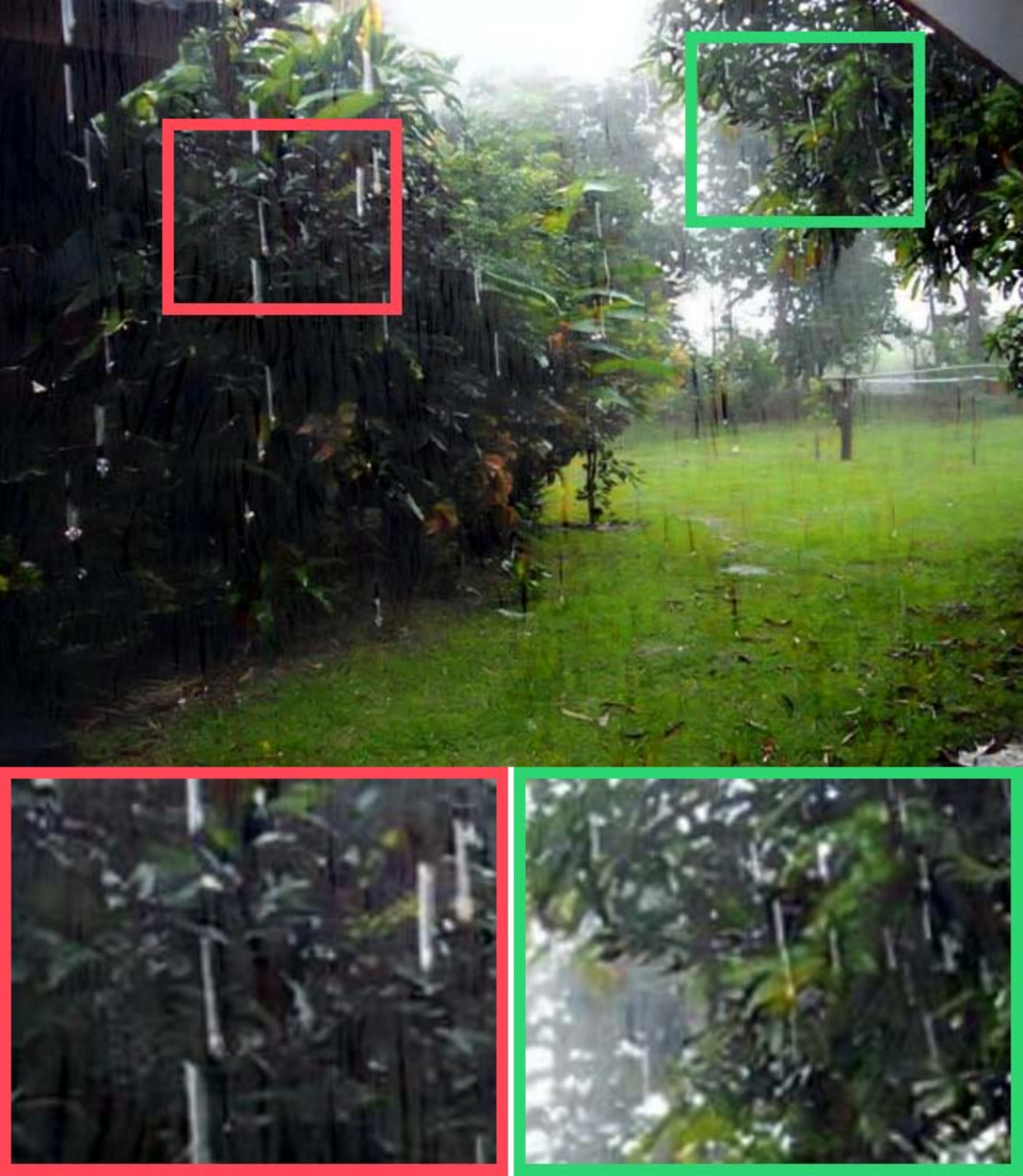}
		&\includegraphics[width=0.15\textwidth, height=0.104\textheight]{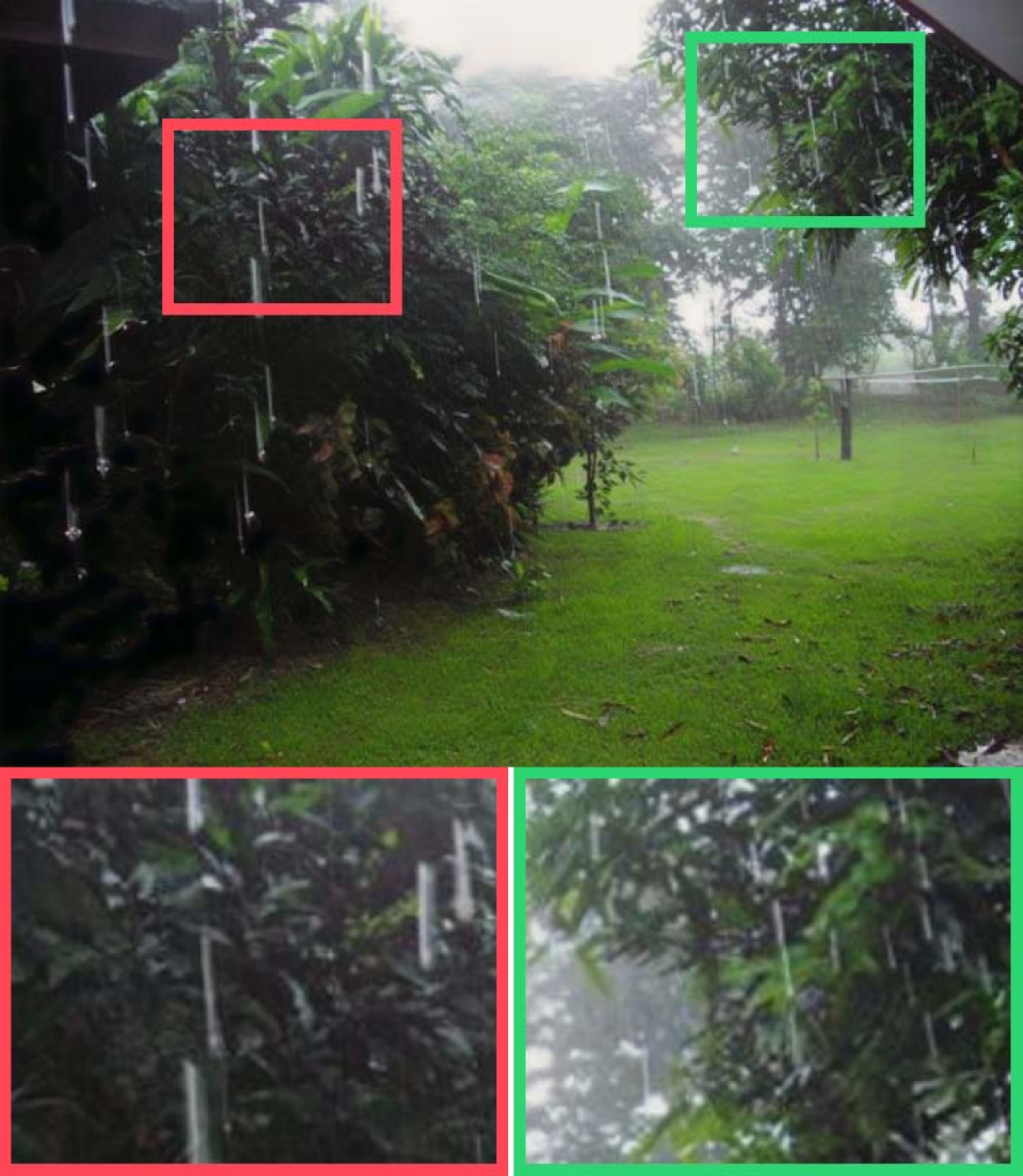}
		&\includegraphics[width=0.15\textwidth, height=0.104\textheight]{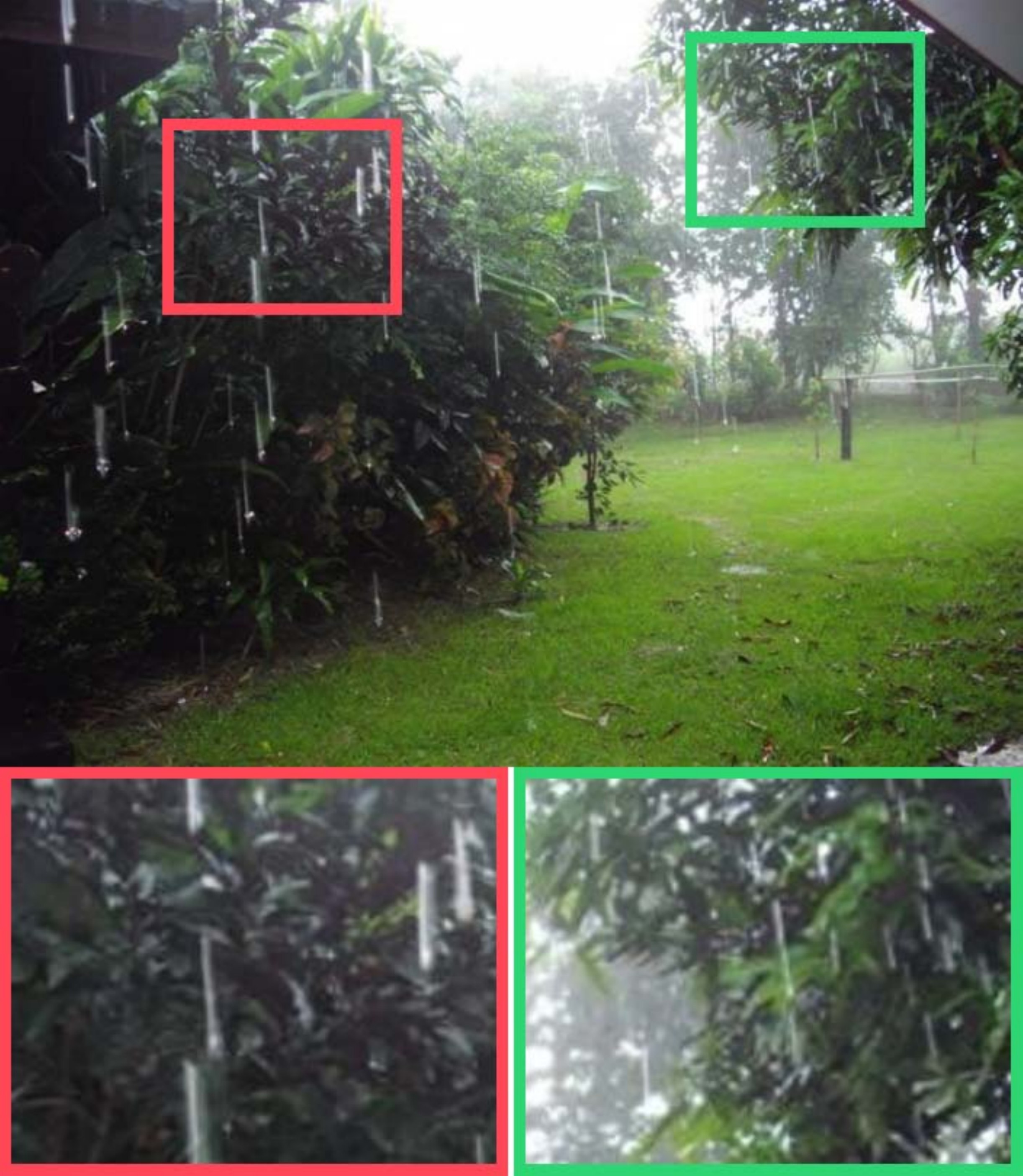}
		&\includegraphics[width=0.15\textwidth, height=0.104\textheight]{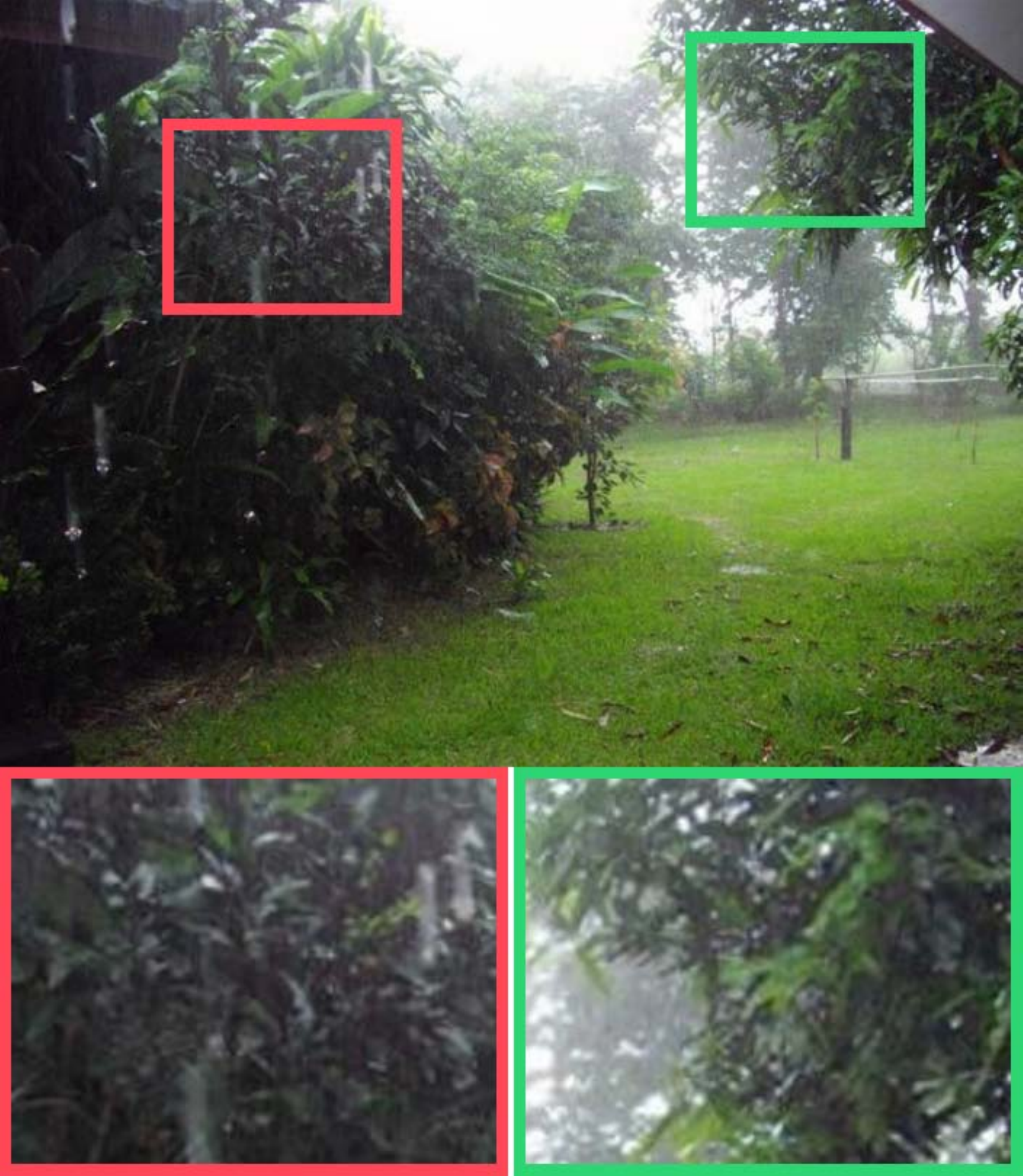}
		&\includegraphics[width=0.15\textwidth, height=0.104\textheight]{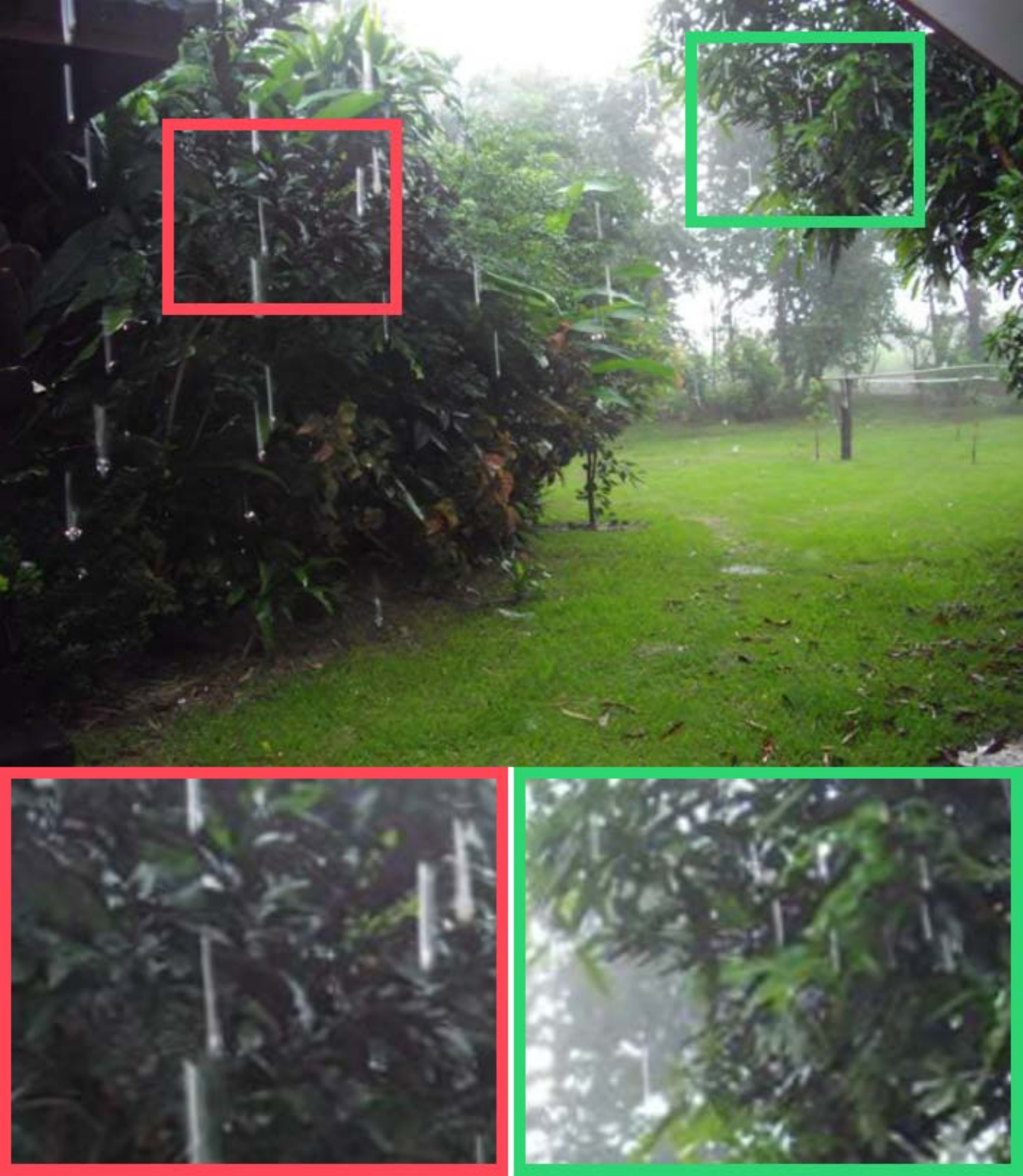}
		&\includegraphics[width=0.15\textwidth, height=0.104\textheight]{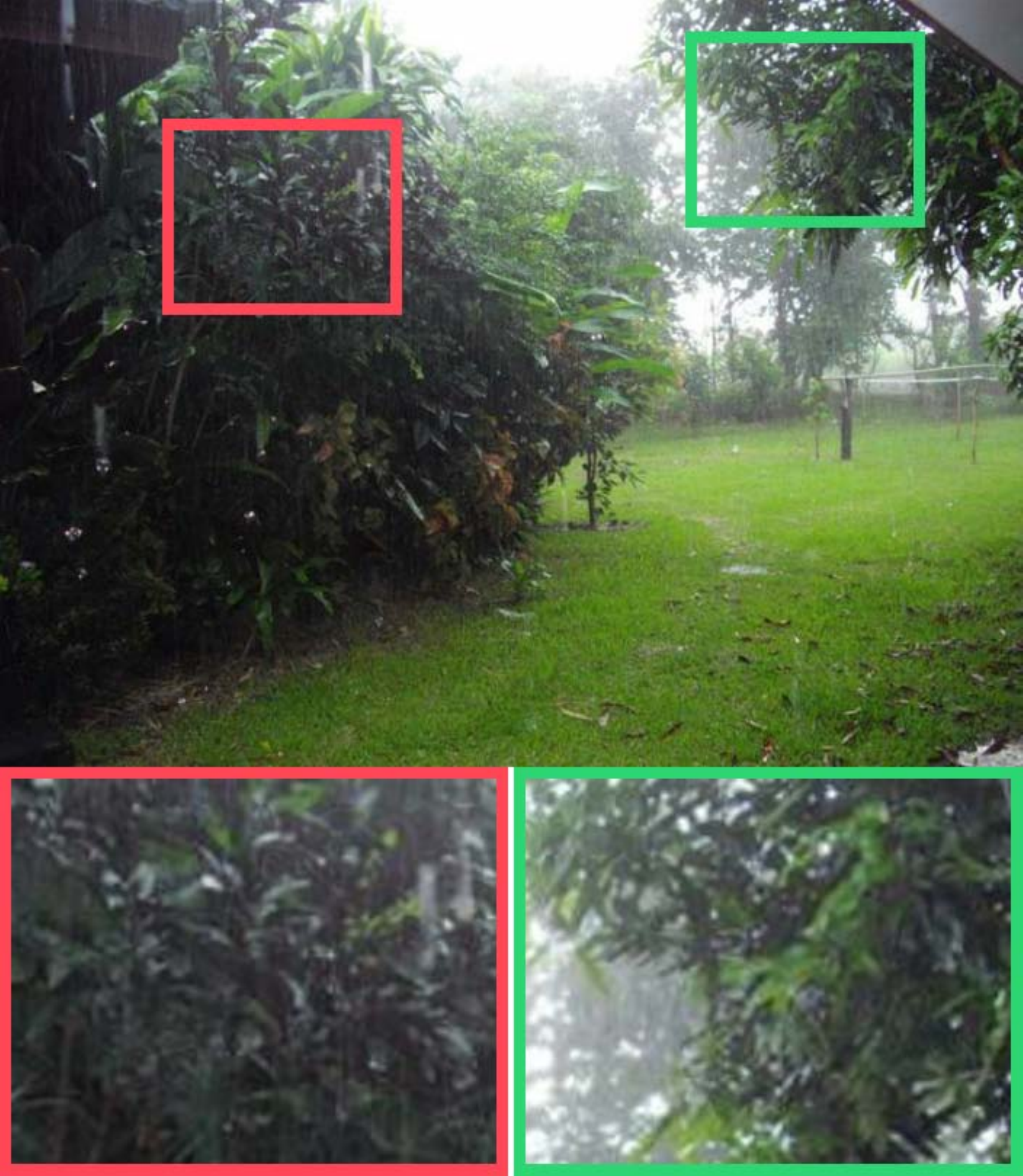}	\\
		3.1154 / 2.1661&3.1762 / 2.1811&3.2183 / 2.2335&3.6224 / 2.5379&3.2893 / 2.2864&3.0312 / 2.1339\\
		SIRR &Syn2Real&MSPFN&DualGCN&MPRNet&Ours\\
	\end{tabular}
	\caption{Subjective comparison of the different deraining methods on real-world data. In the first sample, we can see that our method achieves the best performance on visible rain removal. In the second sample, our method not only removes rain streaks but also preserves details to the greatest extent. The corresponding values on NIQE / PI are reported below. Obviously, our method achieves the best performance on visible rain removal.
}
	\label{fig:Derain_Real}
	\vspace{-0.3cm}
\end{figure*}
\begin{figure}[!tph]
	\centering
	\setlength{\tabcolsep}{1pt}
	\begin{tabular}{cccccccccccc}
		\includegraphics[width=0.5\textwidth, height=0.165\textheight]{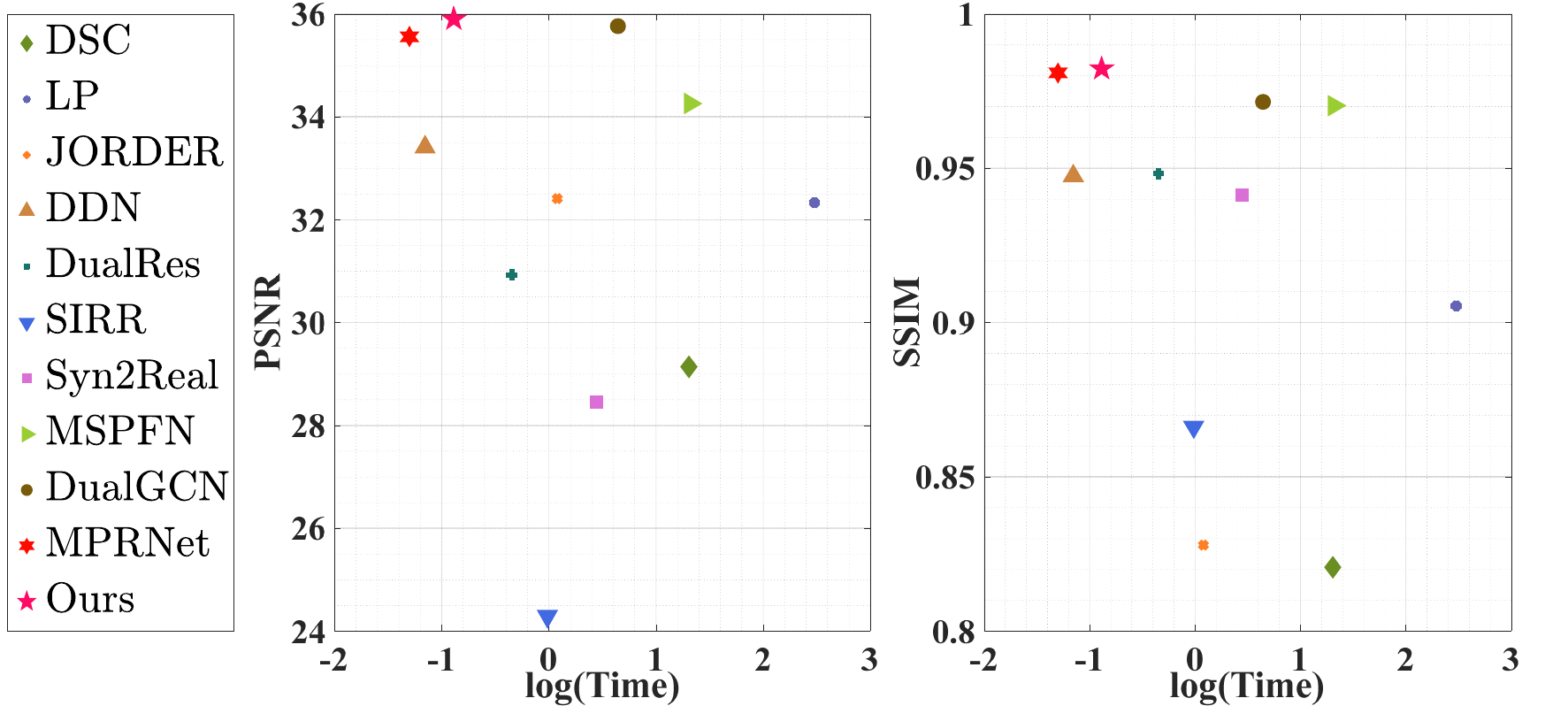}\\
	\end{tabular}
	\caption{ Running time~(after logarithmization) and effectiveness (in terms of PSNR and SSIM) analysis. The proposed method performs better than the others and ranks third in efficiency.}
	\label{fig:time}
\end{figure}
\subsection{Results on Synthetic Images}
\vspace{-.4em}
We apply five synthetic deraining benchmarks, including Rain12~\cite{li2016rain}, Rain100L~\cite{yang2016joint}, Rain1200~\cite{zhang2018density}, Rain1400~\cite{fu2017removing} and Rain100H~\cite{yang2016joint}, in which the first four datasets are degraded with light rain, and the last benchmark is corrupted with heavy rain. For Rain12, the qualitative results are illustrated in Fig.~\ref{fig:syn12com}. We can see that the conventional model-based DSC and LP, and the deep learning-based DualRes, SIRR and MSPFN still retain visible rain streaks. The results of MPRNet tend to introduce additional noise interference. Both DualGCN and our method exhibit an impressive restoration performance.
It is worth mentioning that our method shows a certain degree of information loss in the small clouds and water ripples. This is because the multi-scale feature extraction in our method is able to cover a variety of complicated rain streaks and mistakenly removes the cloud and ripple details as tiny rain streaks.  
As more evidence, the PSNR/SSIM values are reported below each result, where they evaluate the image quality from multiple perspectives, such as contrast and illumination. The higher PSNR and SSIM scores indicate better quality and clearer content. Therefore, even if the result of our method mistakenly polished the minor details, it still obtains the highest metric values. Fig.~\ref{fig:syn1400com} displays the results on Rain1400. Obviously, the conventional methods, DSC and LP, tend to oversmooth the background details where the text information in the red box is blurry. In contrast, our method recovers a clearer and more precise result than other methods.

Visual results on Rain100H are presented in Fig.~\ref{fig:synHcom}. The input image of this dataset suffers from heavy and bright linear obstructions. For the dictionary and sparse coding properties, the conventional methods are weak in characterizing these sharp and heavy patterns, leaving large amounts of visible rain in the restored results. Our method achieves remarkable improvements over these state-of-the-art methods. It can preserve more realistic and credible image details while effectively removing the interfered rain streaks. In particular, the texture on the sleeve has been maintained successfully by our method.

\begin{table*}[h]
	\centering
	\caption{The FLOPs and parameters of the deep learning-based image deraining methods.}
	\begin{tabular}{>{\centering}p{1.5cm}|>{\centering}p{1.2cm}|>{\centering}p{1.2cm}|>{\centering}p{1.2cm}|>{\centering}p{1.2cm}|>{\centering}p{1.2cm}|>{\centering}p{1.2cm}|>{\centering}p{1.2cm}|>{\centering}p{1.2cm}|>{\centering}p{1.2cm}  }
		\hline
		Methods&\tabincell{c}{JORDER\tabularnewline\cite{yang2016joint}}&\tabincell{c}{DDN\tabularnewline\cite{fu2017removing}}&\tabincell{c}{DualRes\tabularnewline\cite{liu2019dual}}&\tabincell{c}{SIRR\tabularnewline\cite{wei2019semi}} &\tabincell{c}{Syn2Real\tabularnewline\cite{yasarla2020syn2real}}&\tabincell{c}{MSPFN\tabularnewline\cite{jiang2020multi}}&\tabincell{c}{DualGCN\tabularnewline\cite{fu2021rain}}&\tabincell{c}{MPRNet\tabularnewline\cite{zamir2021multi}}&Ours\tabularnewline \hline
		FLOPs~(M)&273,221&0.28&242,821&0.23&20,924&3,174,564&5&141,453&3,021\tabularnewline
		Params~(K)&4,169&60&3,729&57&2,605&15,823&2,731&3,637&249\tabularnewline\hline
	\end{tabular}
	\label{tab:flops}
\end{table*}
\begin{figure*}[htb]
	\centering
	\setlength{\tabcolsep}{1pt}
	\begin{tabular}{cccccccccccc}
		
		\includegraphics[width=0.18\textwidth,height=0.085\textheight]{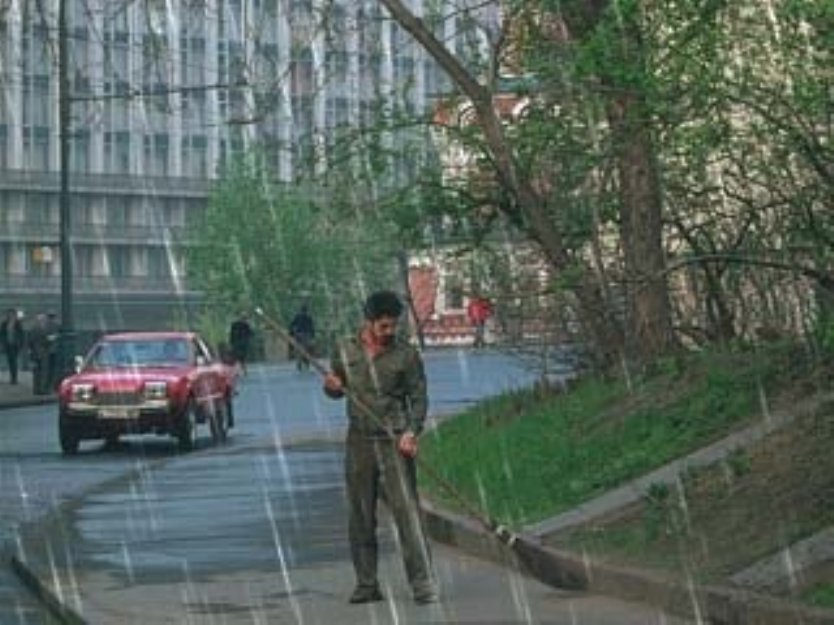}
		&\includegraphics[width=0.18\textwidth,height=0.085\textheight]{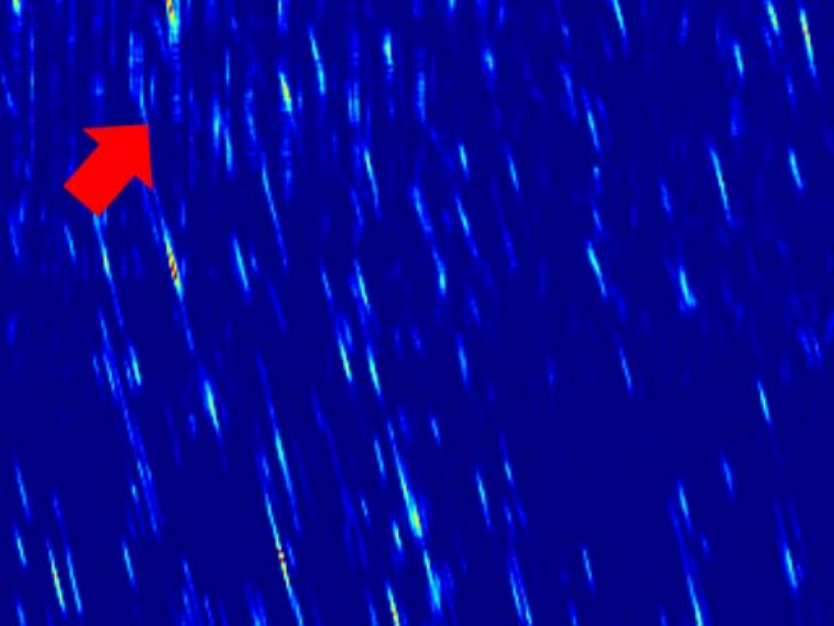}
		&\includegraphics[width=0.18\textwidth,height=0.085\textheight]{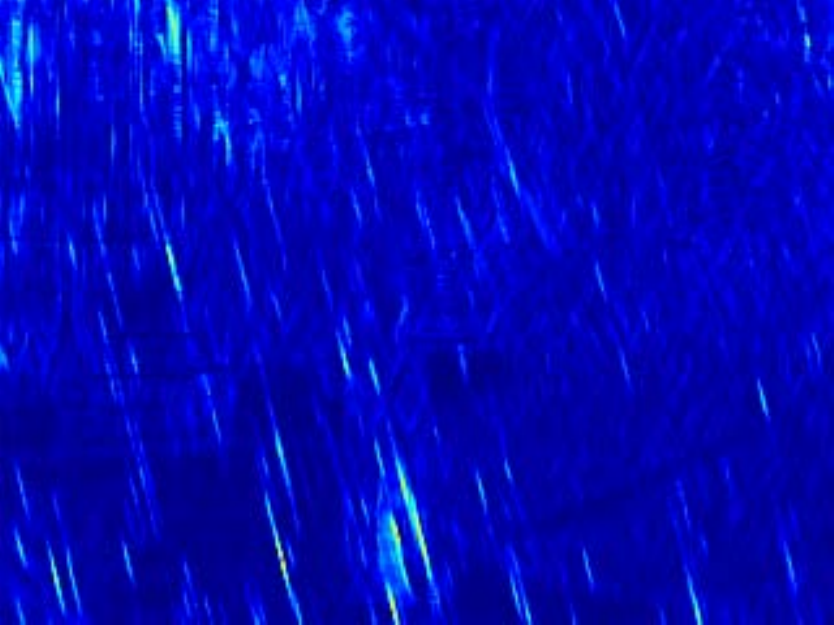}
		&\includegraphics[width=0.18\textwidth,height=0.085\textheight]{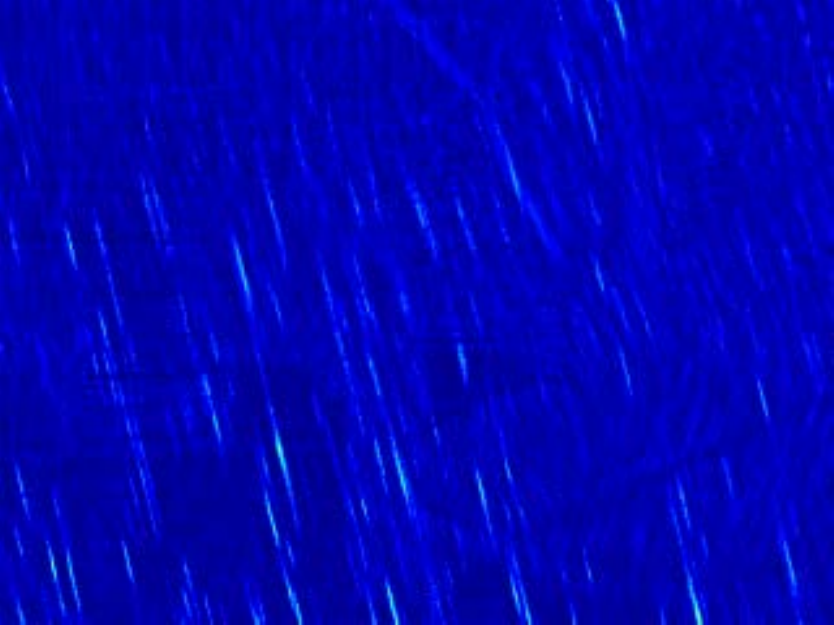}
		&\includegraphics[width=0.18\textwidth,height=0.085\textheight]{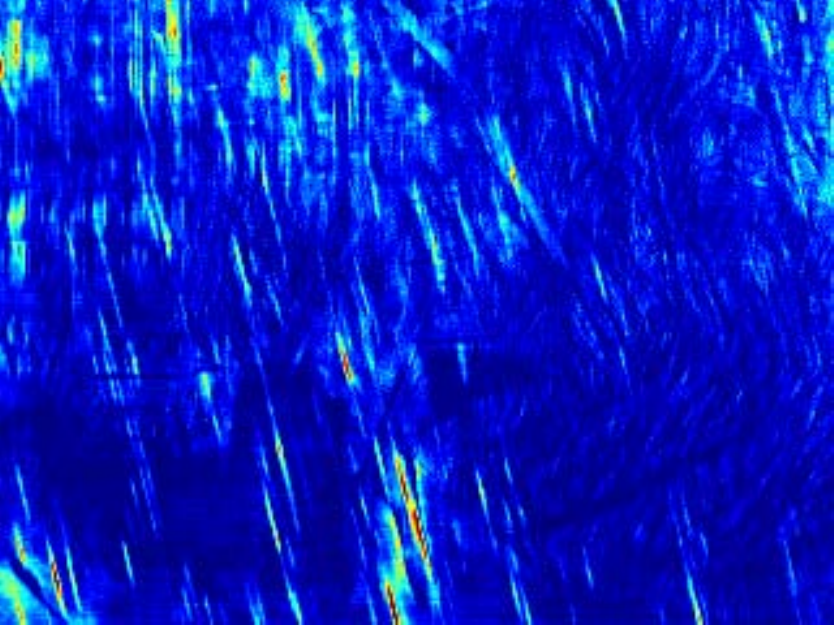}\\Input&JORDER&DDN&DualRes&SIRR\\
		\includegraphics[width=0.18\textwidth,height=0.085\textheight]{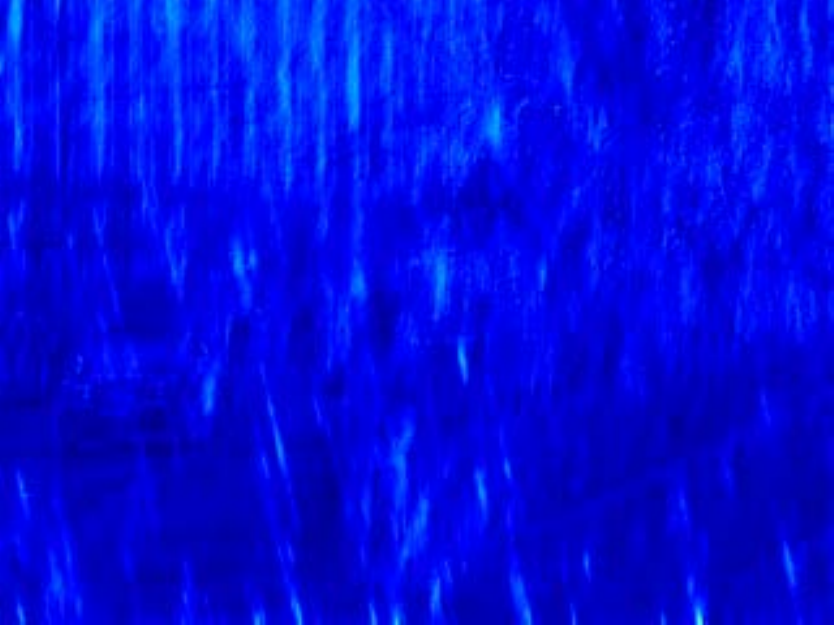}
		&\includegraphics[width=0.18\textwidth,height=0.085\textheight]{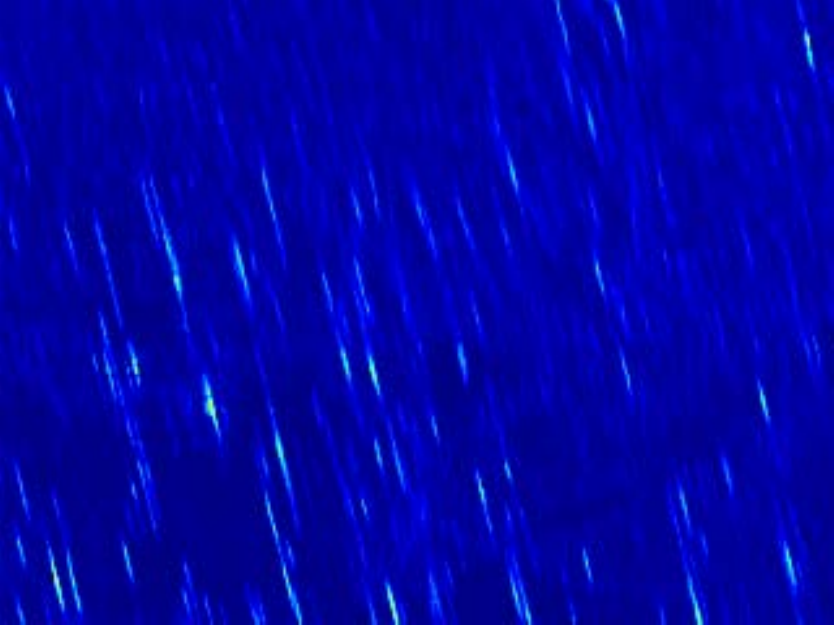}
		&\includegraphics[width=0.18\textwidth,height=0.085\textheight]{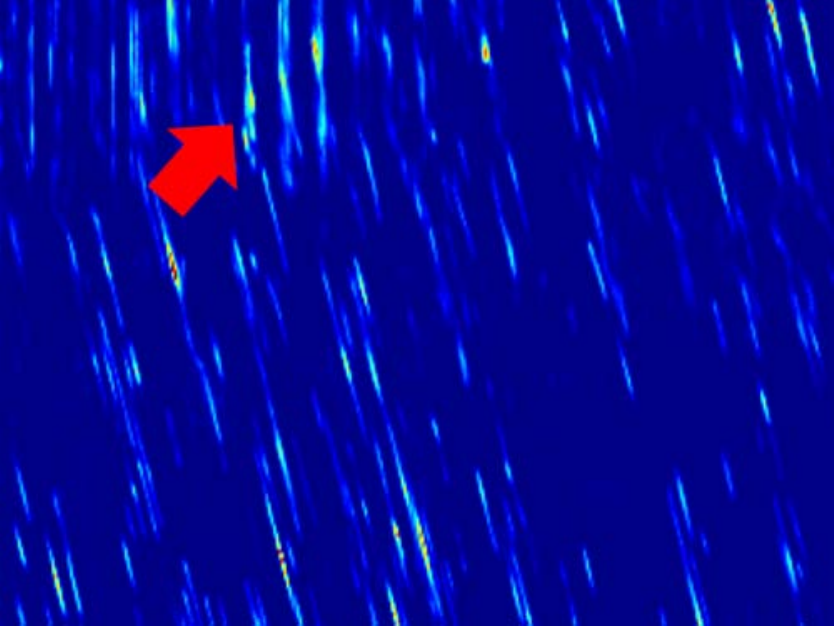}
		&\includegraphics[width=0.18\textwidth,height=0.085\textheight]{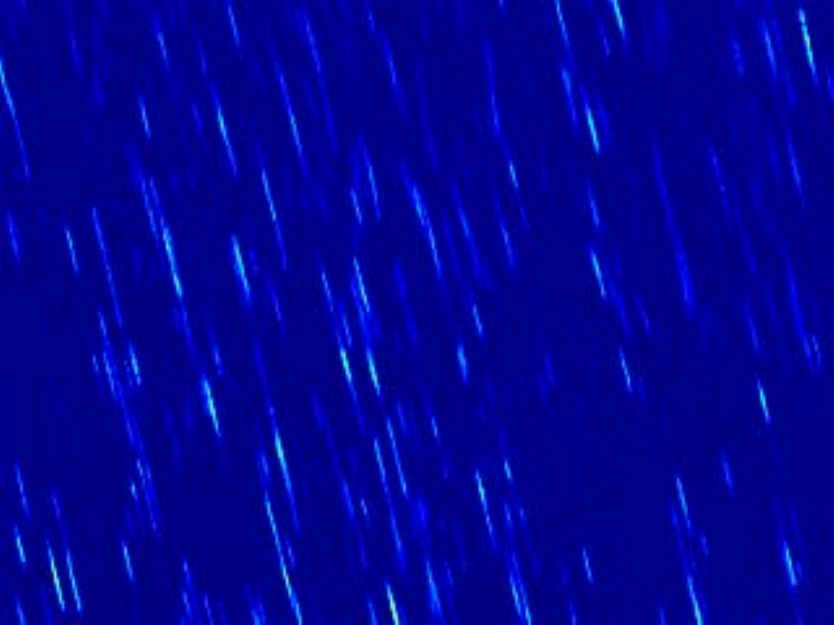}
		&\includegraphics[width=0.18\textwidth,height=0.085\textheight]{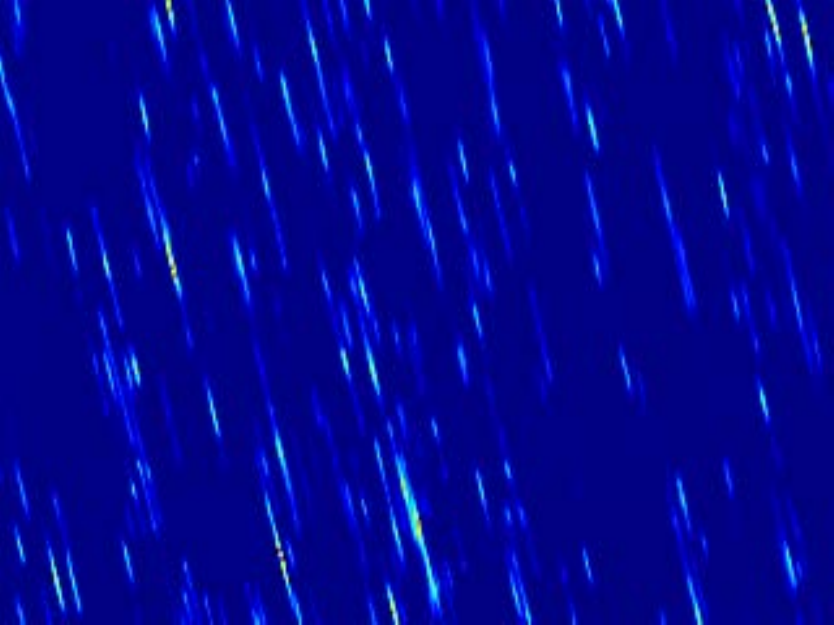}\\Syn2Real&MSPFN&DualGCN&MPRNet&Ours\\
		
	\end{tabular}
	\caption{Illustration of rain extractions. Compared with JORDER and DualGCN, our method shows superiority in distinguishing rain from the background details, while JORDER and DualGCN classify the background context as undesirable rain in the area indicated by the red arrow. The others show a limited effect on the rain removal, retaining a large amount of visible rain in the background.}
	\label{fig:Derain_rain}
\end{figure*}
\begin{figure*}[htb]
	\centering
	\setlength{\tabcolsep}{1.0pt}
	\begin{tabular}{cccccccccccc}
		
		\includegraphics[width=0.18\textwidth,height=0.131\textheight]{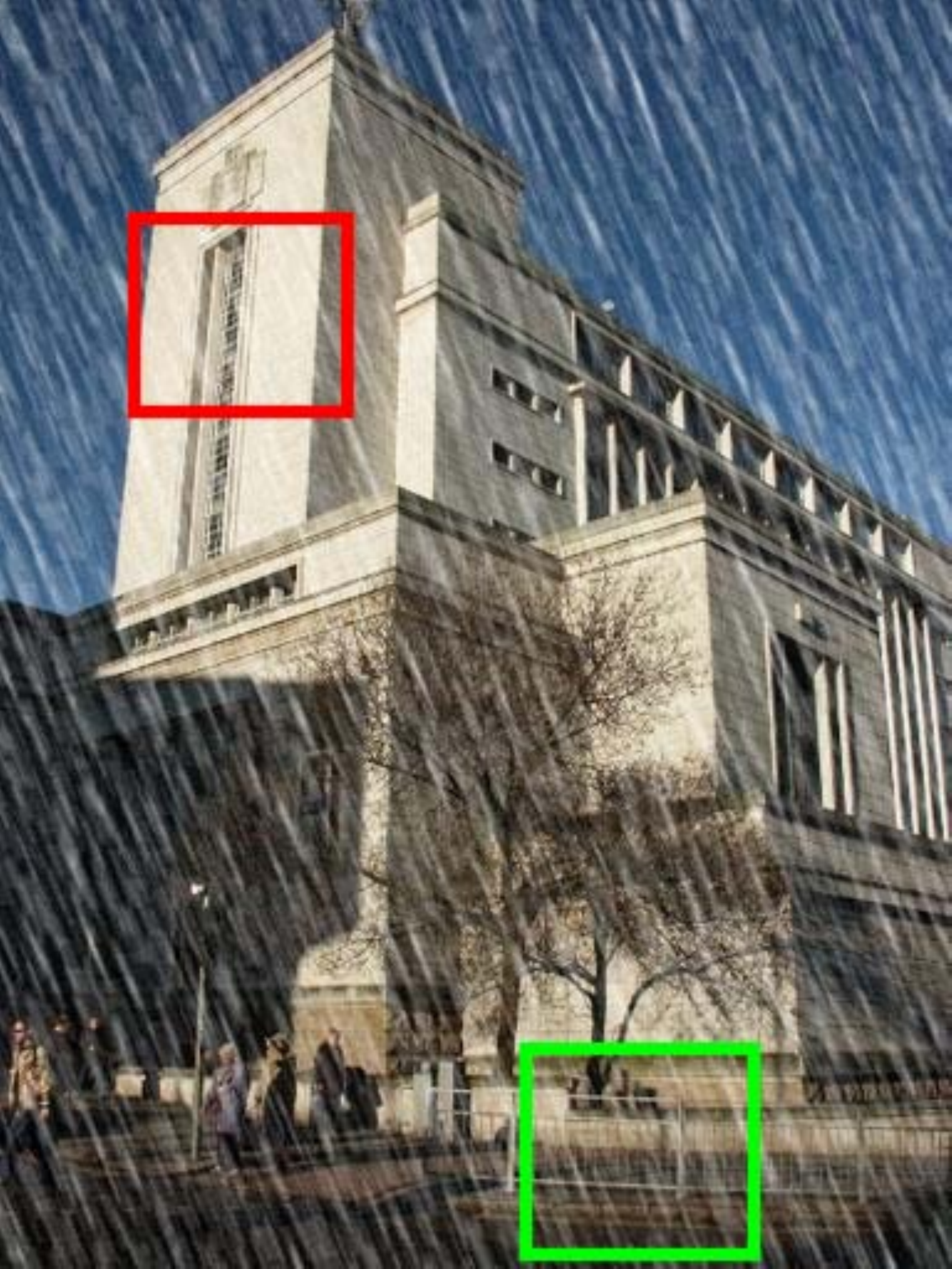}
		&\includegraphics[width=0.18\textwidth,height=0.131\textheight]{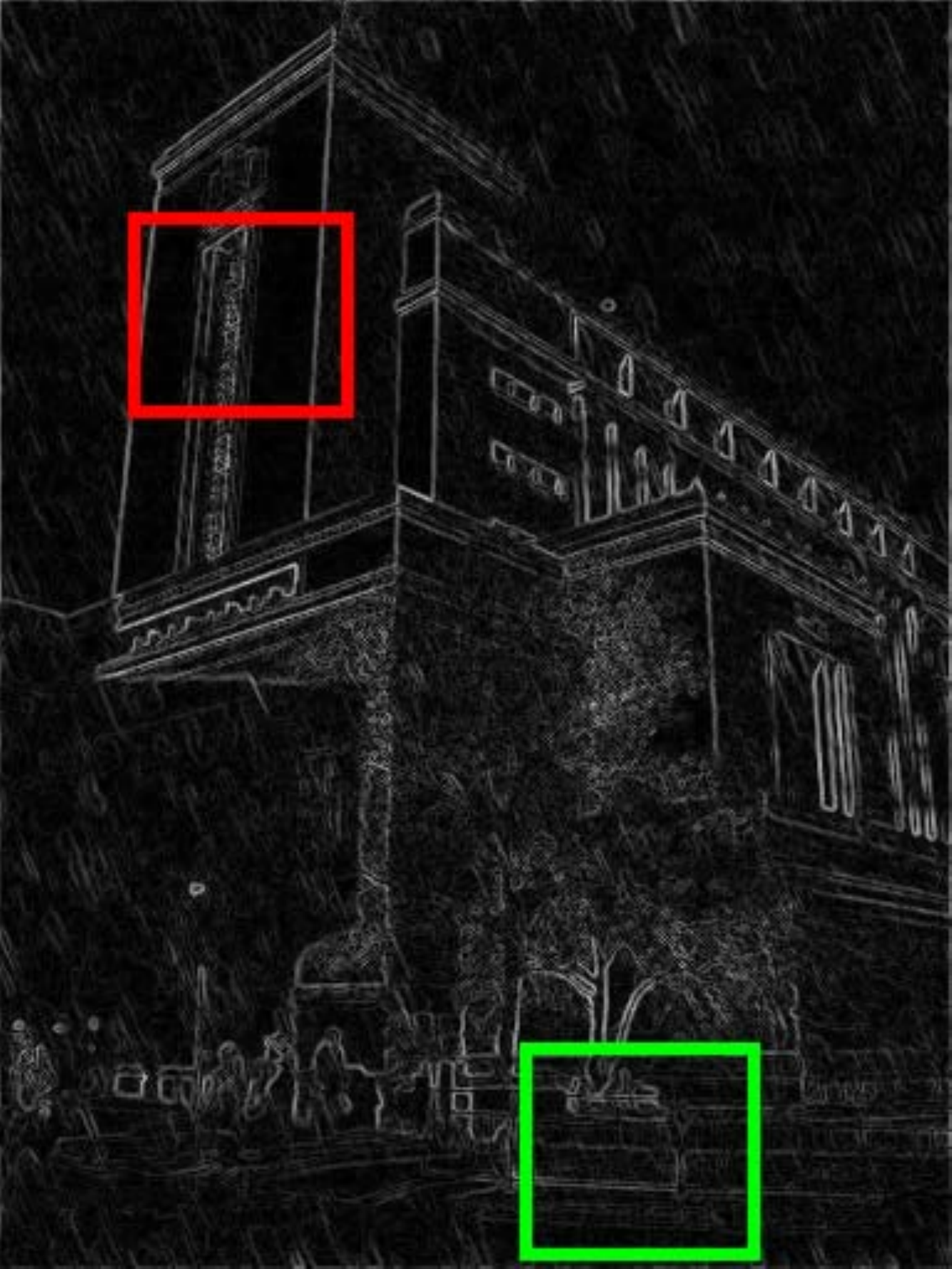}
		&\includegraphics[width=0.18\textwidth,height=0.131\textheight]{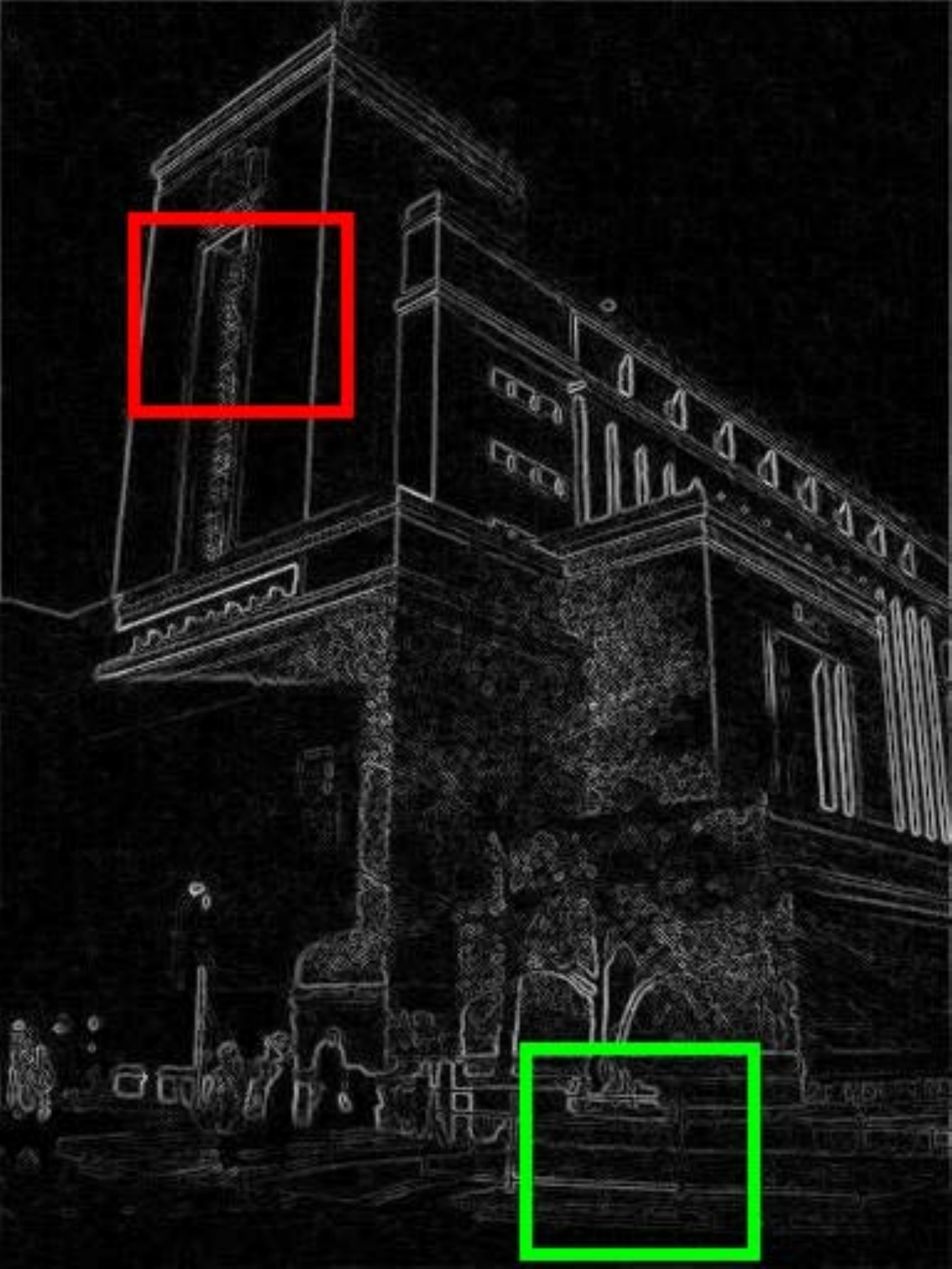}
		&\includegraphics[width=0.18\textwidth,height=0.131\textheight]{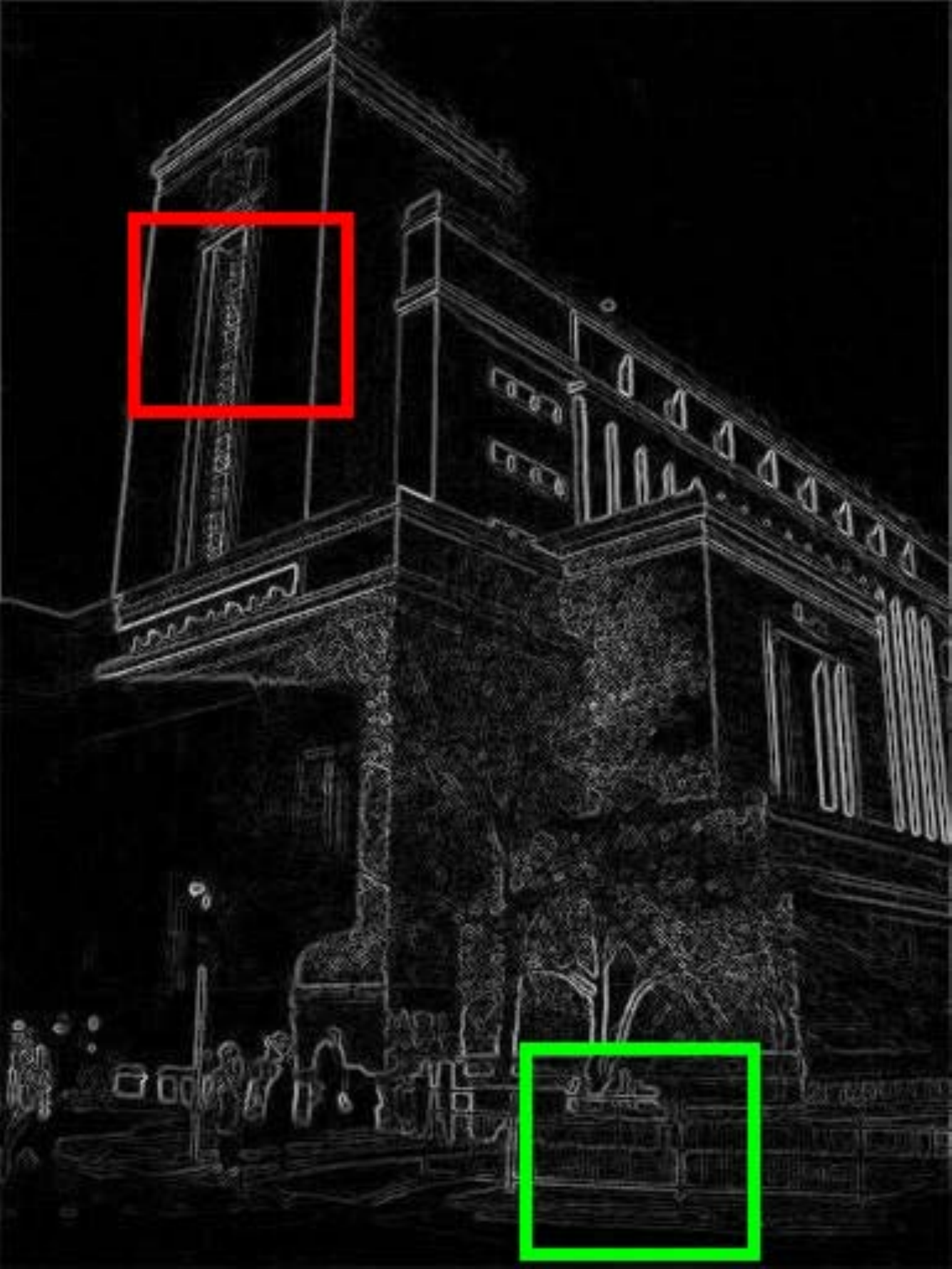}
		&\includegraphics[width=0.18\textwidth,height=0.131\textheight]{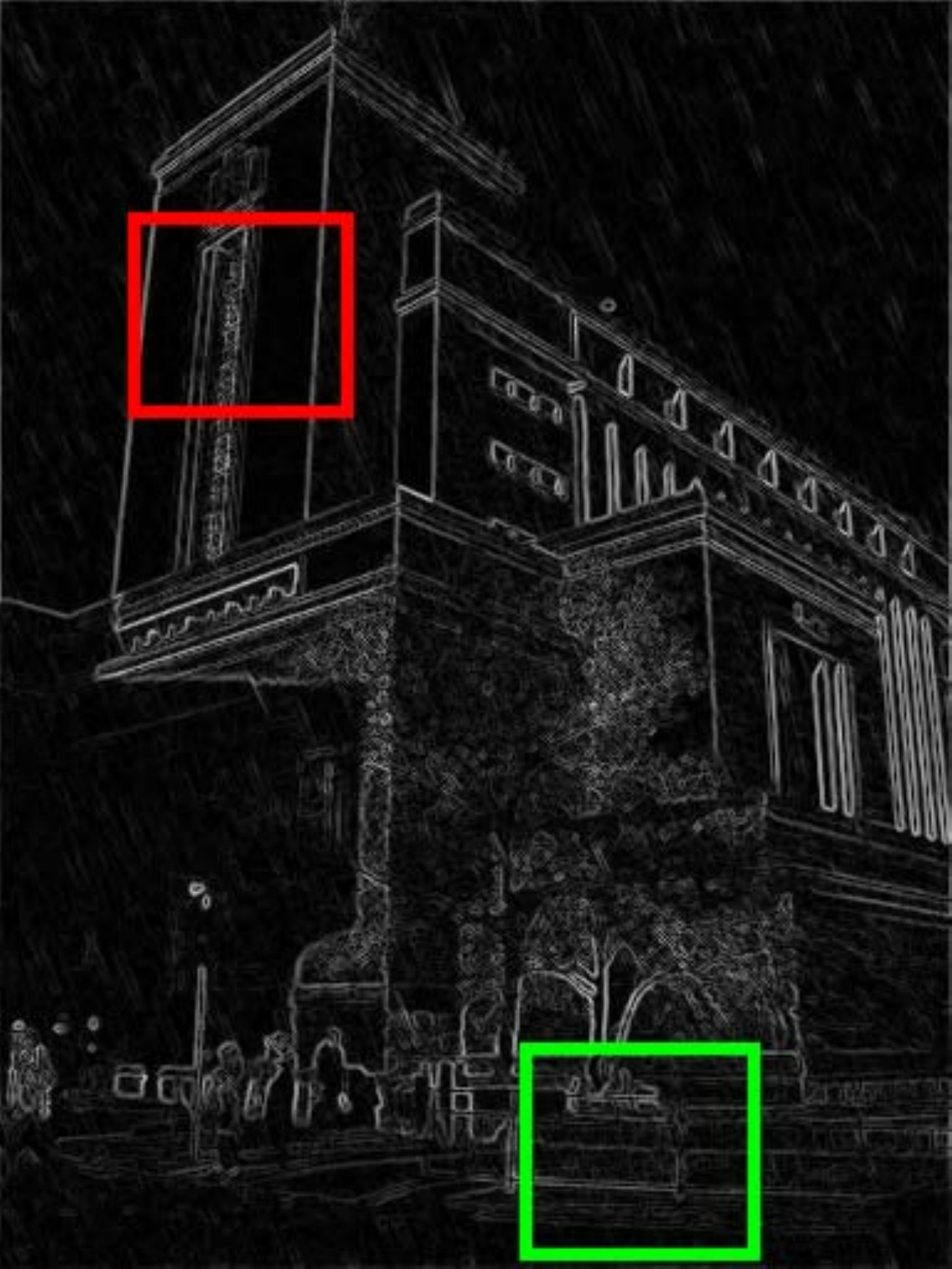}\\Input&JORDER&DDN&DualRes&SIRR\\
		\includegraphics[width=0.18\textwidth,height=0.131\textheight]{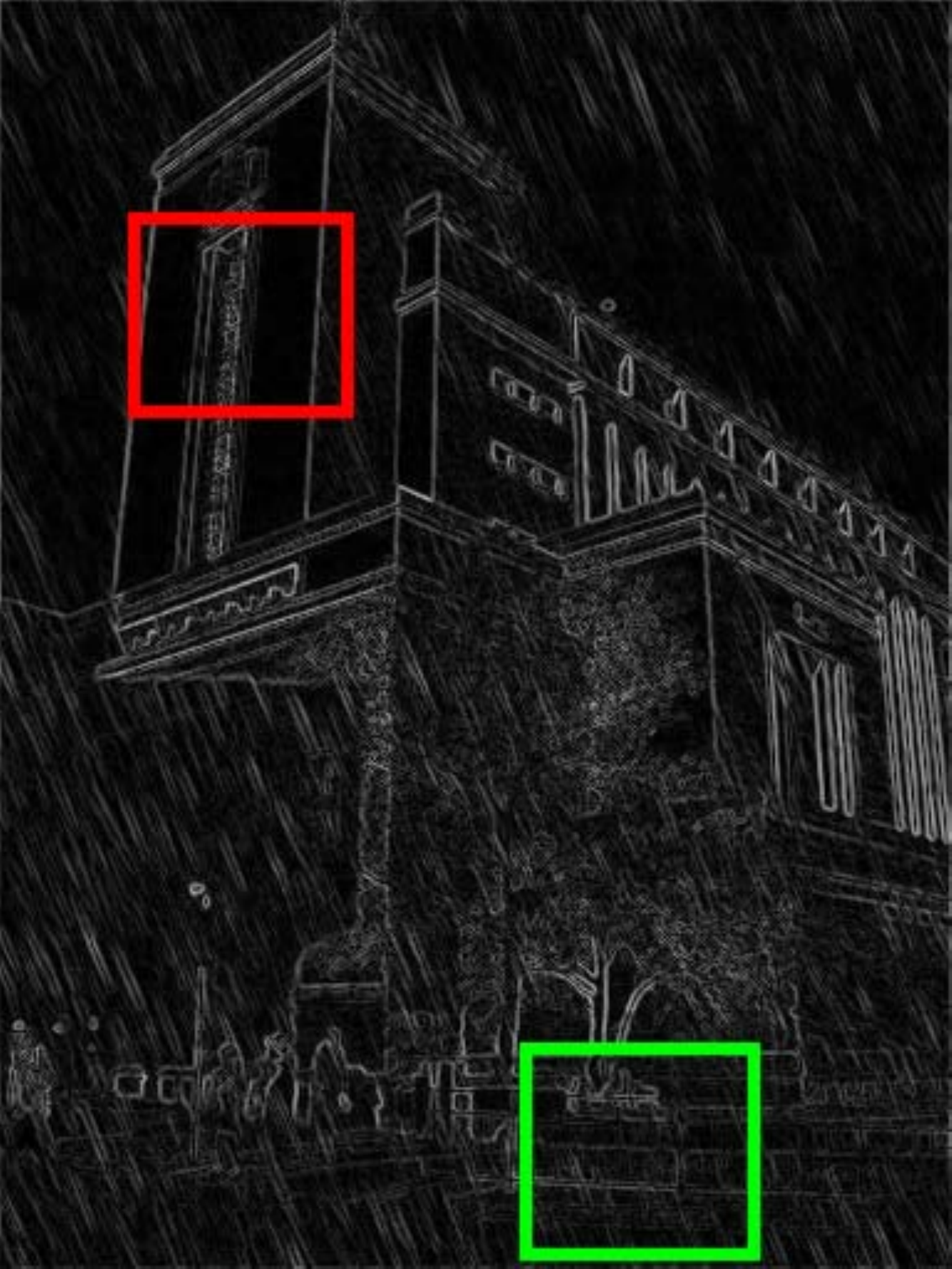}
		&\includegraphics[width=0.18\textwidth,height=0.131\textheight]{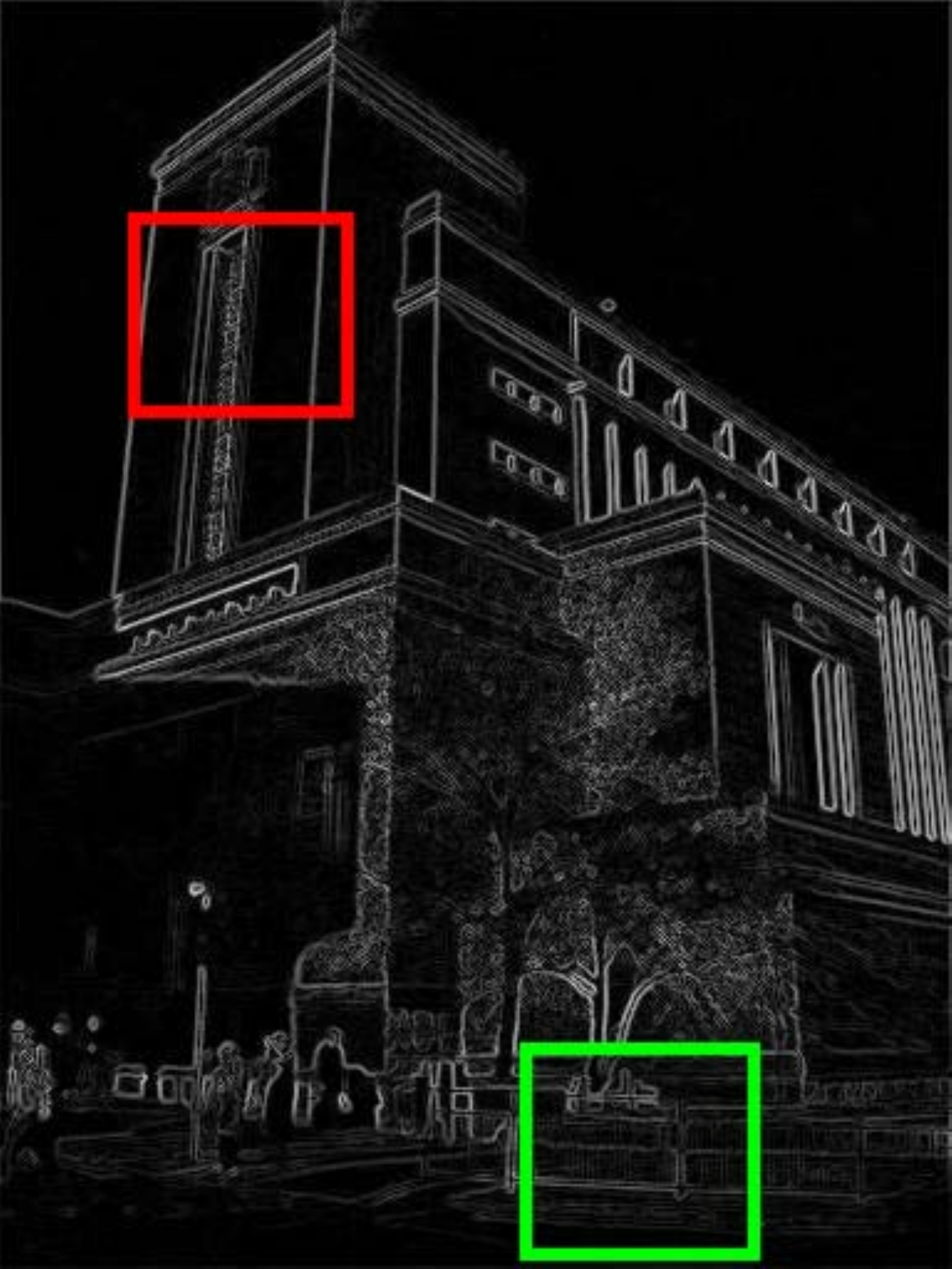}
		&\includegraphics[width=0.18\textwidth,height=0.131\textheight]{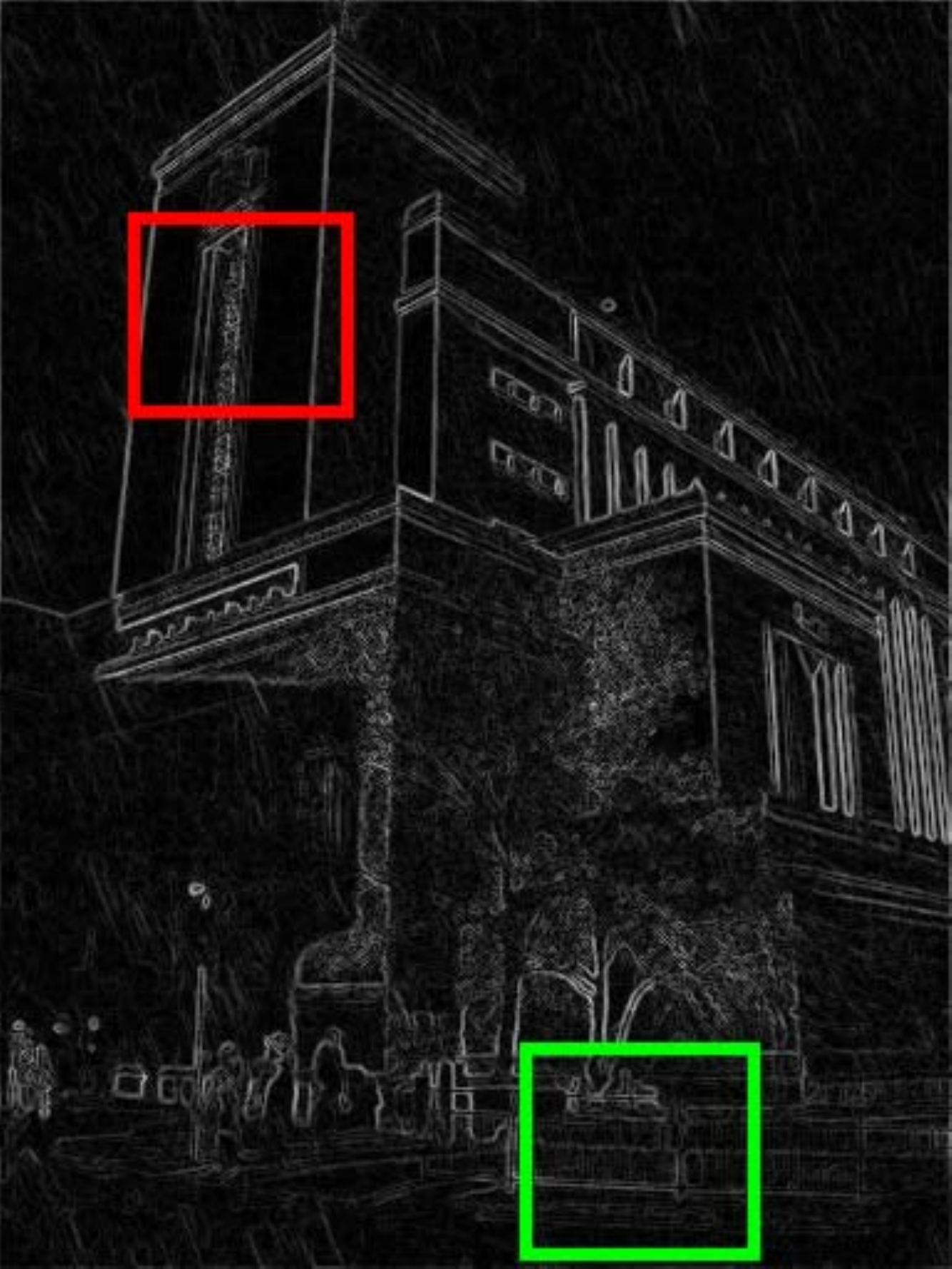}
		&\includegraphics[width=0.18\textwidth,height=0.131\textheight]{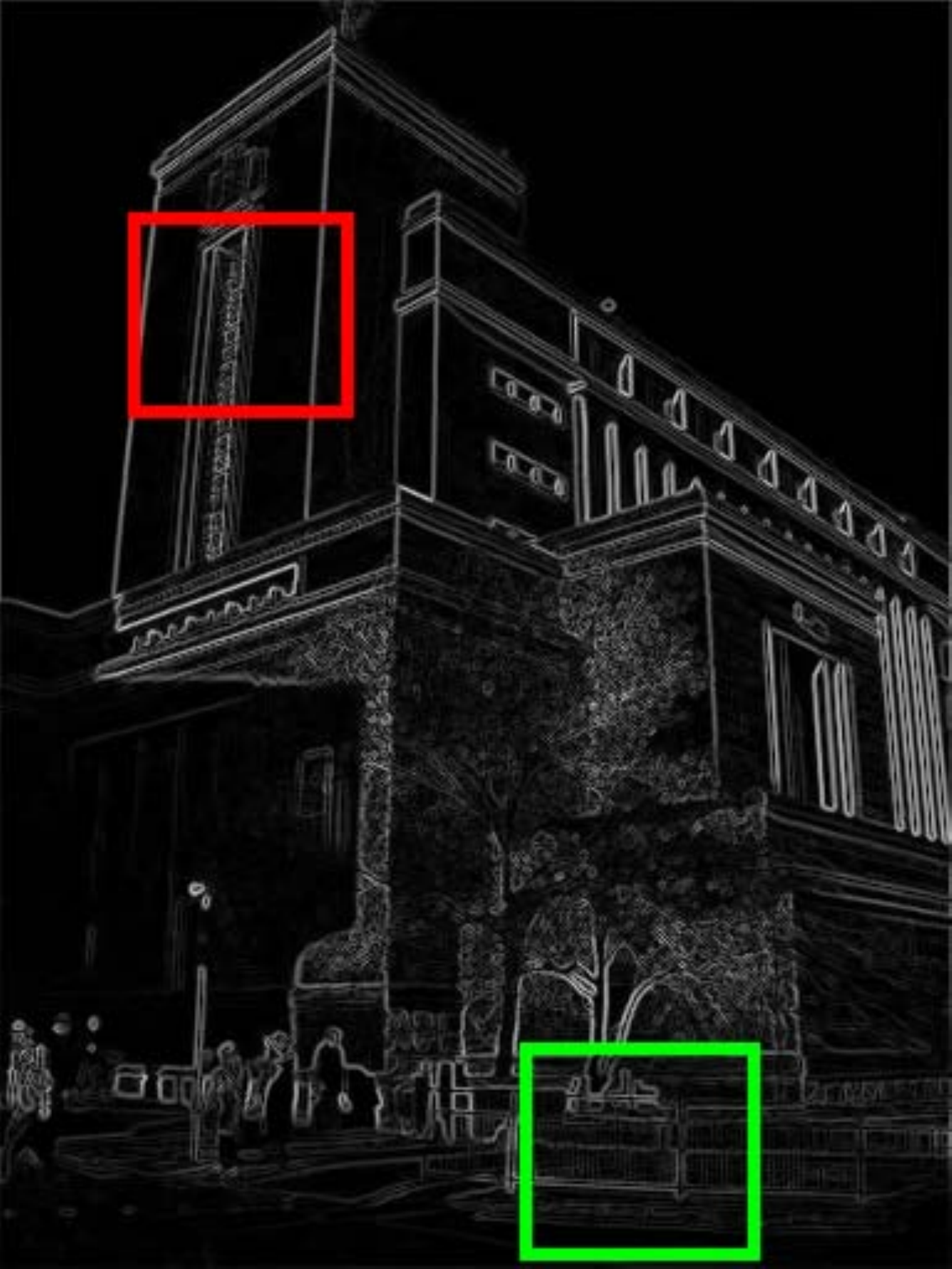}
		&\includegraphics[width=0.18\textwidth,height=0.131\textheight]{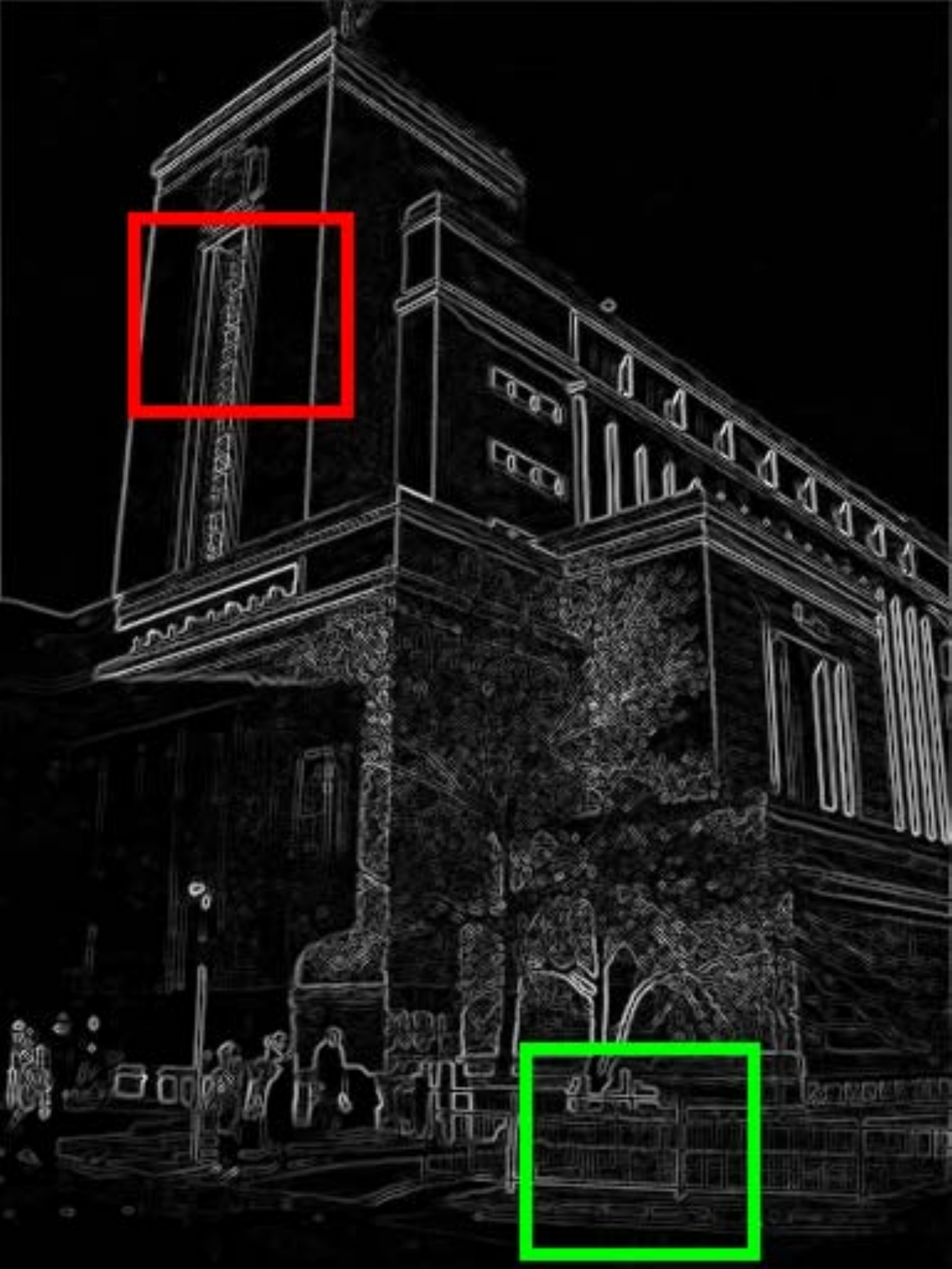}\\Syn2Real&MSPFN&DualGCN&MPRNet&Ours\\
	\end{tabular}
	\caption{Illustration of the detail preservation. Our method shows superiority in detail preservation, especially in the regions boxed in red and green. }
	\label{fig:Derain_details}
\end{figure*}
\begin{figure}[!htb]
	\centering
	\setlength{\tabcolsep}{1pt}
	\begin{tabular}{cccccccccccc}
		\includegraphics[width=0.49\textwidth,height=0.13\textheight]{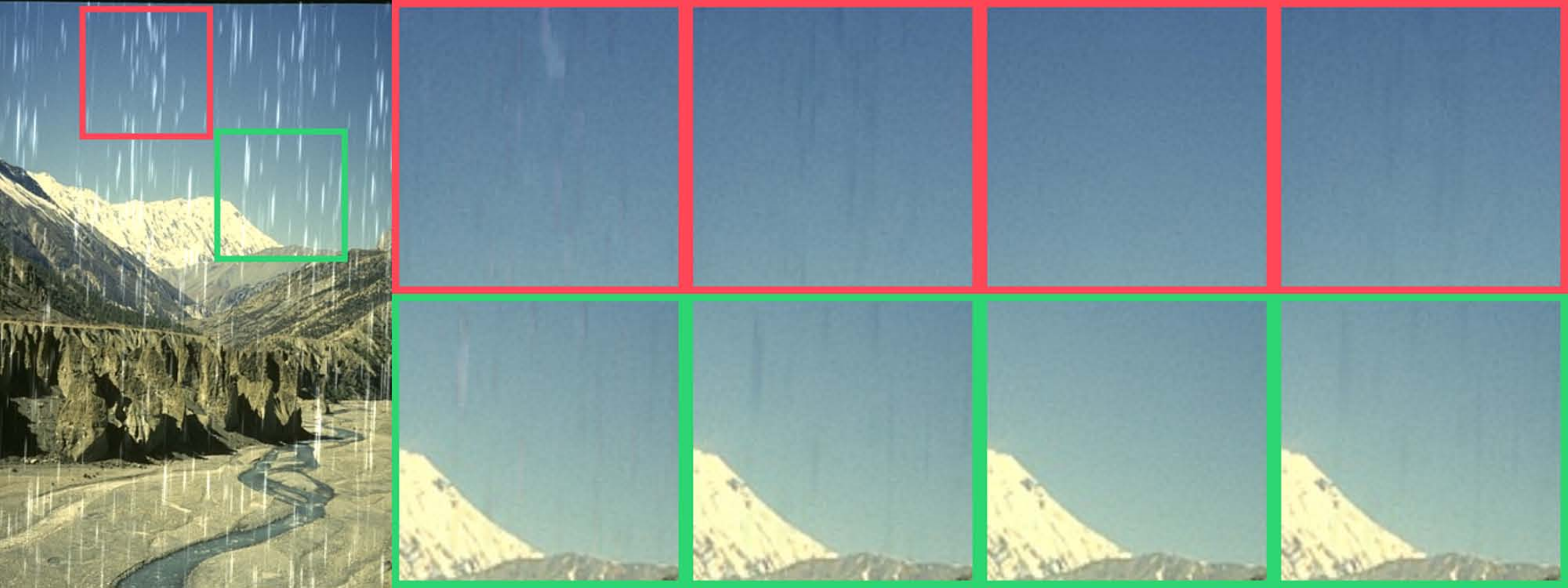}\\
		\includegraphics[width=0.49\textwidth,height=0.13\textheight]{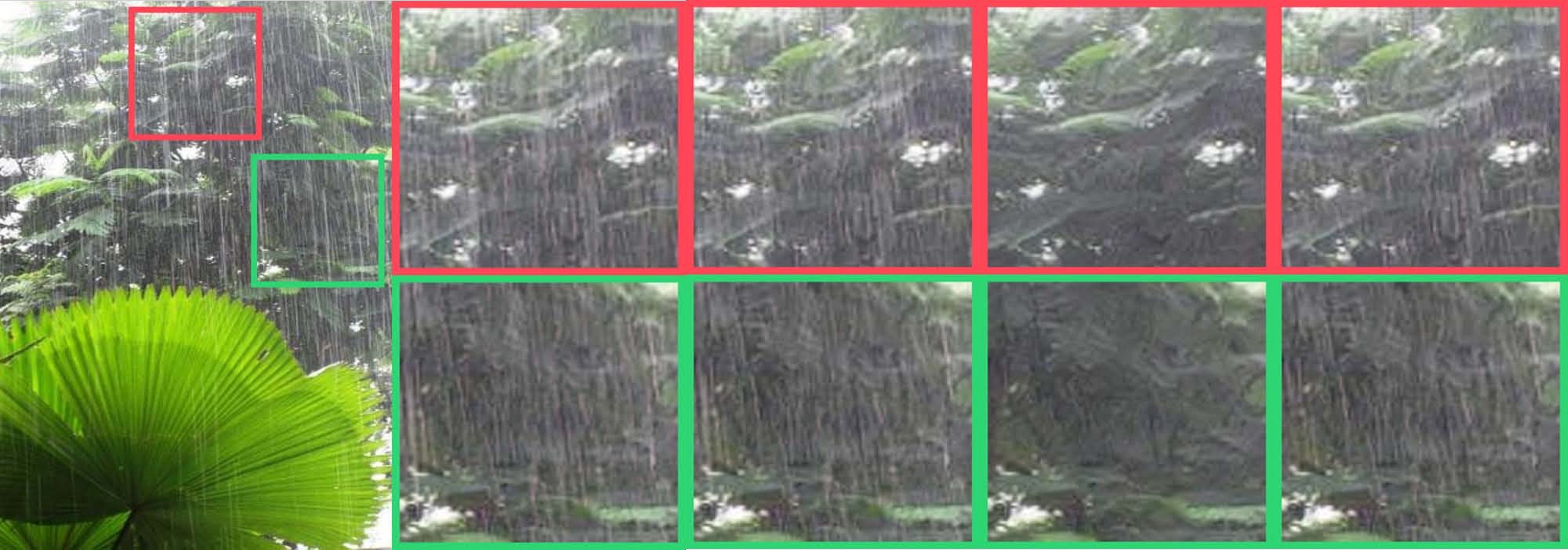}\\
		\includegraphics[width=0.49\textwidth,height=0.13\textheight]{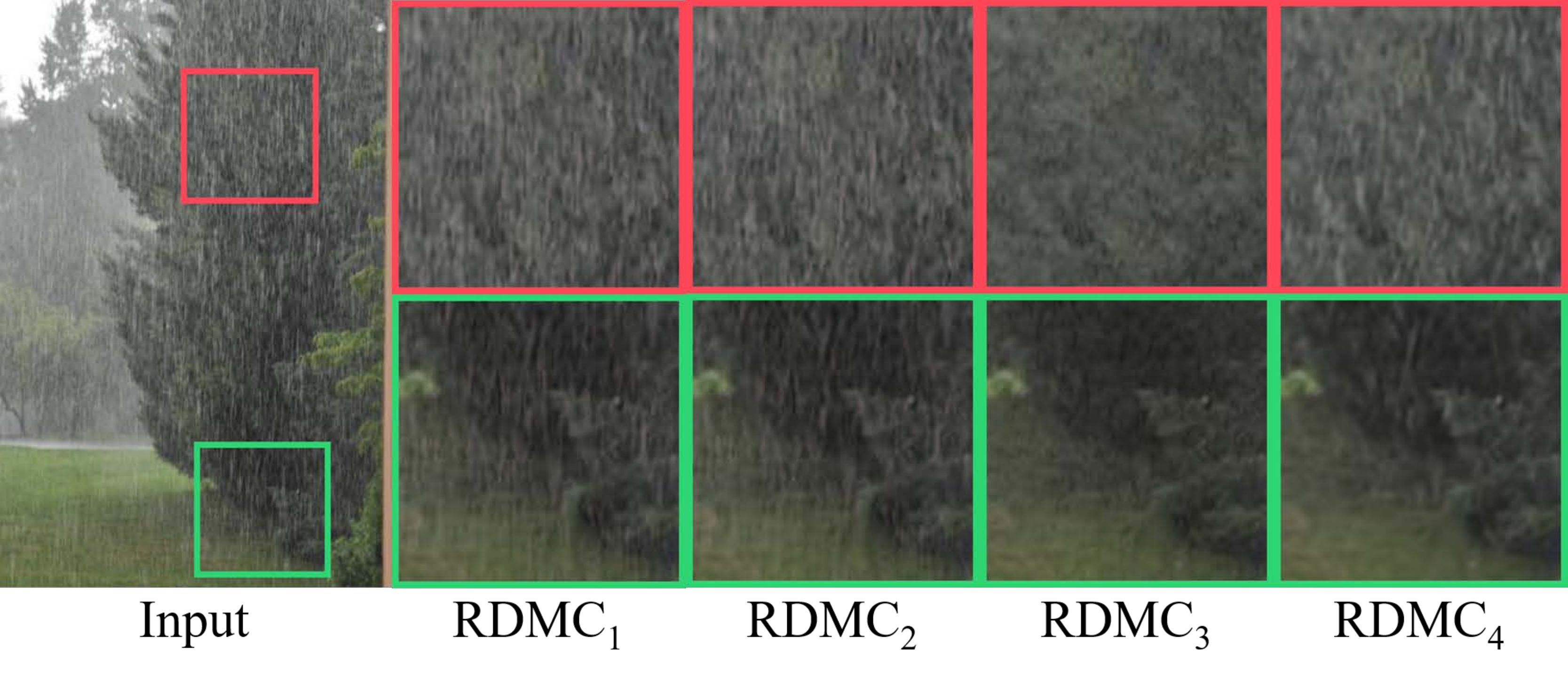}\\
	\end{tabular}
	\caption{Analysis of recursive unfolding factors. The results show that the effectiveness is gradually improved along with the recursion in the first three stages but the results present little improvement after the third stage, and there is even a slight decrease on the second sample.}
	\label{fig:Derain_iteration}
	%	\vspace{-0.5cm}
\end{figure}

The quantitative results are reported in Table.~\ref{tab:Derain}. On the Rain12 and Rain1400 datasets, the proposed method achieves the best results in terms of PSNR and SSIM among the ten competitive methods. On Rain1200, our method ranks first in PSNR, surpassing the second method~(i.e., JORDER) by 2.3dB, while in SSIM, we rank second by a narrow margin of 0.0078dB. DualGCN and MPRNet rank first on Rain100L and Rain100H, respectively. Meanwhile, we rank second on these two datasets. It is worth mentioning that the training data of MPRNet are gathered from a series of existing synthetic rainy data, where a large portion comes from the same source as Rain100H~\cite{yang2016joint}, resulting in a strong simulation of this dataset and generating excellent quantitative results. Considering all the results on the five testing datasets, our RDMC achieves the top two results on each dataset, showing relatively stable performances under diverse conditions over the other methods, without an extremely variant performance.

\subsection{Results on Real-world Images}

We conduct additional comparisons on real-world data~\cite{fu2017removing,yang2016joint,zhang2019image,wei2019semi} to further demonstrate the generalization and robustness of the proposed method. The visual results are illustrated in Fig.~\ref{fig:Derain_Real}. Obviously, in the first case, DSC, LP, MSPFN and DualGCN fail to remove the undesirable streaks. JORDER, DDN and SIRR tend to introduce additional artifacts that degrade the images further. In addition, most methods leave a large amount of intensive rain in the second case; only JORDER and DualGCN perform a distinct removal of visible rain, but they still retain tiny rain amounts in the results.
In contrast, the proposed method achieves promising results, which removes the majority of the visible rain and recovers the results with more credible content.
The quantitative comparison is reported in Table.~\ref{tab:real}. NIQE and PI are employed as evaluation metrics, and the image quality is inversely proportional to the numerical results. We observe that the proposed method ranks first on both the NIQE and PI metrics among the ten competitive methods, followed by SIRR and DualGCN. To this end, the objective results are roughly consistent with subjective ones.

\subsection{Running Time and Complexity}
Fig.~\ref{fig:time} exhibits the running time and performance of the different methods, where the logarithm is enforced on the time variable for a friendlier presentation. All the experiments are conducted on a GeForce RTX 2080 Ti GPU. For a fair comparison, we record the running time of processing one 321*481 image. Our method achieves the best performance with relatively high efficiency. Although our method executes the elemental network recursively, it is a slightly slower than the one-kick methods, where our method takes 0.138s to infer one image, and MPRNet and DDN spend 0.046s and 0.066s, respectively. To compare the complexity with the deep learning-based deraining methods, we report the floating point operations per second~(FLOPs) and the parameters of each model in Table.~\ref{tab:flops}. It reveals that the proposed method performs on par with the state-of-the-art MPRNet reported in Table.~\ref{tab:Derain}, requiring significantly fewer parameters and computations to achieve comparable performance. 
Although DDN and SIRR require the most negligible computation and parameters, they show the weakest rain removal abilities. Therefore, our method tends to yield satisfactory results at relatively low computation costs.

\begin{table}[]
	\begin{center}
		\centering
		\caption{Quantitative comparison of different recursive stage~($T$). We set~$T=1,2,3$ and~$4$. The metric evaluation demonstrates that~$T=3$ is optimal which is consistent with the subjective comparison.}
		\label{tab:ablation_iteration}
		\begin{tabular}{>{\centering}p{1.3cm}|>{\centering}p{1.1cm}>{\centering}p{1.1cm}|>{\centering}p{1.1cm}>{\centering}p{1.1cm}}
			\hline
			{Dataset}&\multicolumn{2}{c|}{Rain100L}&\multicolumn{2}{c}{Real-world}\tabularnewline\hline
			{Measure}&PSNR~$\uparrow$&SSIM~$\uparrow$&NIQE~$\downarrow$&PI~$\downarrow$\tabularnewline\hline
			
			RDMC$_1$&32.4553&0.9710&3.6328&2.3185\tabularnewline
			RDMC$_2$&32.8868&0.9746&3.5061&2.2520\tabularnewline
			RDMC$_3$&\textbf{36.5899}&\textbf{0.9893}&\textbf{3.2414}&\textbf{2.1049}\tabularnewline
			RDMC$_4$&34.1713&0.9818&3.2740&2.2163\tabularnewline\hline		\end{tabular}%
	\end{center}
\end{table}

\subsection{Evaluation of Rain Streaks Removal}
To measure the rain removal capability, we further compare the extracted rain maps among the different methods. In Fig.~\ref{fig:Derain_rain}, we visualize the rain streak layer with pseudocolor mapping.
DDN, DualRes, SIRR and Syn2Real are prone to misunderstand the scenario information as rain streaks, and many semantic details are extracted into the rain maps. MSPFN and MPRNet are good at handling simple and sparse rain but show little effect on inconspicuous rain in complex backgrounds. Although JORDER and DualGCN seem to extract the rain streaks accurately, the structural information on the building indicated by the red arrow is mistakenly regarded as rain. As expected, our method shows superiority in rain extraction on both simple and diverse rain, surpassing the other methods.
\begin{figure}[t]
	\centering
	\setlength{\tabcolsep}{1pt}
	\begin{tabular}{cccccccccccc}
		\includegraphics[width=0.15\textwidth,height=0.085\textheight]{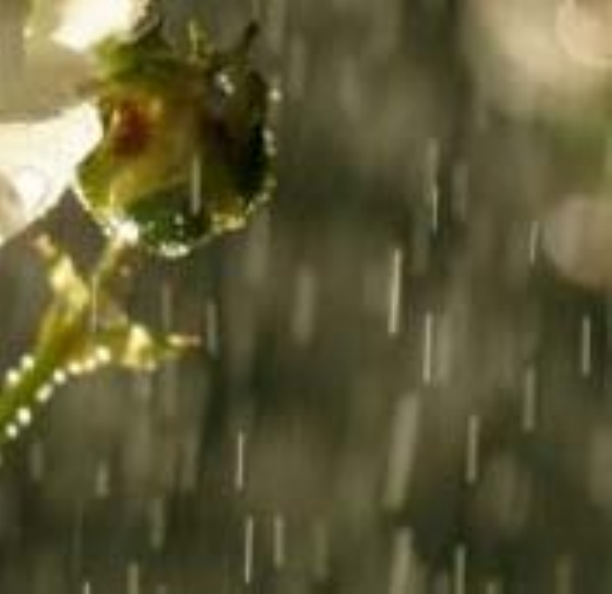}
		&\includegraphics[width=0.15\textwidth,height=0.085\textheight]{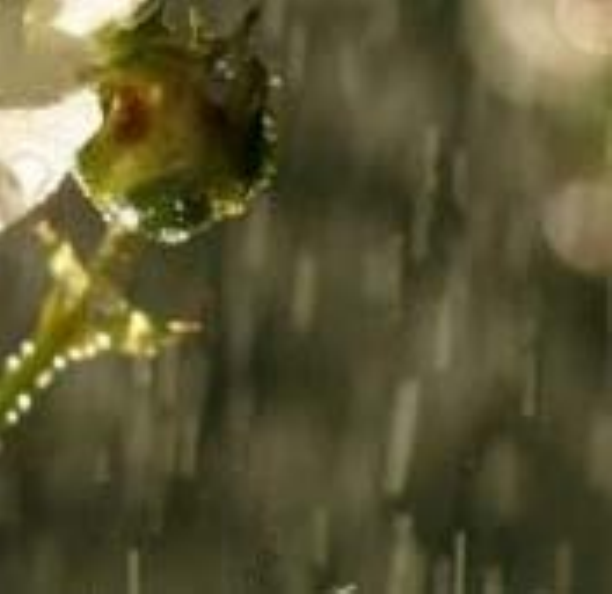}
		&\includegraphics[width=0.15\textwidth,height=0.085\textheight]{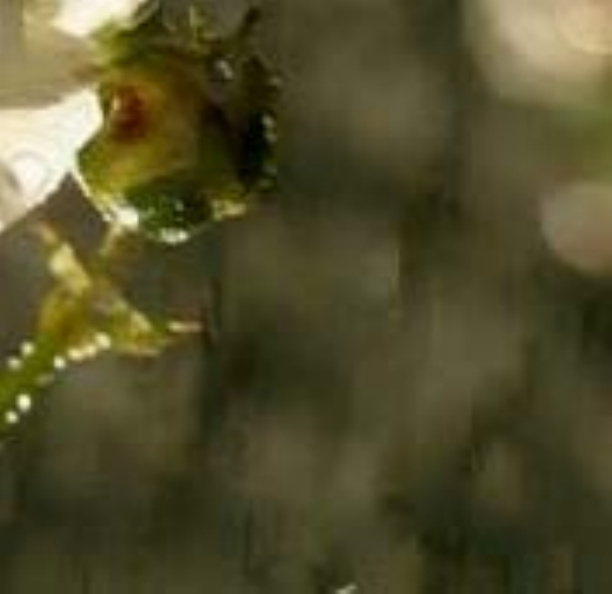}\\
		3.9807 / 3.1064&3.9017 / 3.0572&3.7331 / 3.0297\\
		\includegraphics[width=0.15\textwidth,height=0.085\textheight]{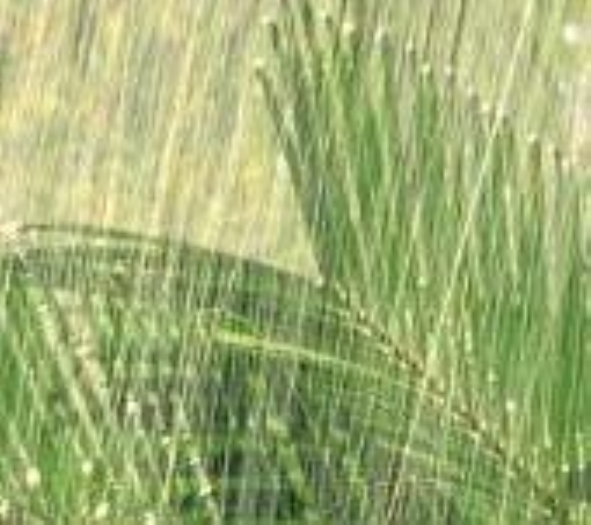}
		&\includegraphics[width=0.15\textwidth,height=0.085\textheight]{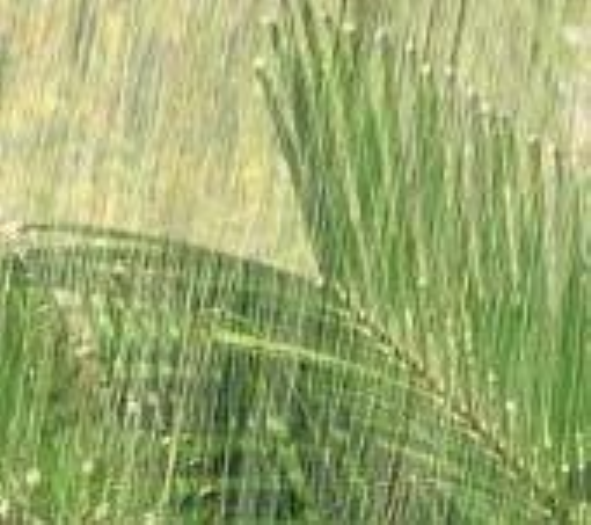}
		&\includegraphics[width=0.15\textwidth,height=0.085\textheight]{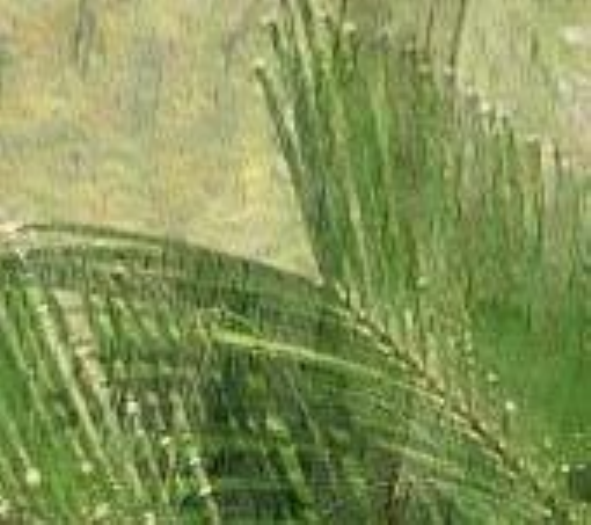}\\
		5.9199 / 3.6029&5.5103 / 3.4002&5.2962 / 3.2734\\
		\includegraphics[width=0.15\textwidth,height=0.085\textheight]{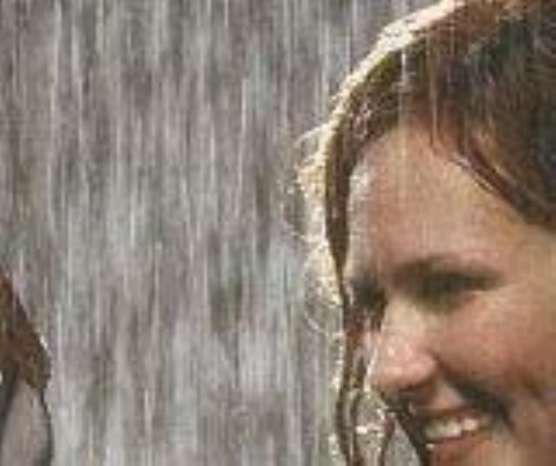}
		&\includegraphics[width=0.15\textwidth,height=0.085\textheight]{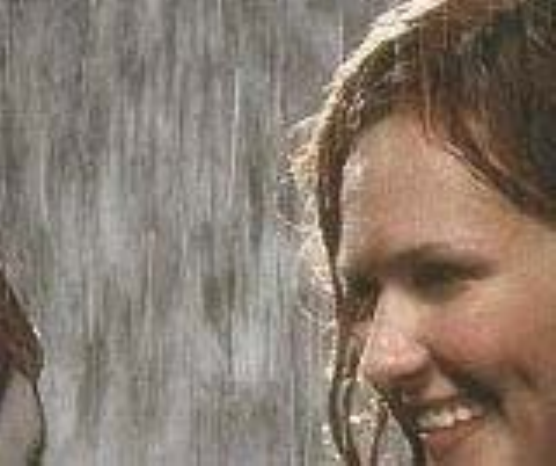}
		&\includegraphics[width=0.15\textwidth,height=0.085\textheight]{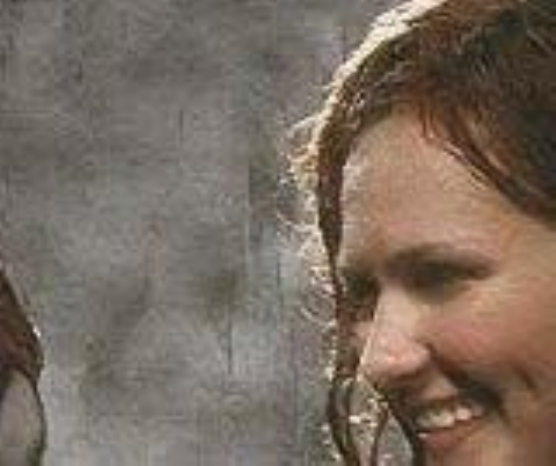}\\
		4.0871 / 2.5465&3.9705 / 2.4796&3.9211 / 2.3492\\
		(a)&(b)&(c)\\	
	\end{tabular}
	\caption{Analysis of the contrastive prior~(CP). (a) The input real-world rainy images. (b) The results without CP. (c) The results with CP. The NIQE / PI values are reported below each image. 
	}  
	\label{fig:Derain_vgg}
\end{figure}
\subsection{Evaluation of Details Preservation }
We extract the high-frequency information of the restored results and compare the details preservation of the different methods in Fig.~\ref{fig:Derain_details}. We can see that JORDER, SIRR, Syn2Real and DualGCN leave large amount of rain in the results, while DDN and MSPFN erase many of the informative details.
In contrast, the proposed method leaves the least undesirable rain and preserves rich and credible details, such as the edges of the windows and the fence outside the building framed in the red and green boxes, respectively. Therefore, our method significantly advances the state-of-the-art in content information preservation.

\subsection{Analysis and Discussion}
\subsubsection{Effectiveness of Recursive Numbers}

We discuss the effect of the recursive unfolding factor in Fig.~\ref{fig:Derain_iteration}, where the recursive stage number~(T) is set as~$1, 2, 3$ and~$4$. Obviously, the effectiveness is gradually improved before the fourth stage. In particular, for the three real-world cases of Fig.~\ref{fig:Derain_iteration}, the results of RDMC$_4$ are even worse than those of RDMC$_3$. Objective results are reported in Table.~\ref{tab:ablation_iteration}. It can be found that RDMC$_3$ achieves the best performance, which is consistent with the visual appearance. Therefore, the optimal recursive number is set as~$3$, and the experiments involved in this paper are conducted under this setting.

%\begin{table}[htp]
%	\begin{center}
%		\centering
%		\caption{\textcolor{red}{Quantitative comparison about the selection of recursive stage number~($T$). We set~$T=1,2,3,4$, and the corresponding PSNR and SSIM values are reported. The metric evaluation demonstrates that~$T=3$ is optimal, which is consistent with the subjective comparison.}}
%		\label{tab:ablation_iteration}
%		\begin{tabular}{cccccccccccc}
%			\hline
%			&&RDMC$_1$&RDMC$_2$&RDMC$_3$&RDMC$_4$\\ \hline
%			\multirow{2}{*}{Rain100L}&PSNR~$\uparrow$ &32.4553&32.8868&\textbf{36.5899}&34.1713\\
%			&SSIM~$\uparrow$ &0.9710&0.9746&\textbf{0.9893}&0.9818\\\hline
%			\multirow{2}{*}{Real-world}&NIQE~$\downarrow$&3.6328 &3.5061 &\textbf{3.2414} &3.2740 \\
%			&PI~$\downarrow$&2.3185 &2.2520 &\textbf{2.1049}&2.2163\\\hline
%		\end{tabular}%
%	\end{center}
%\end{table}

\subsubsection{Effectiveness of Contrastive Prior}
To analyse the contribution of the contrastive prior, a comparison was made between the non-contrast version and the contrastive prior constrained version. Fig.~\ref{fig:Derain_vgg} illustrates the results exhibited on real-world data. Compared with the non-contrast version, the contrastive prior receives results closer to the rain-free background and far away from the rainy images. Specifically, the results under the contrastive prior shown in the last column present an evident improvement in rain removal, while the results of the noncontrastive prior shown in the middle column retain a large amount of visible rain. Therefore, the contrastive prior provides a plausible constraint that positively affects the recovered images.

\begin{figure}[htp]
	\centering
	\setlength{\tabcolsep}{1pt}
	\begin{tabular}{cccccccccccc}
		\includegraphics[width=0.15\textwidth,height=0.132\textheight]{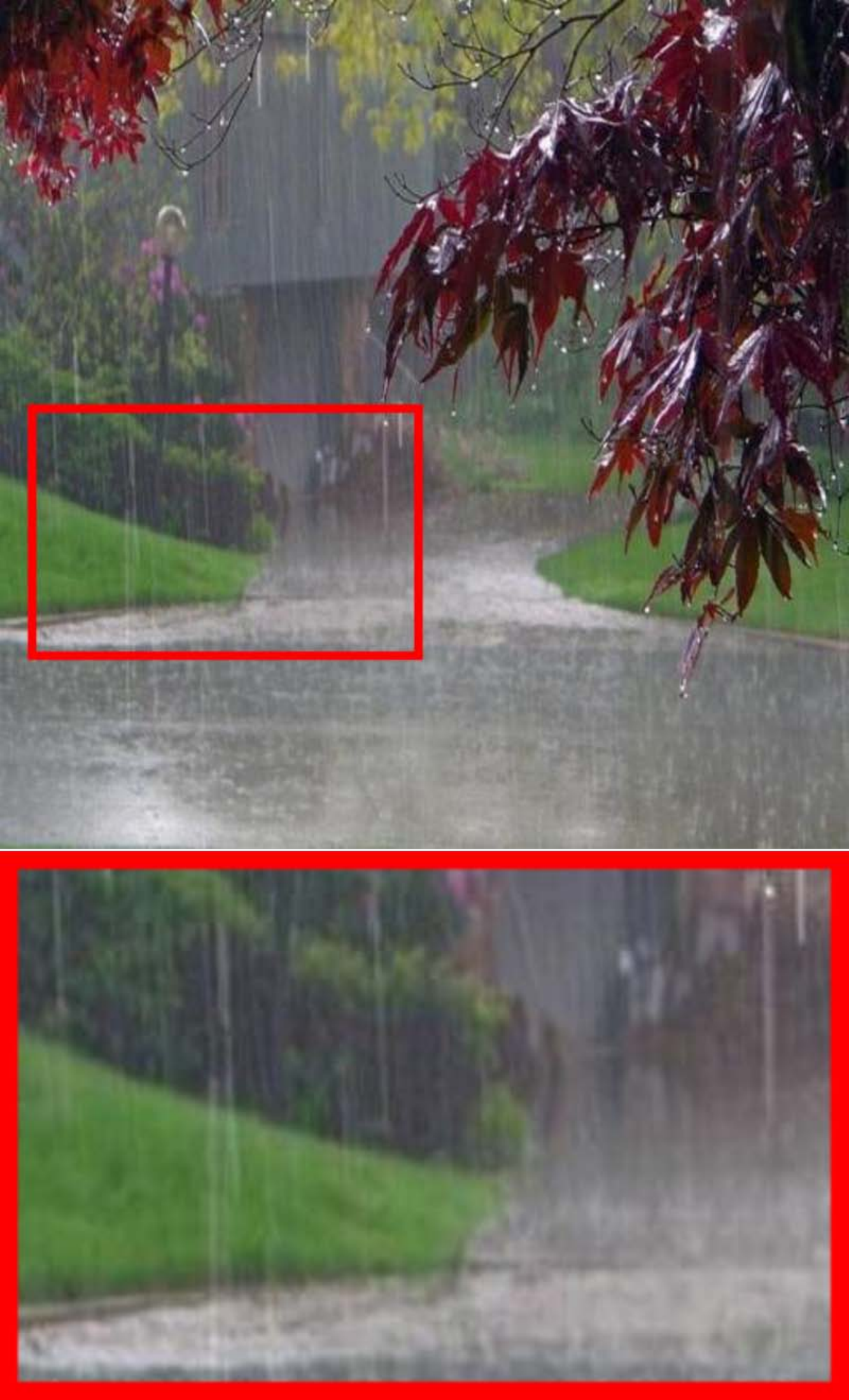}
		&\includegraphics[width=0.15\textwidth,height=0.132\textheight]{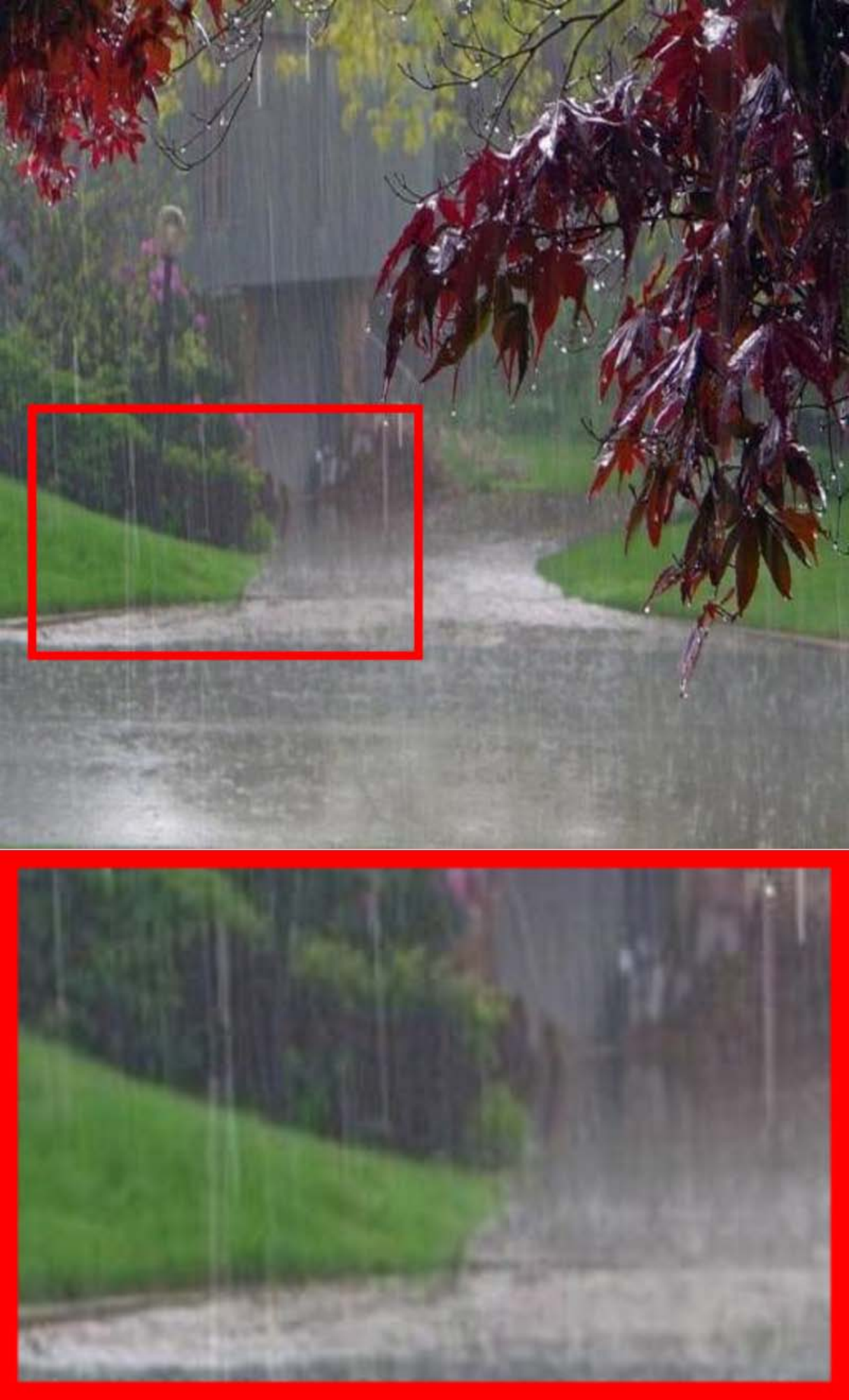}
		&\includegraphics[width=0.15\textwidth,height=0.132\textheight]{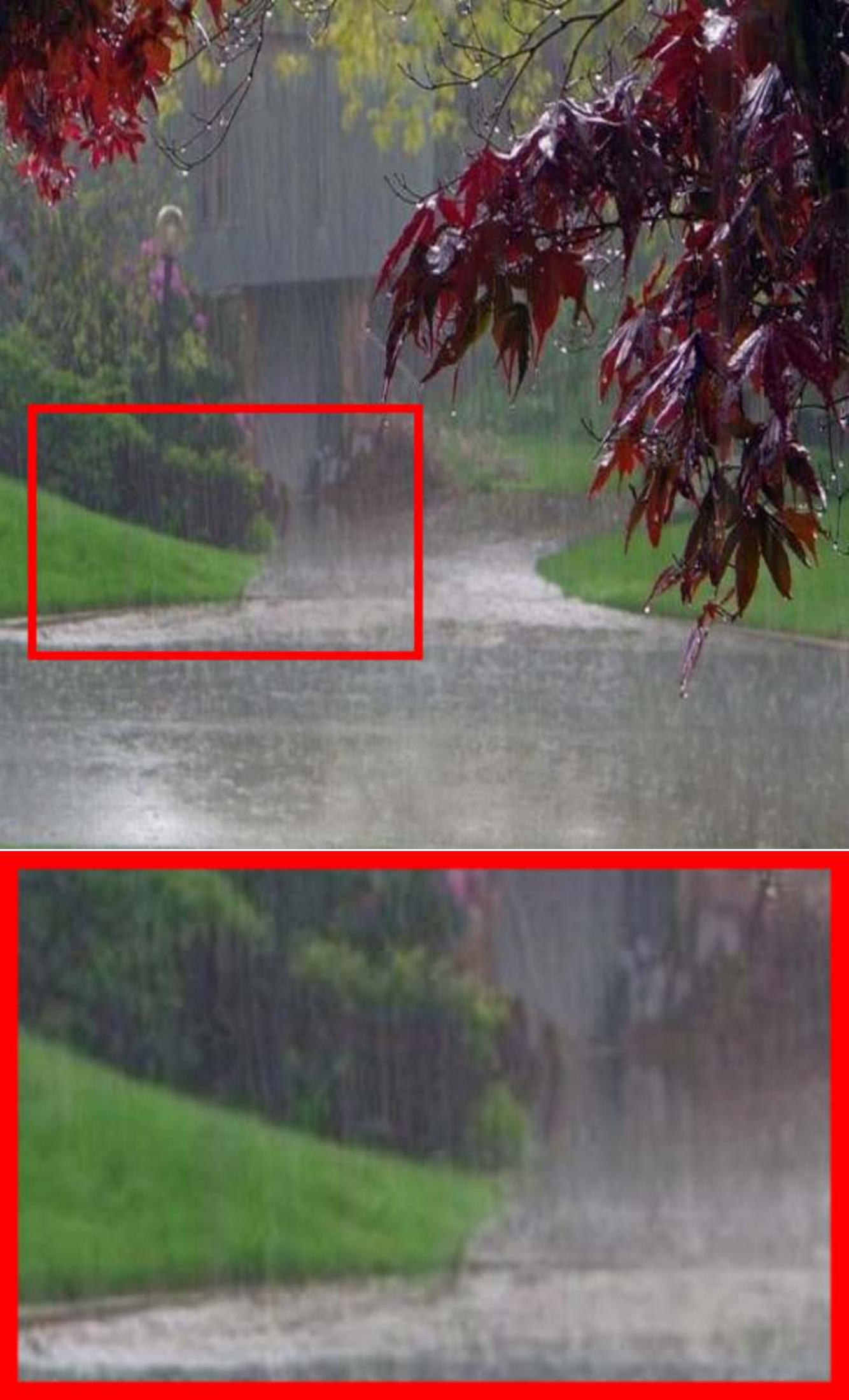}\\
		2.6578 / 1.8964&2.6034 / 1.8771&2.5806 / 1.8561\\
		\includegraphics[width=0.15\textwidth,height=0.132\textheight]{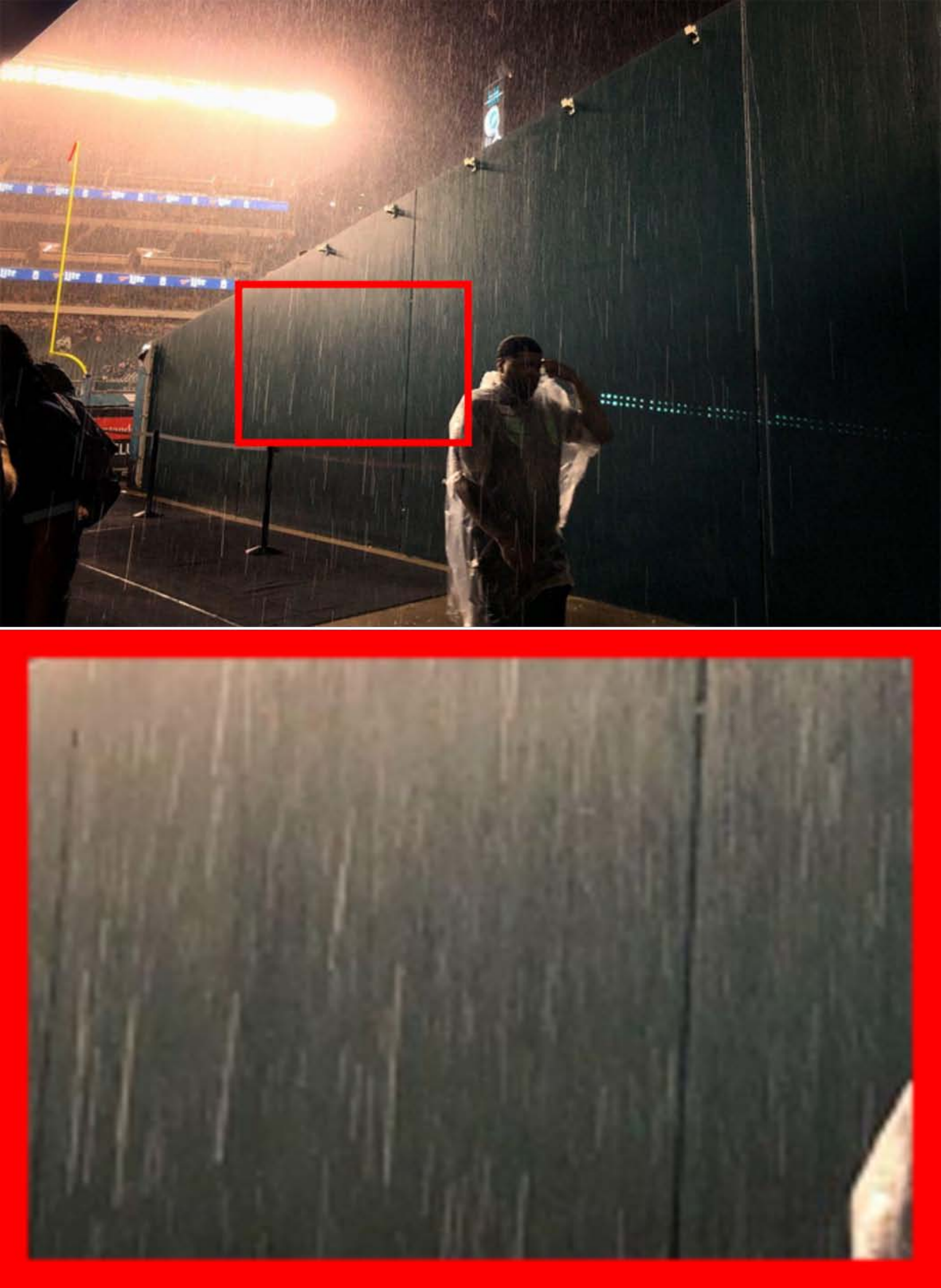}
		&\includegraphics[width=0.15\textwidth,height=0.132\textheight]{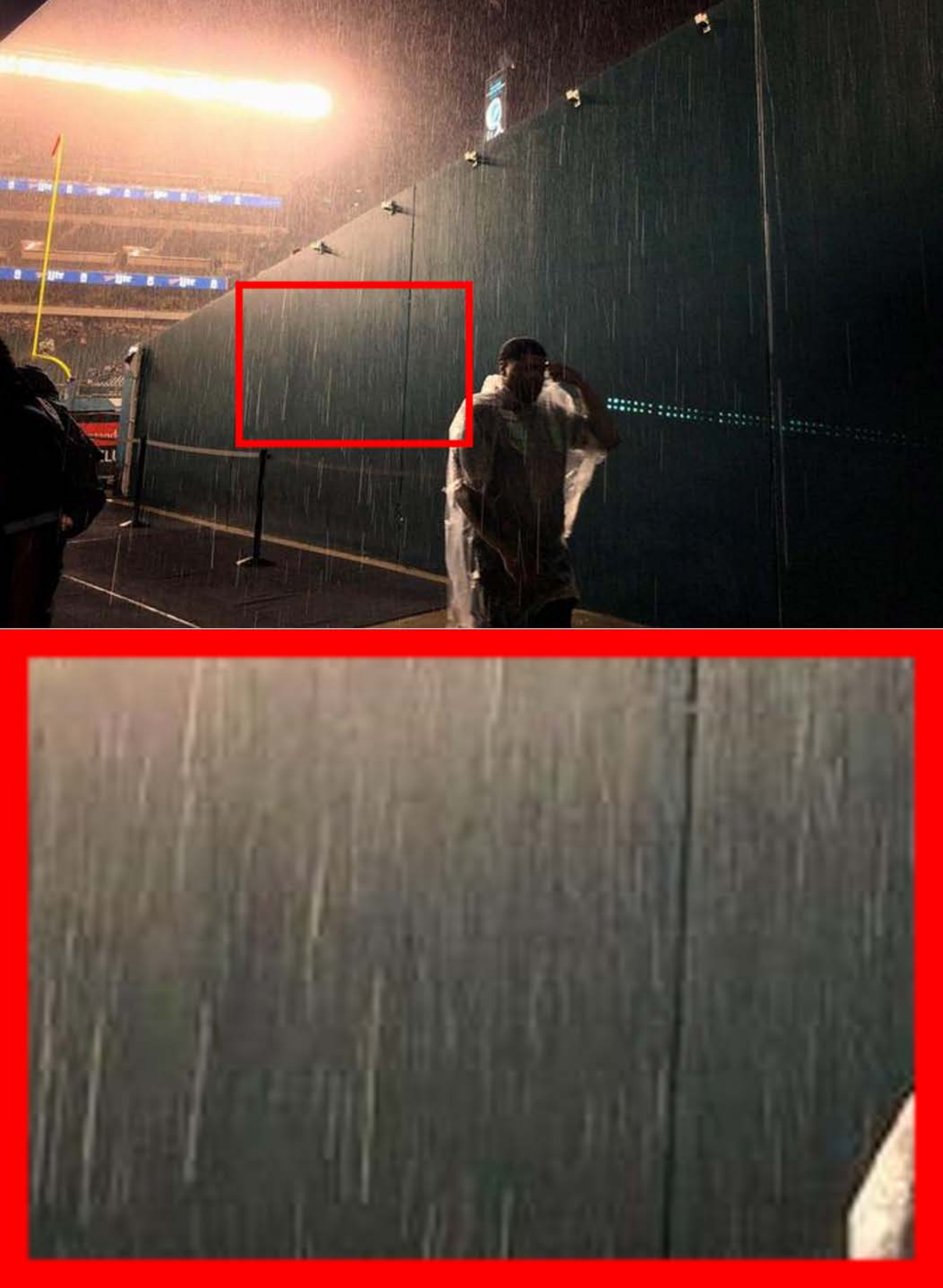}
		&\includegraphics[width=0.15\textwidth,height=0.132\textheight]{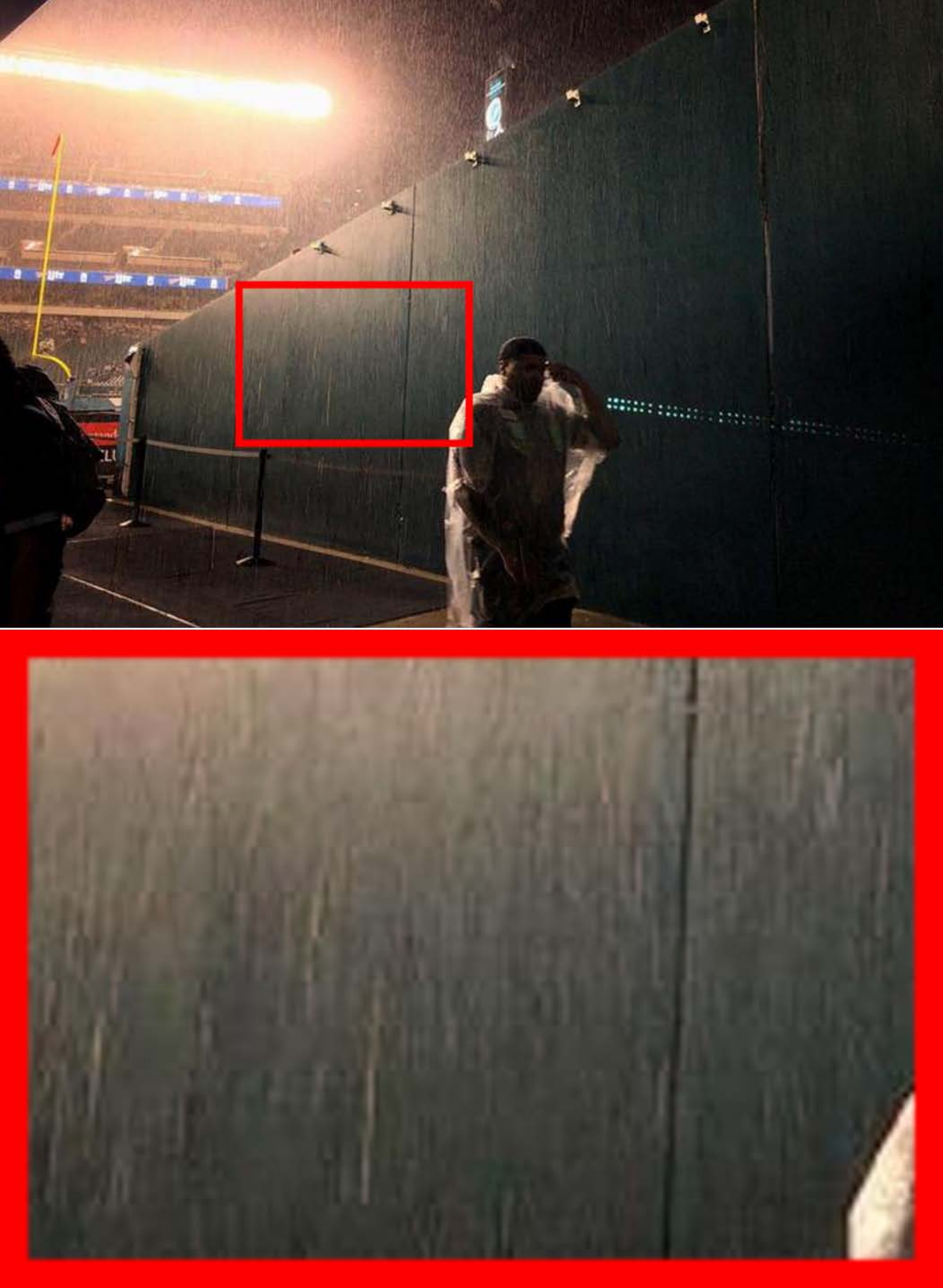}\\	
		4.2046 / 2.6232&3.9884 / 2.5102&3.8573 / 2.4402\\
		(a)&(b)&(c)\\	
	\end{tabular}
	\caption{Ablation study on dynamic cross-level feature recruitment~(DCR). (a)~The input rainy images. (b)~The comparative results without DCR. (c)~The results adopted with DCR. The metrics values on NIQE and PI are reported below. }
	\label{fig:Derain_darts}
\end{figure}
\begin{table}[h]
	\begin{center}
		\centering
		\caption{The performance of each advanced version built upon the baseline, including with and w/o MFE, DCR, and CP. We employed Unet as the baseline. }
		\label{tab:ablation_components}
		\begin{tabular}{c|>{\centering}p{0.7cm}|>{\centering}p{0.7cm}|>{\centering}p{0.7cm}|>{\centering}p{0.8cm}|>{\centering}p{0.8cm}}
			\hline
			Models & MFE& DCR & CP & NIQE~$\downarrow$ & PI~$\downarrow$ 	\tabularnewline \hline
			UNet~(baseline) 
			& \ding{55} & \ding{55} & \ding{55} & 3.9142&2.6121	\tabularnewline
			UNet+MFE
			& \ding{52} & \ding{55} & \ding{55} &3.8378&2.5219	\tabularnewline 
			RDMC w/o CP
			&\ding{52} & \ding{52} & \ding{55} &3.3423&2.2054	\tabularnewline
			RDMC(Ours)
			&\ding{52}& \ding{52}& \ding{52}&\textbf{3.1713}&\textbf{2.1982}	\tabularnewline \hline
		\end{tabular}%
	\end{center}
\end{table}
\begin{figure}[h]
	\centering
	\setlength{\tabcolsep}{1pt}
	\begin{tabular}{cccccccccccc}
		\includegraphics[width=0.24\textwidth,height=0.087\textheight]{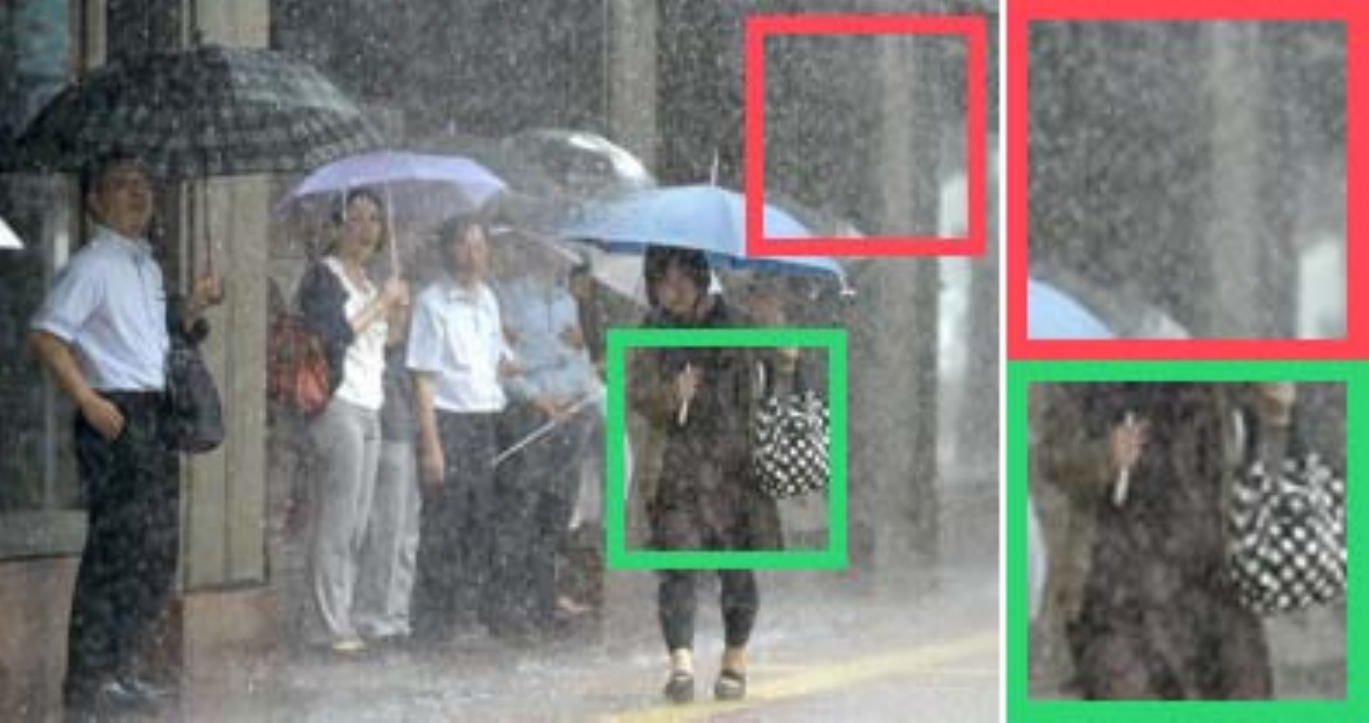}
		&\includegraphics[width=0.24\textwidth,height=0.087\textheight]{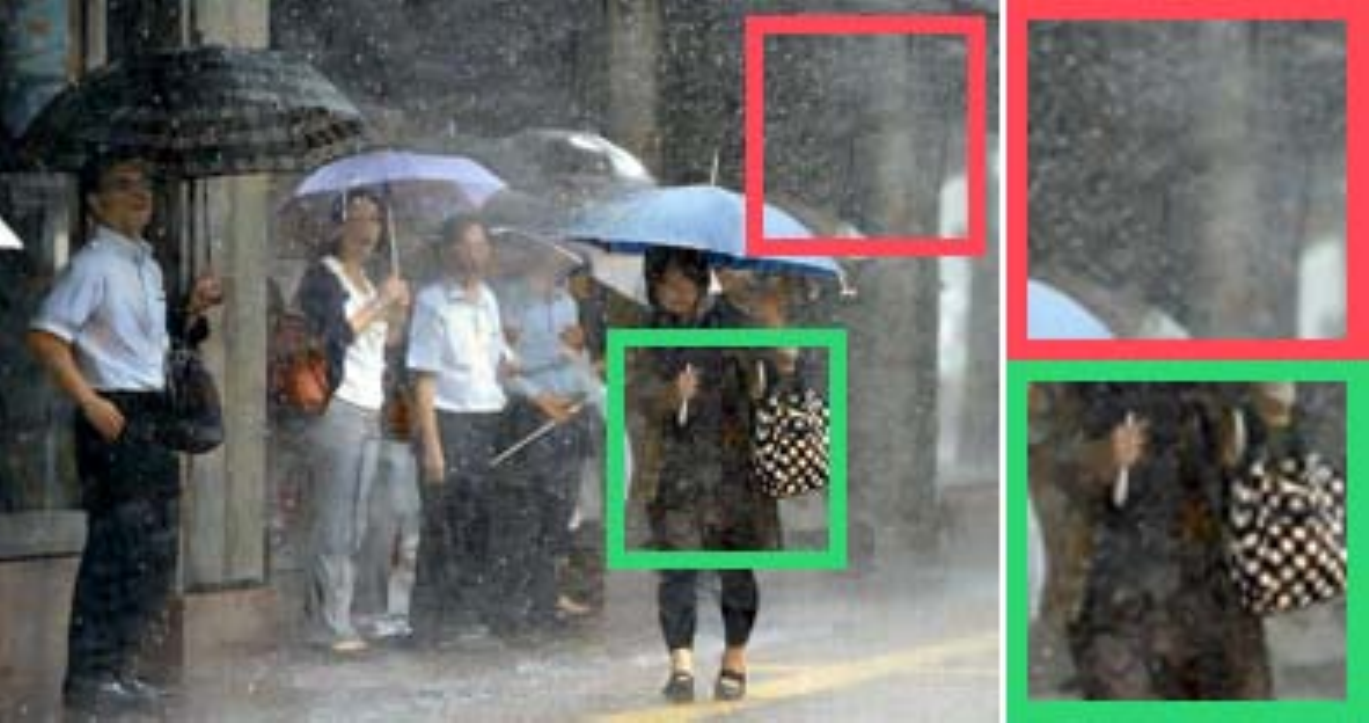}\\
		
		\includegraphics[width=0.24\textwidth,height=0.087\textheight]{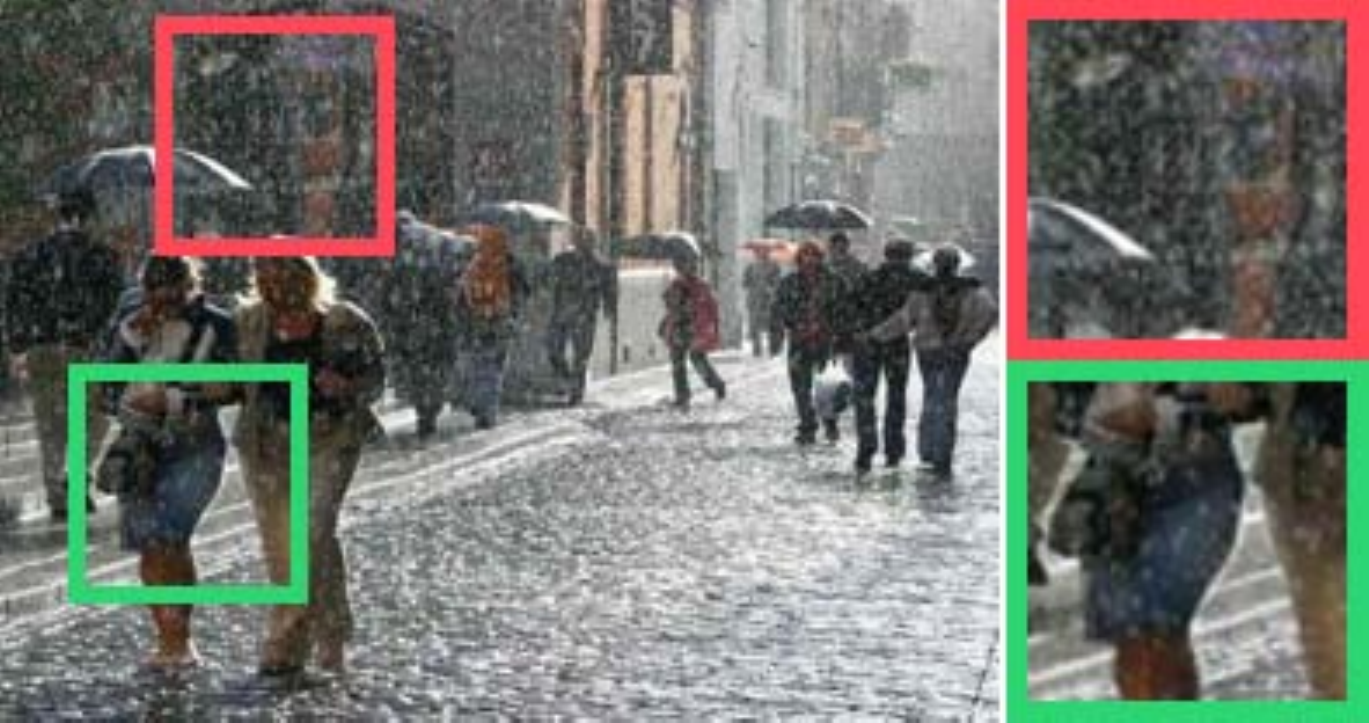}
		&\includegraphics[width=0.24\textwidth,height=0.087\textheight]{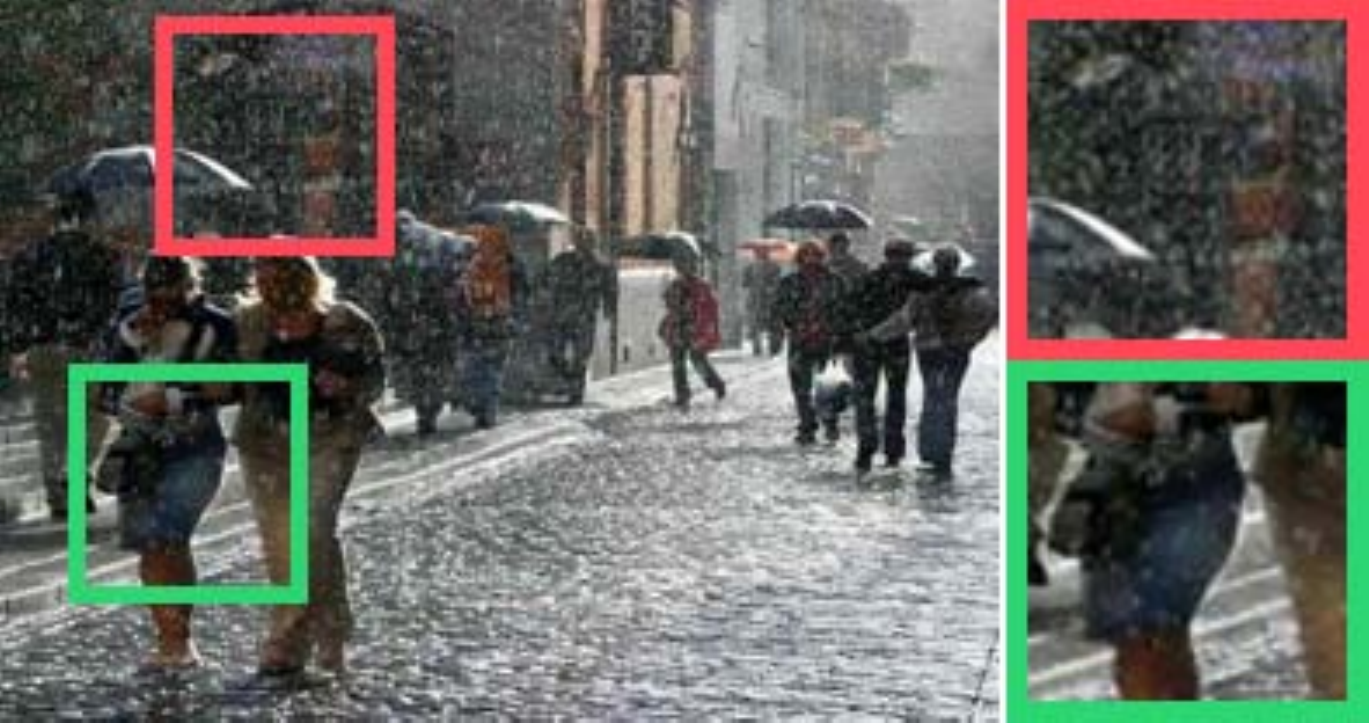}\\
		Input&Ours\\	
	\end{tabular}
	\caption{Failure cases of the proposed RDMC on real-world datasets, where the ambiguous and accumulated rain cannot be extracted by the proposed network, leaving a large amount of interfered rain in the results.}
	\label{fig:fail}
	\vspace{-0.2cm}
\end{figure}

\begin{figure*}[h]
	\centering
	\setlength{\tabcolsep}{1pt}
	\begin{tabular}{cccccccccccc}
		\includegraphics[width=0.16\textwidth, height=0.075\textheight]{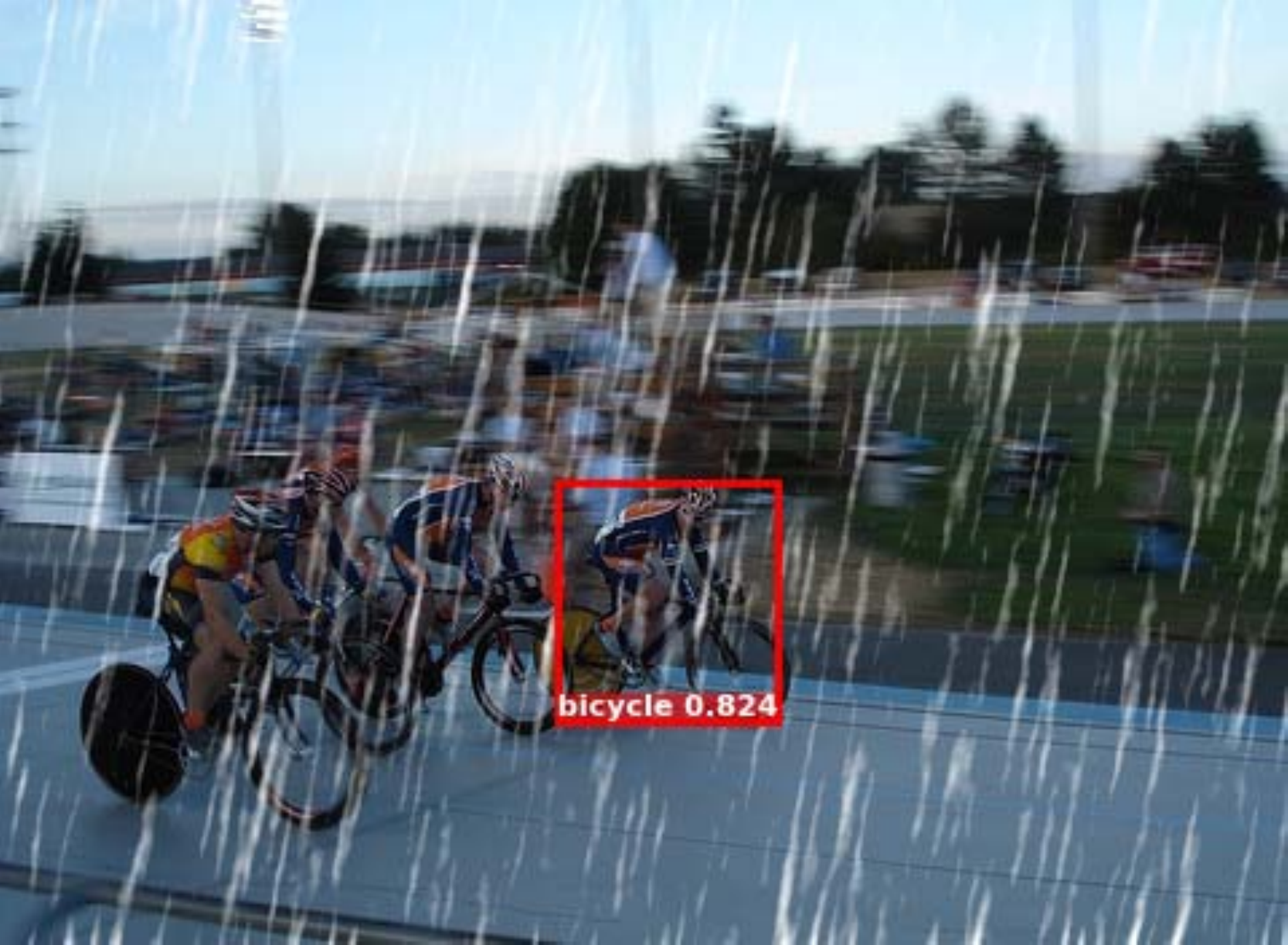}
		&\includegraphics[width=0.16\textwidth, height=0.075\textheight]{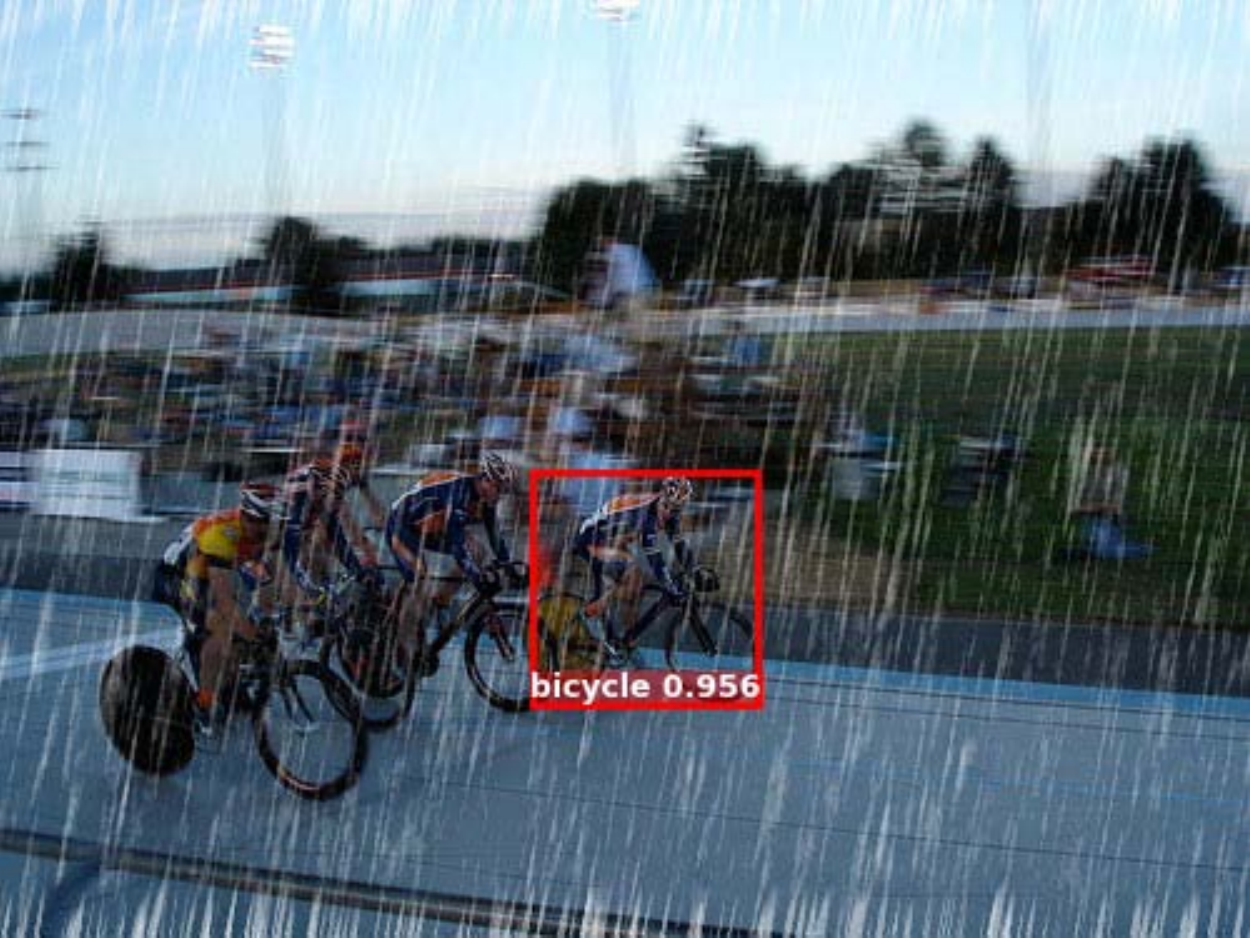}
		&\includegraphics[width=0.16\textwidth, height=0.075\textheight]{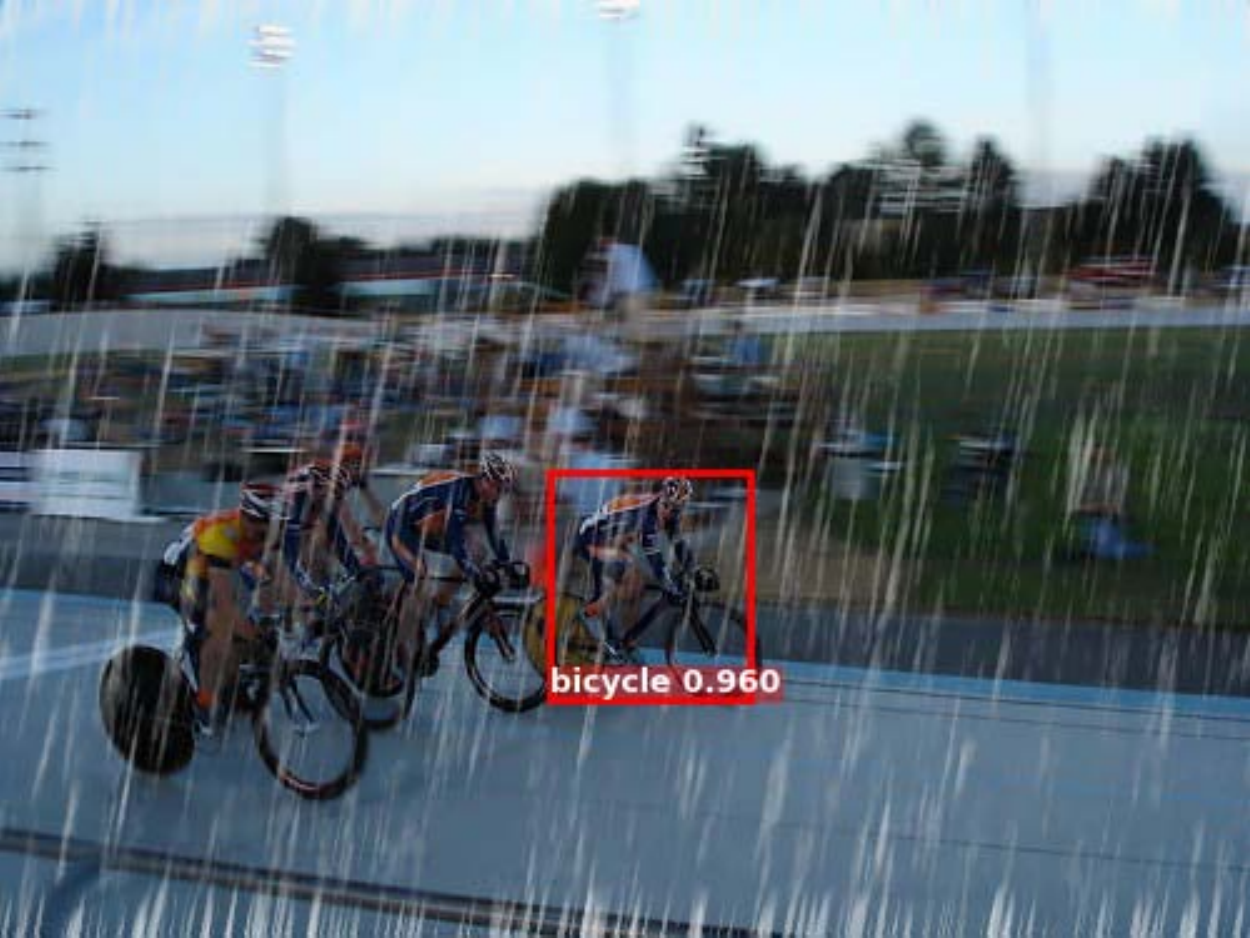}
		&\includegraphics[width=0.16\textwidth, height=0.075\textheight]{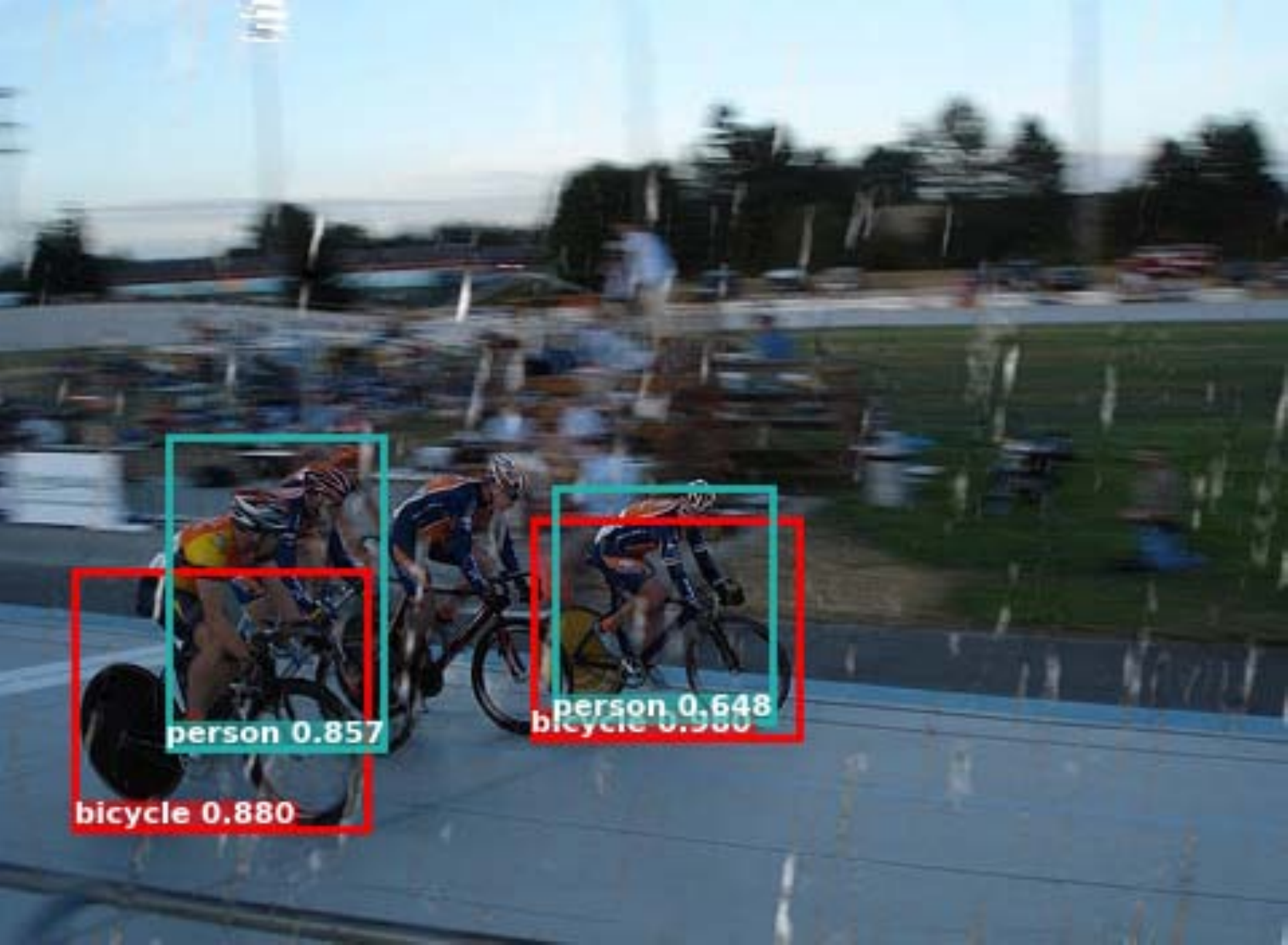}
		&\includegraphics[width=0.16\textwidth, height=0.075\textheight]{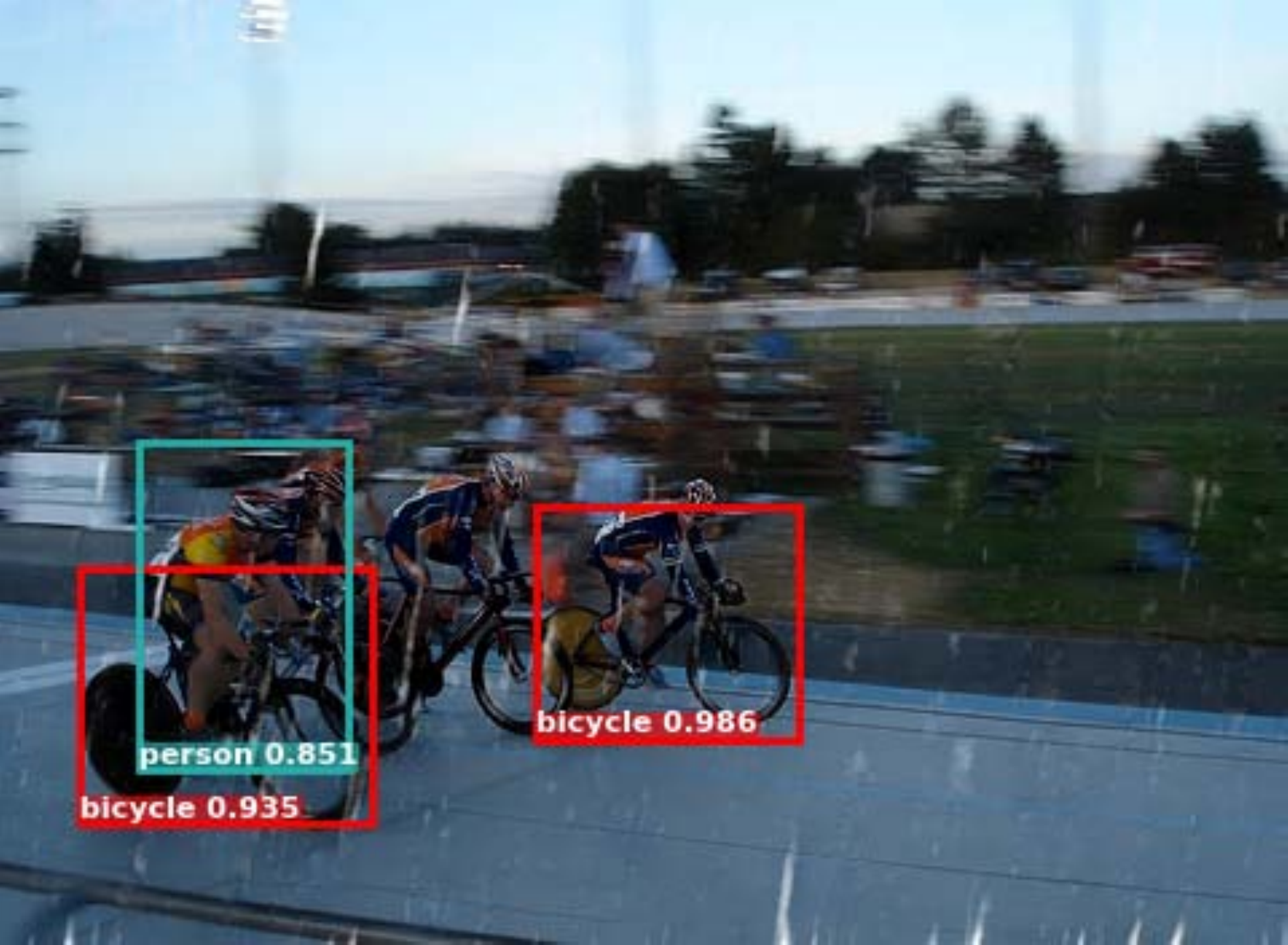}
		&\includegraphics[width=0.16\textwidth, height=0.075\textheight]{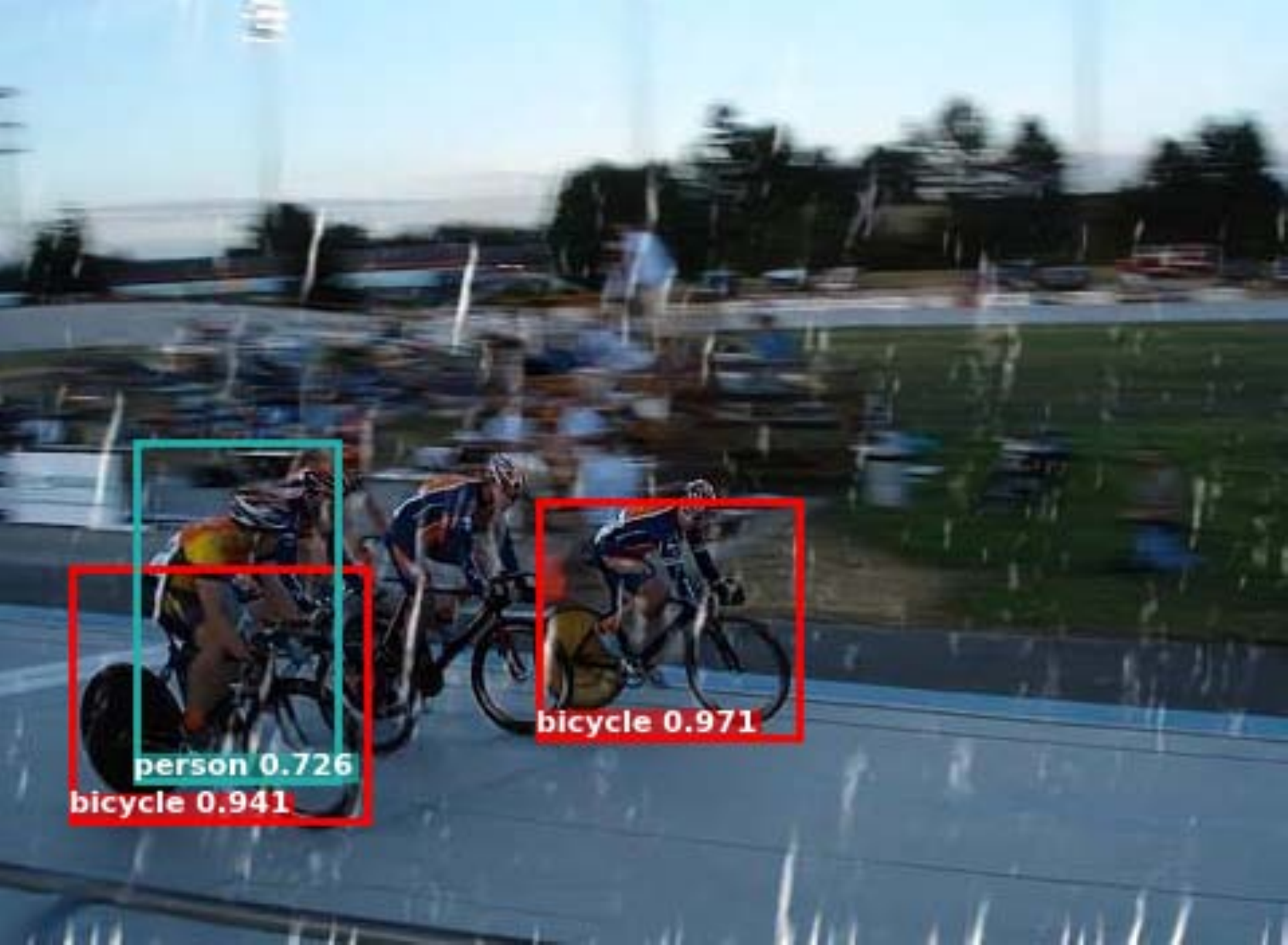}
		\\
		Input&DSC&LP&JORDER&DDN&DualRes\\
		\includegraphics[width=0.16\textwidth, height=0.075\textheight]{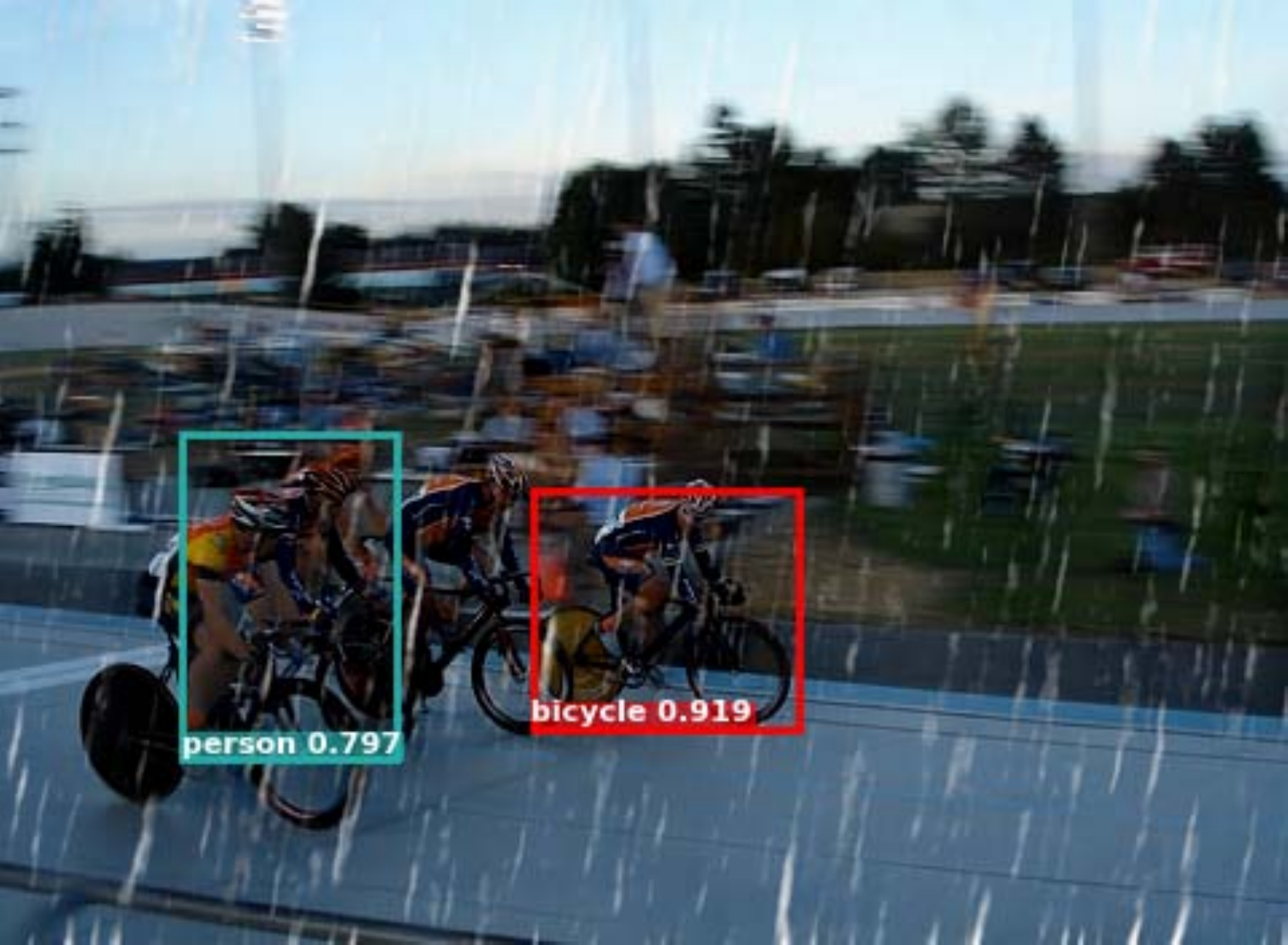}
		&\includegraphics[width=0.16\textwidth, height=0.075\textheight]{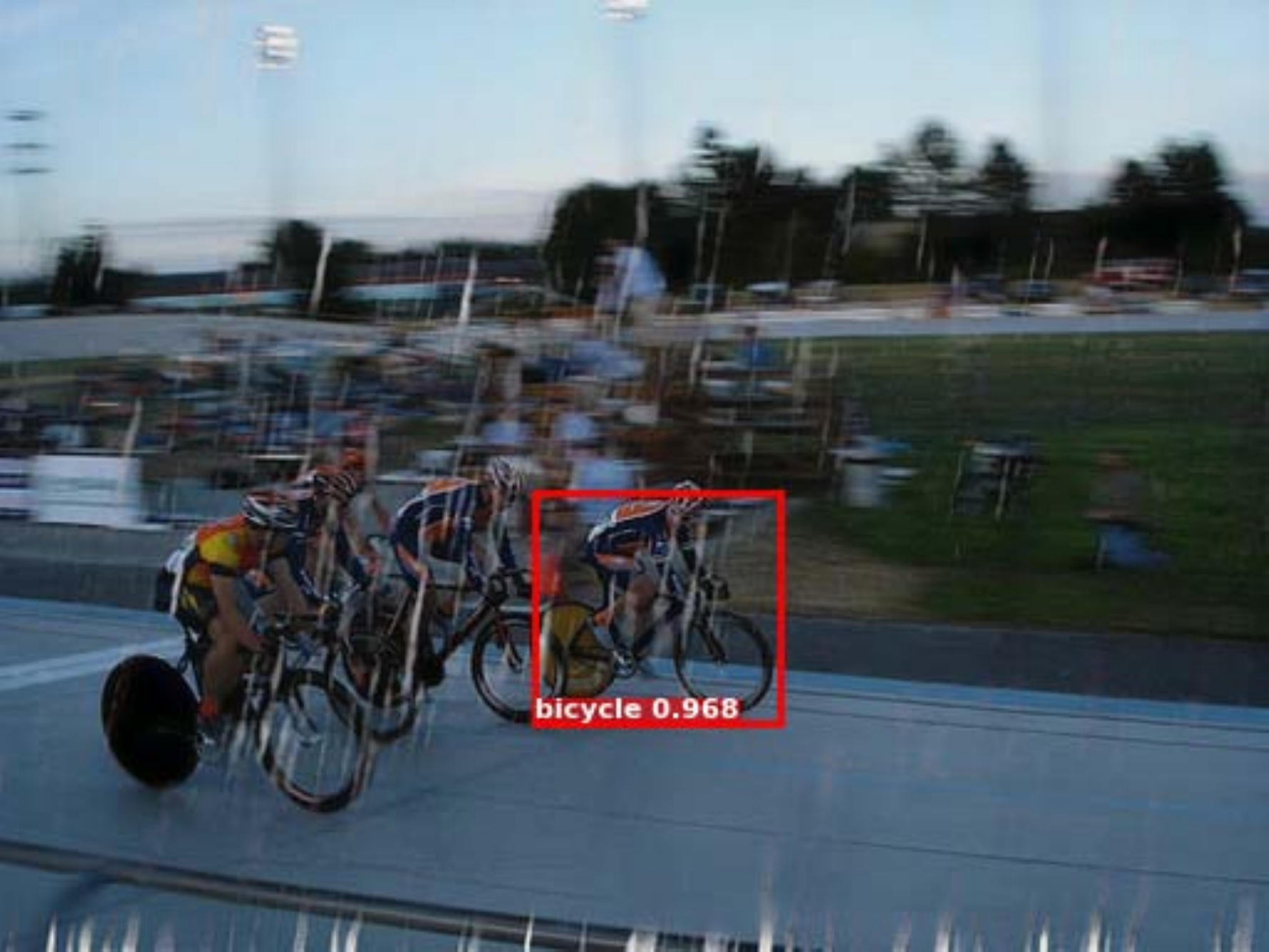}
		&\includegraphics[width=0.16\textwidth, height=0.075\textheight]{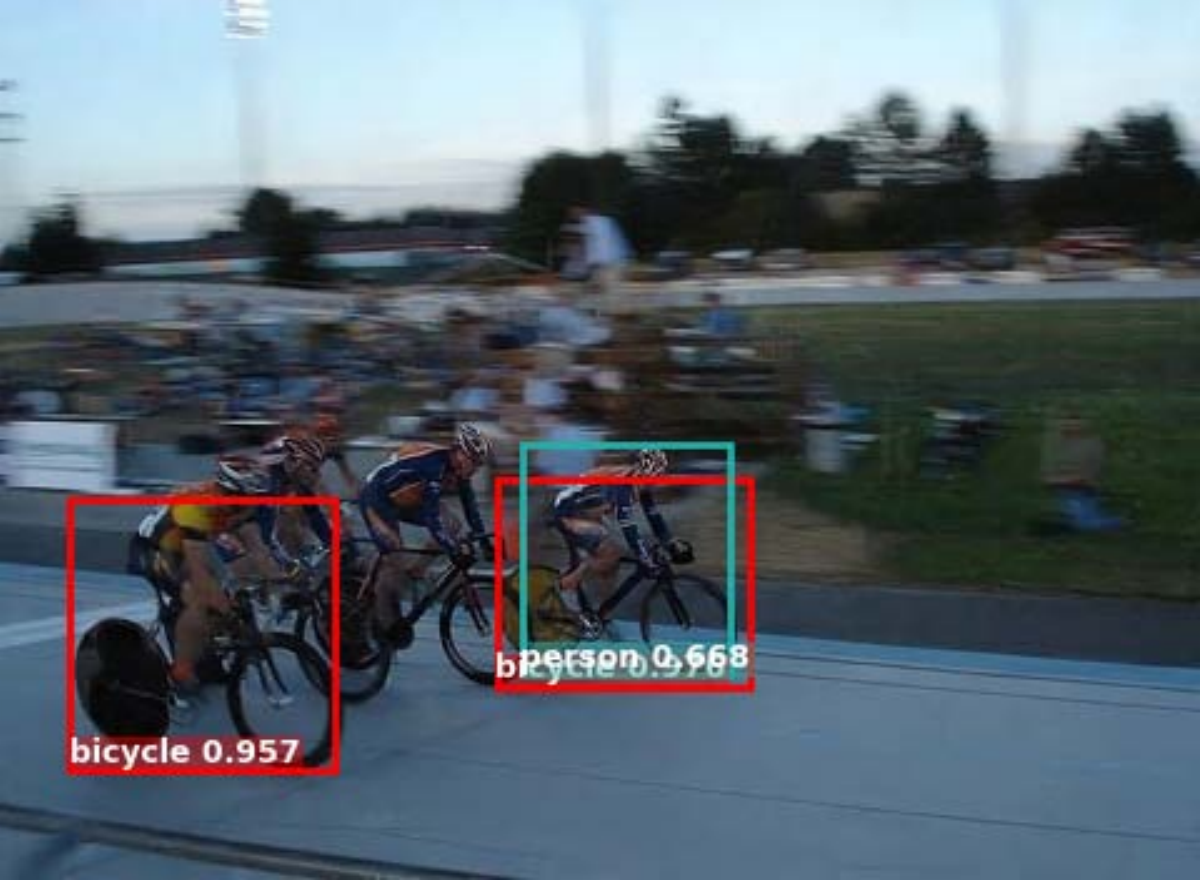}
		&\includegraphics[width=0.16\textwidth, height=0.075\textheight]{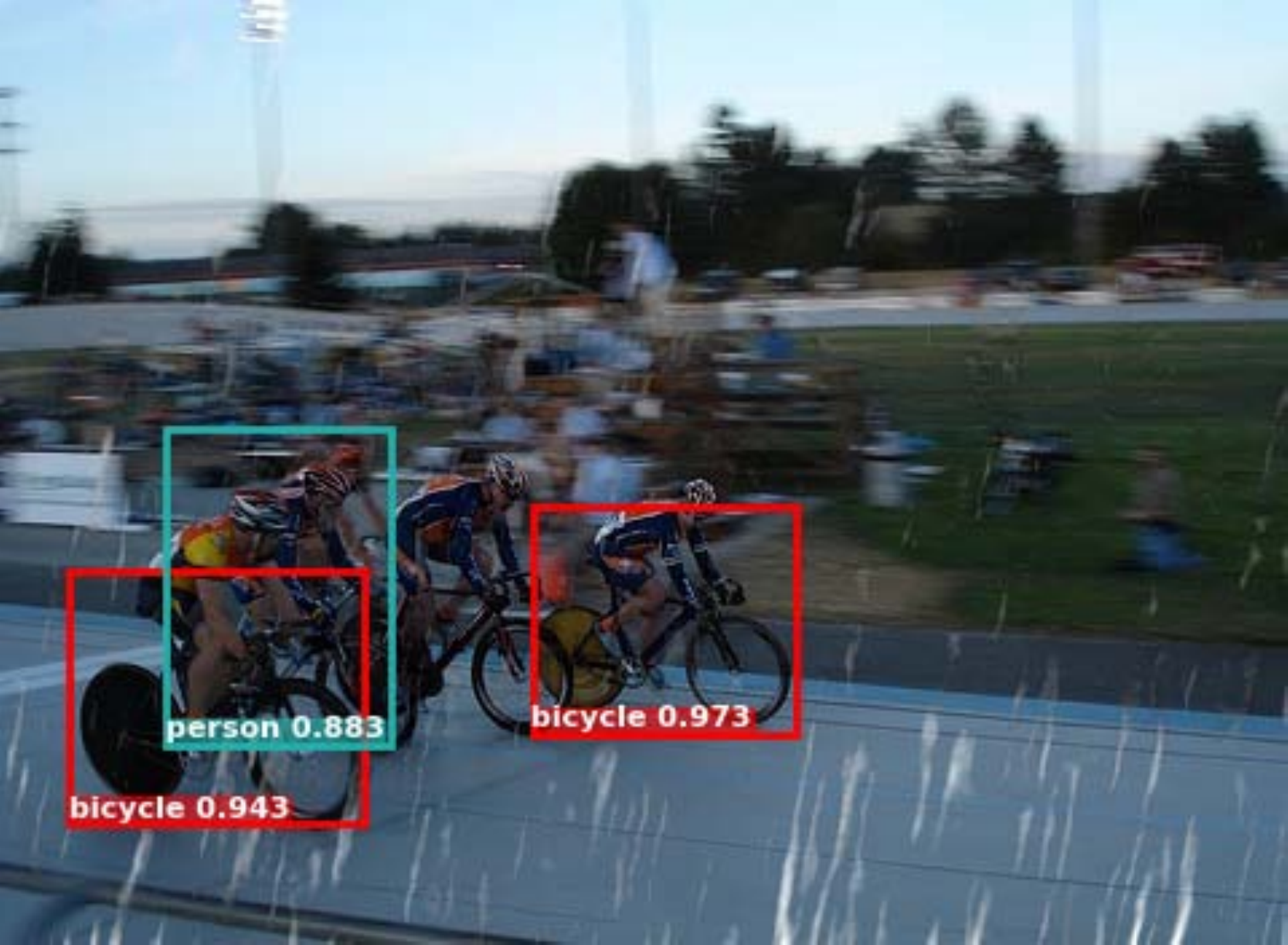}
		&\includegraphics[width=0.16\textwidth, height=0.075\textheight]{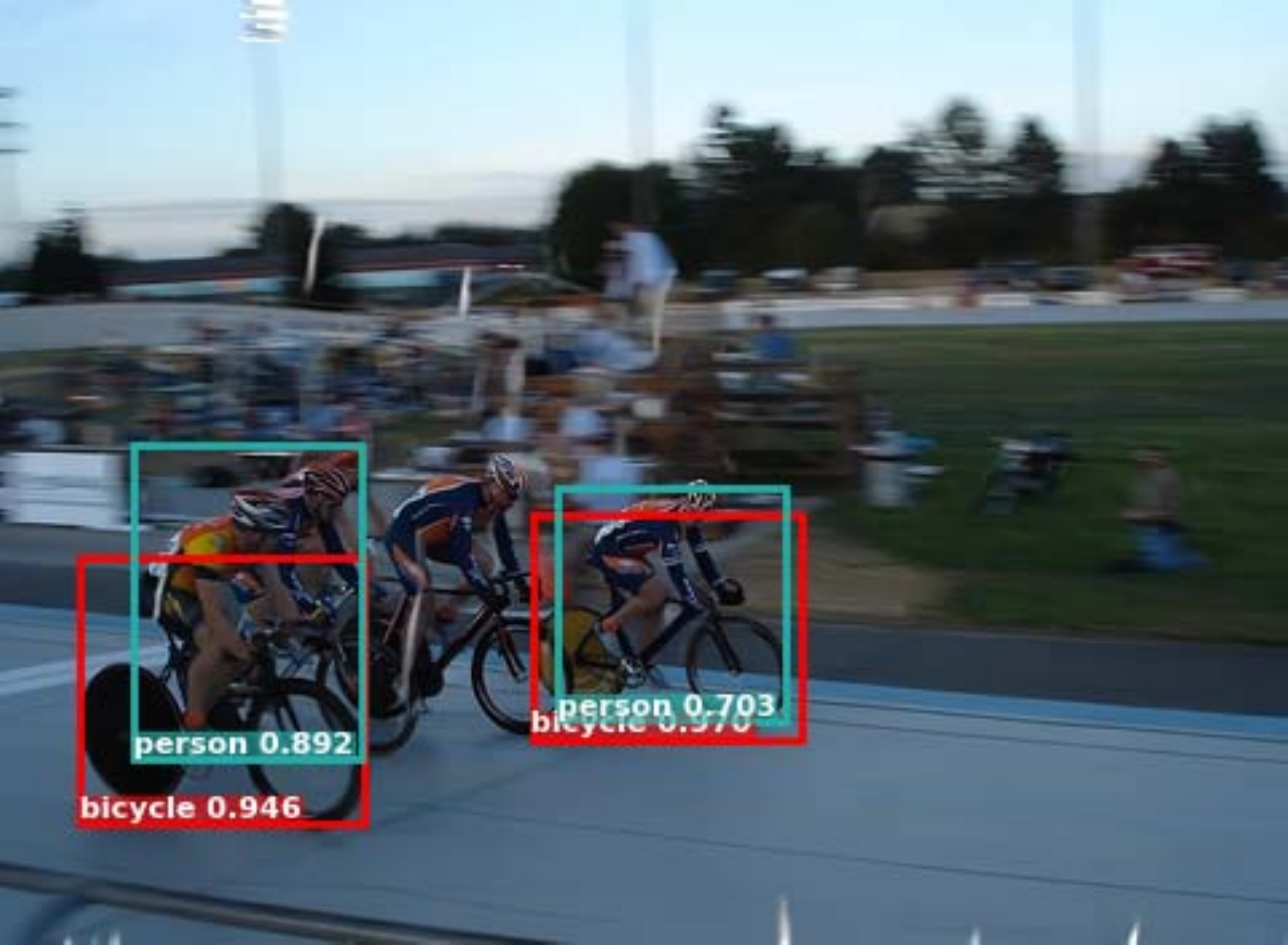}
		&\includegraphics[width=0.16\textwidth, height=0.075\textheight]{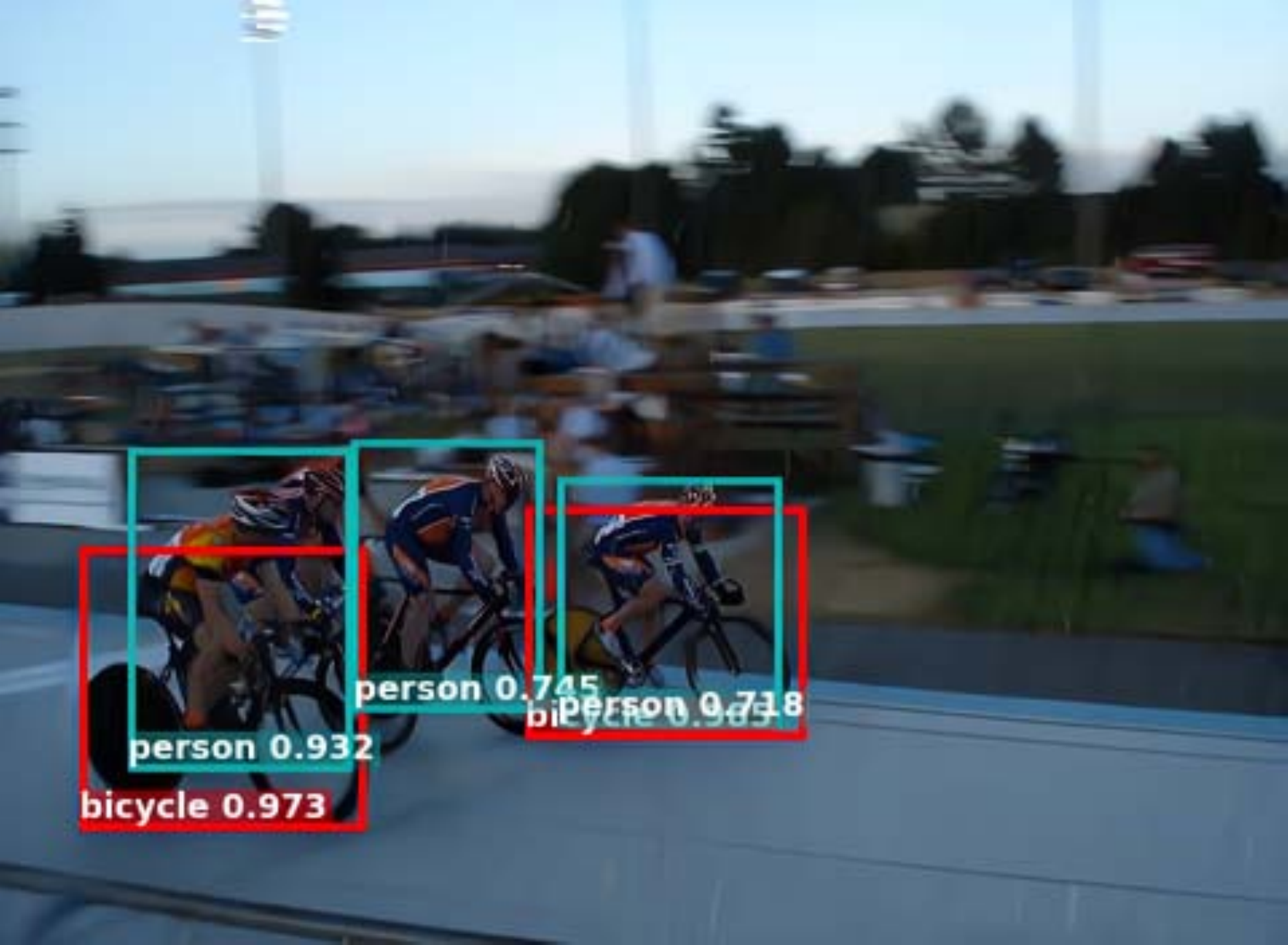}\\	
		SIRR&Syn2Real&MSPFN&DualGCN&MPRNet&Ours\\	
	\end{tabular}
	\caption{Examples of object detection results after the different deraining algorithms. The detected number and confidence score of each object illustrate that the proposed method is more friendly to the subsequent application.}
	\label{fig:Derain_ssd}
\end{figure*}

\begin{figure*}[h]
	\centering
	\setlength{\tabcolsep}{1pt}
	\begin{tabular}{cccccccccccc}
		\includegraphics[width=0.16\textwidth, height=0.075\textheight]{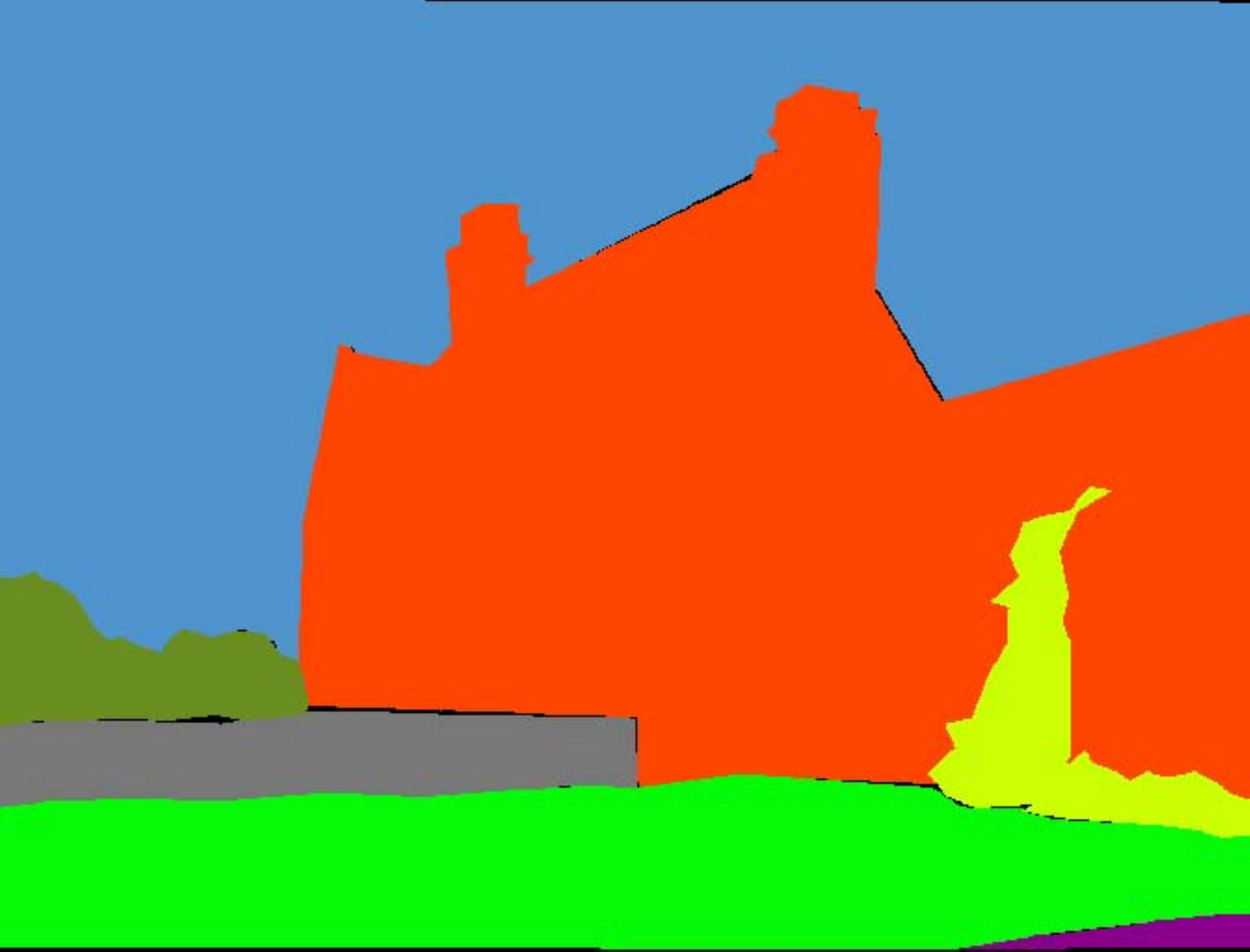}
		&\includegraphics[width=0.16\textwidth, height=0.075\textheight]{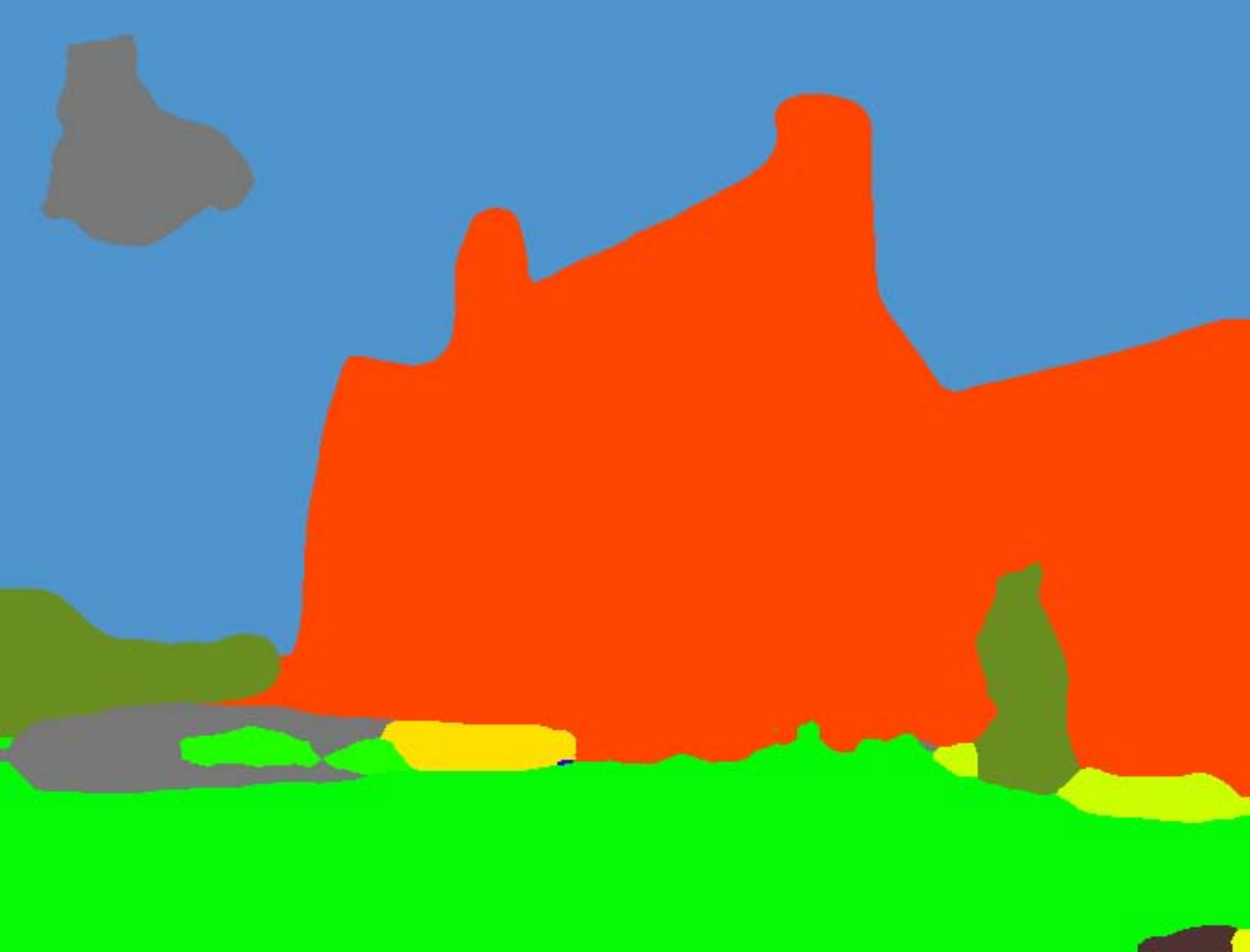}
		&\includegraphics[width=0.16\textwidth, height=0.075\textheight]{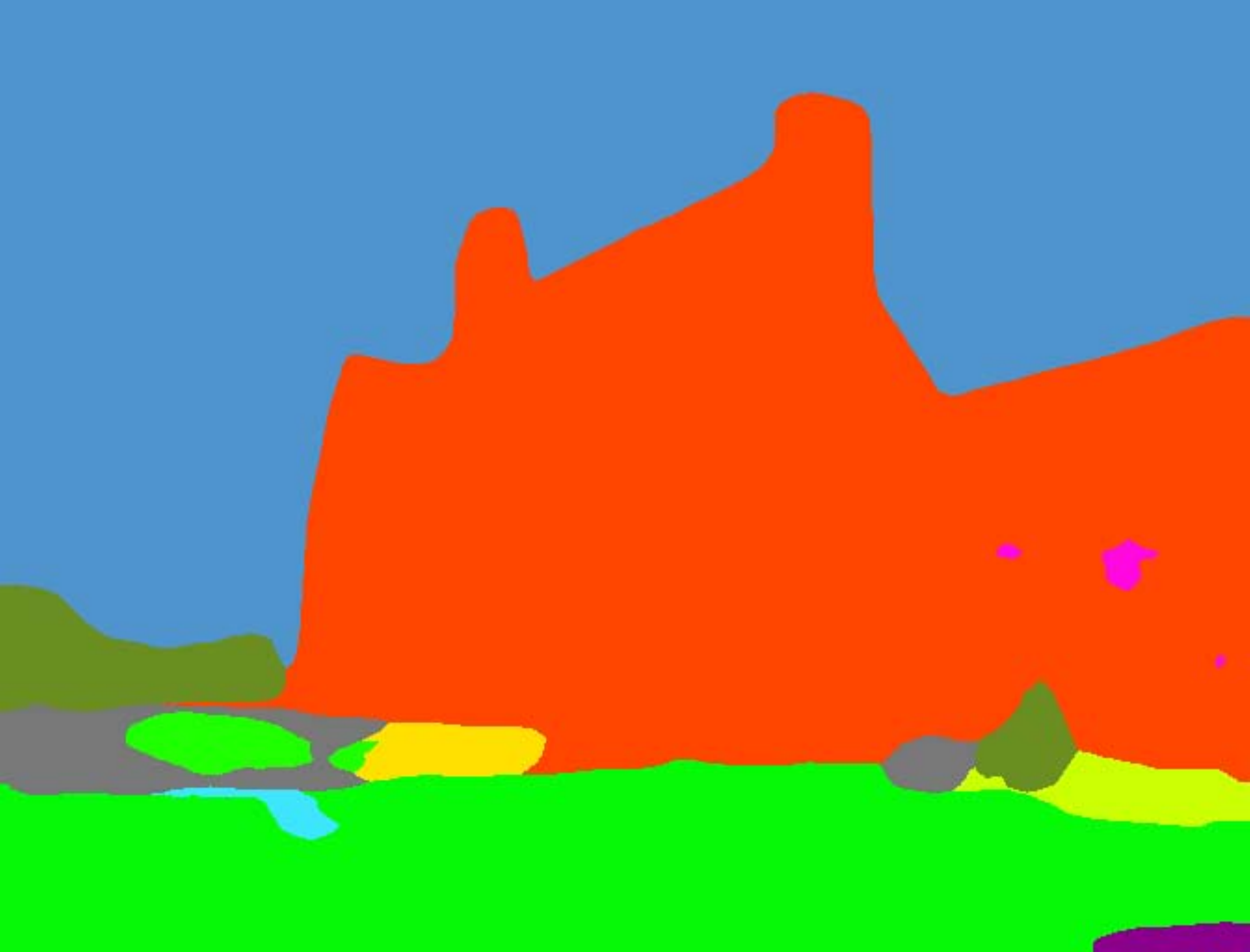}
		&\includegraphics[width=0.16\textwidth, height=0.075\textheight]{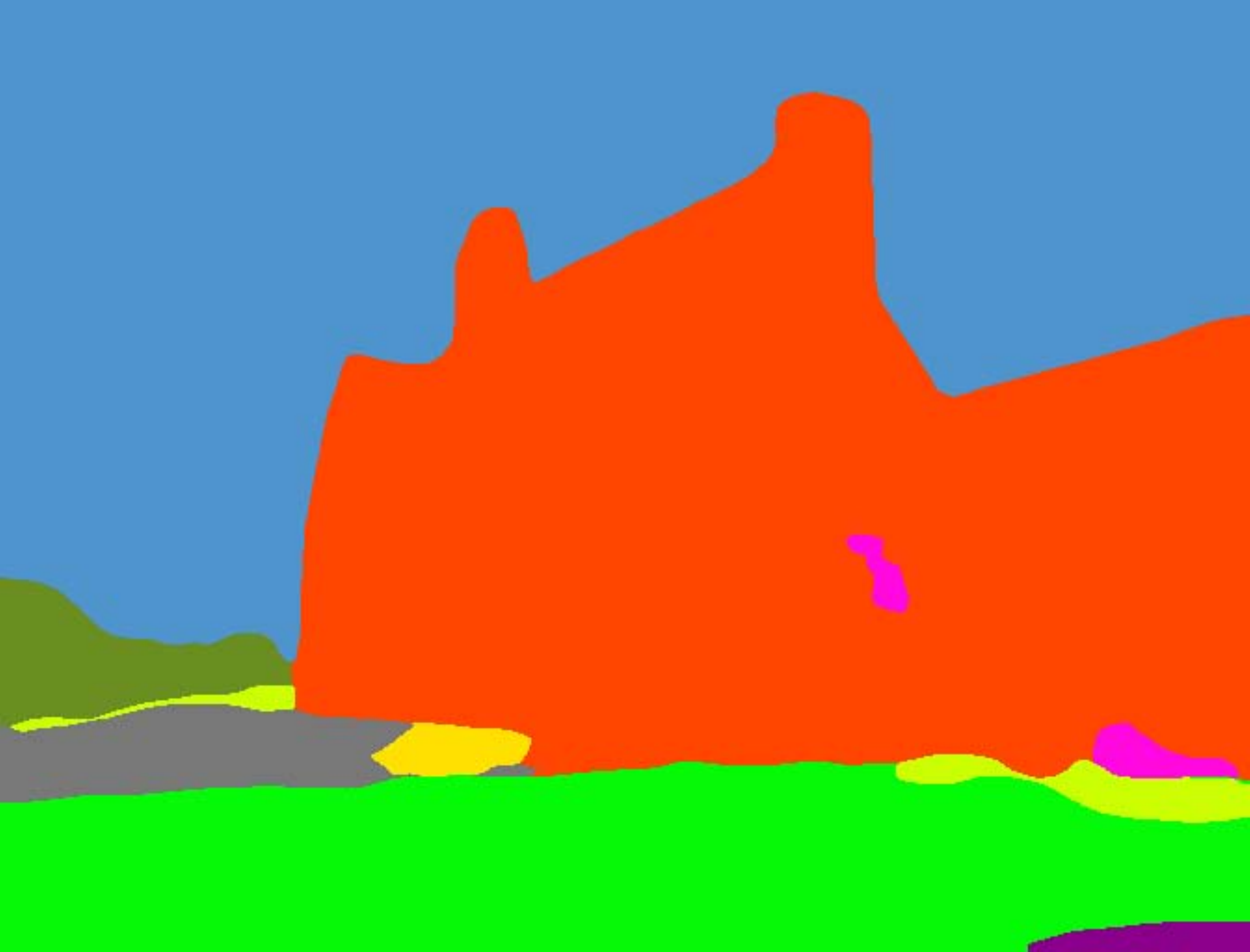}
		&\includegraphics[width=0.16\textwidth, height=0.075\textheight]{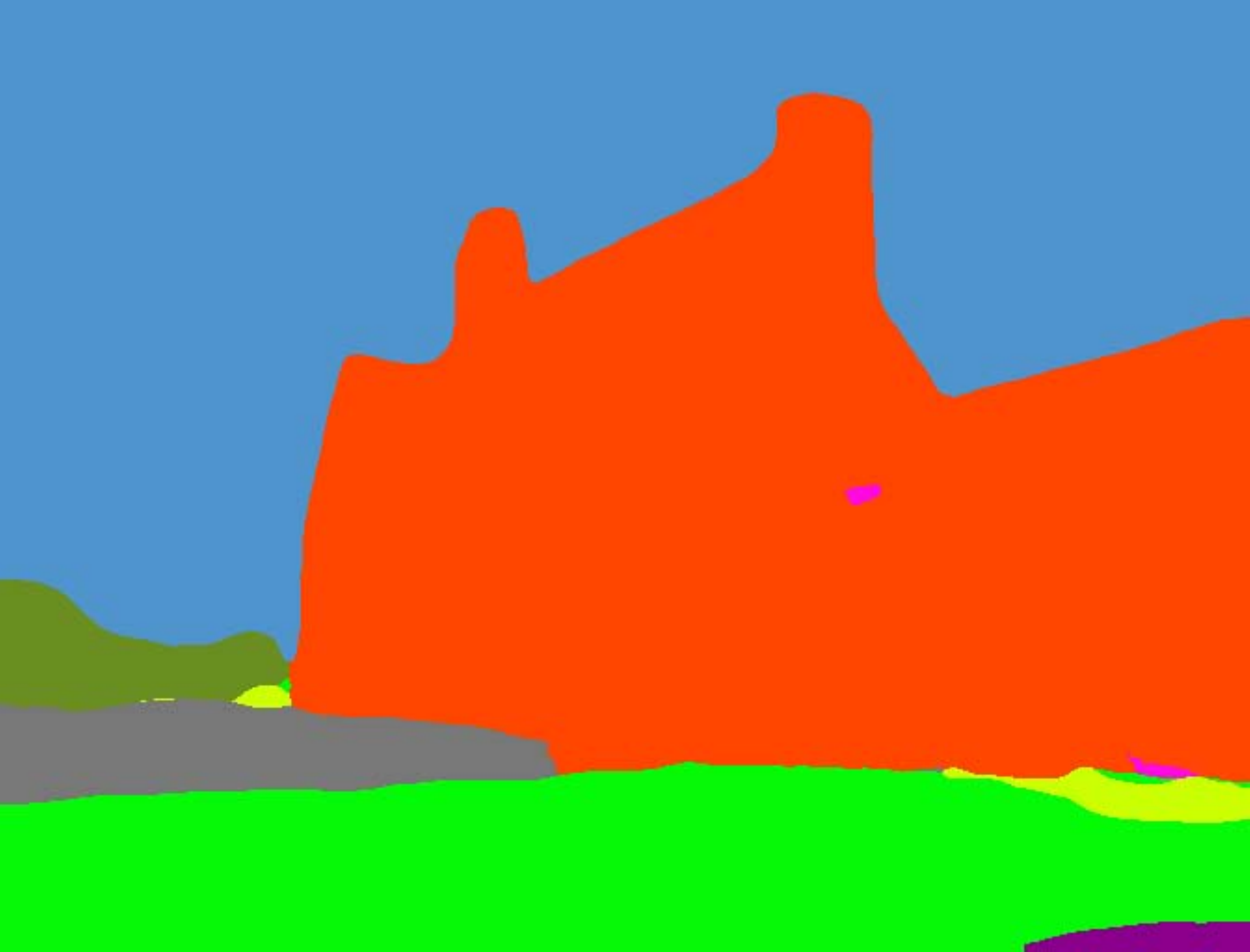}
		&\includegraphics[width=0.16\textwidth, height=0.075\textheight]{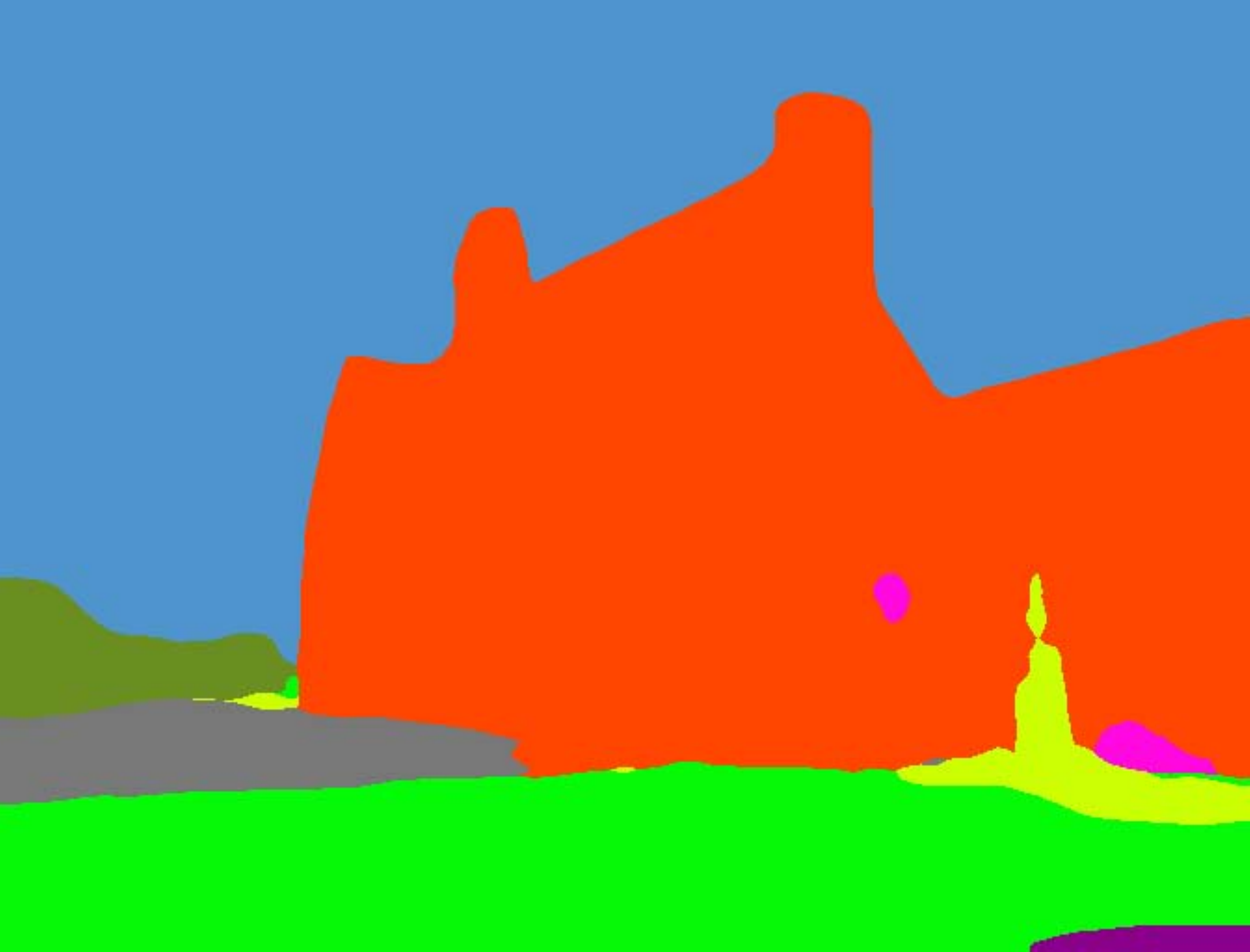}
		\\
		Ground truth&DSC&LP&JORDER&DDN&DualRes\\
		\includegraphics[width=0.16\textwidth, height=0.075\textheight]{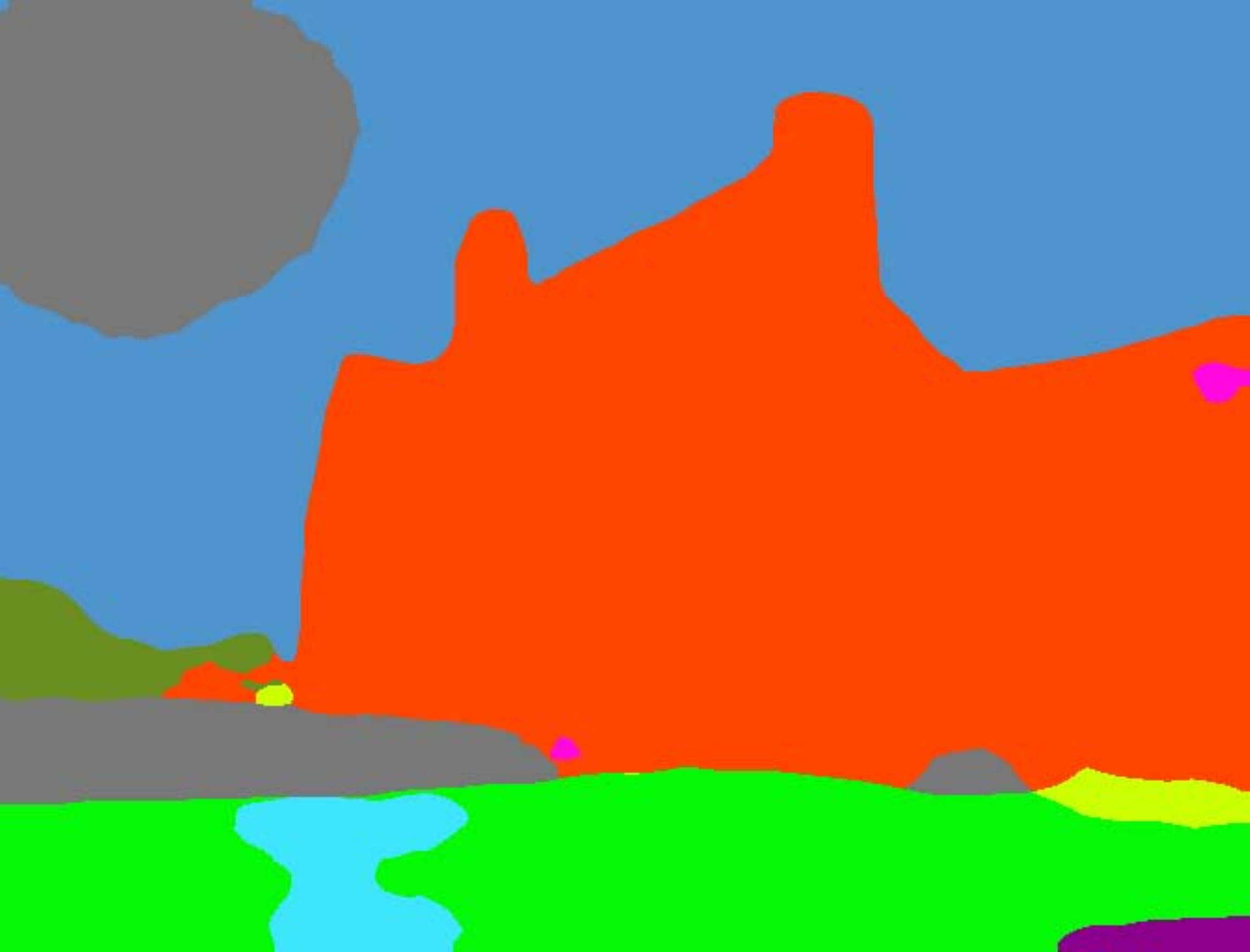}
		&\includegraphics[width=0.16\textwidth, height=0.075\textheight]{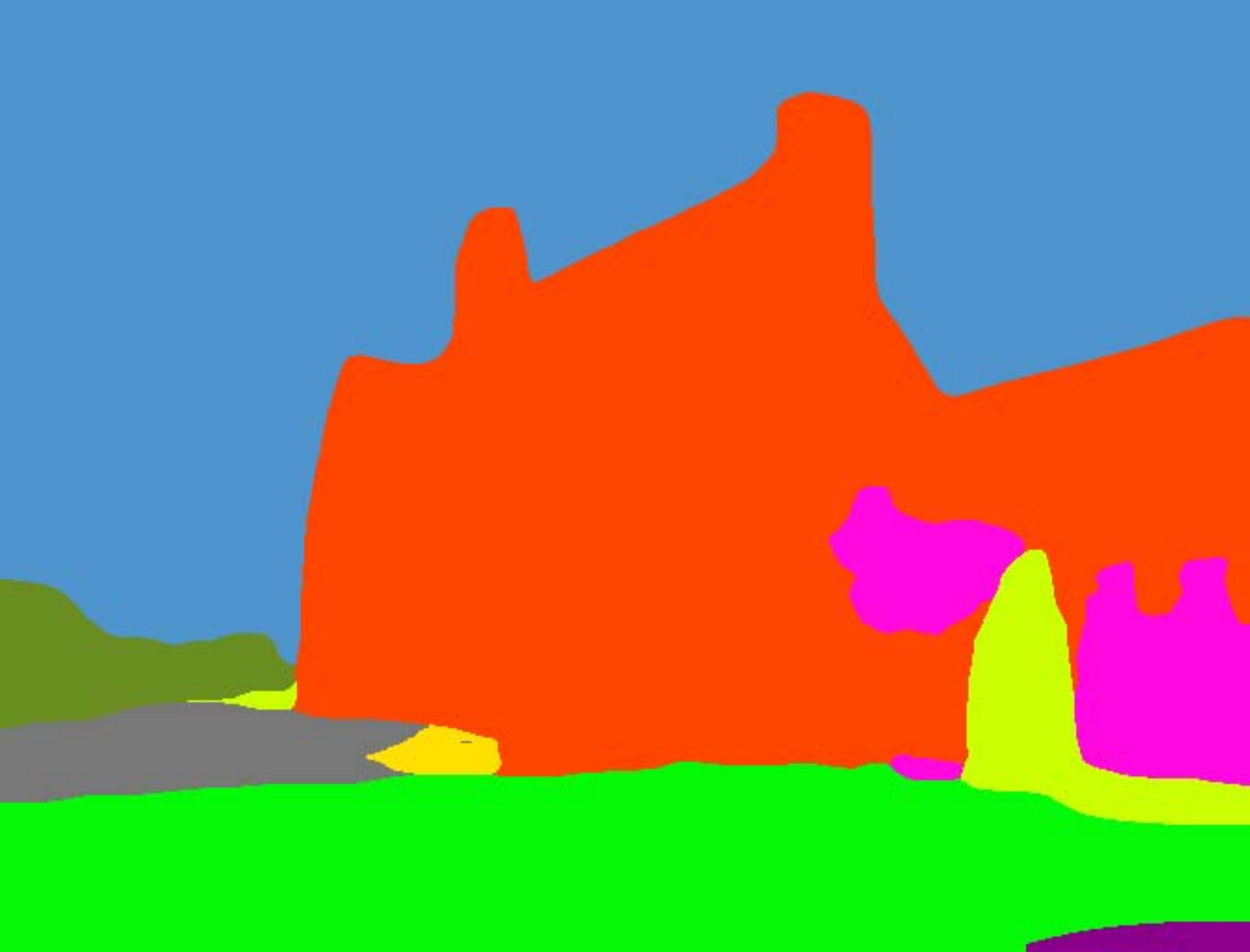}
		&\includegraphics[width=0.16\textwidth, height=0.075\textheight]{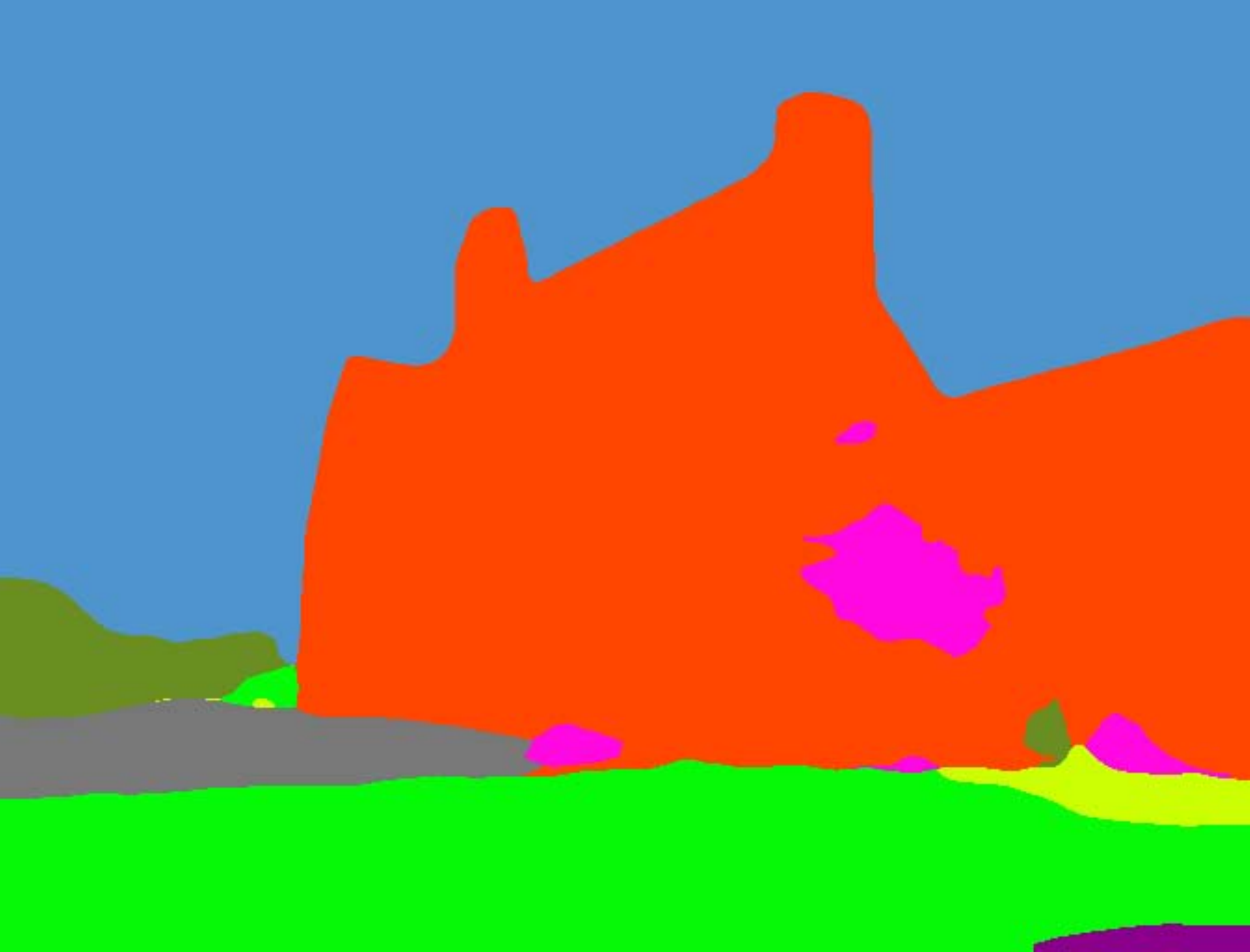}
		&\includegraphics[width=0.16\textwidth, height=0.075\textheight]{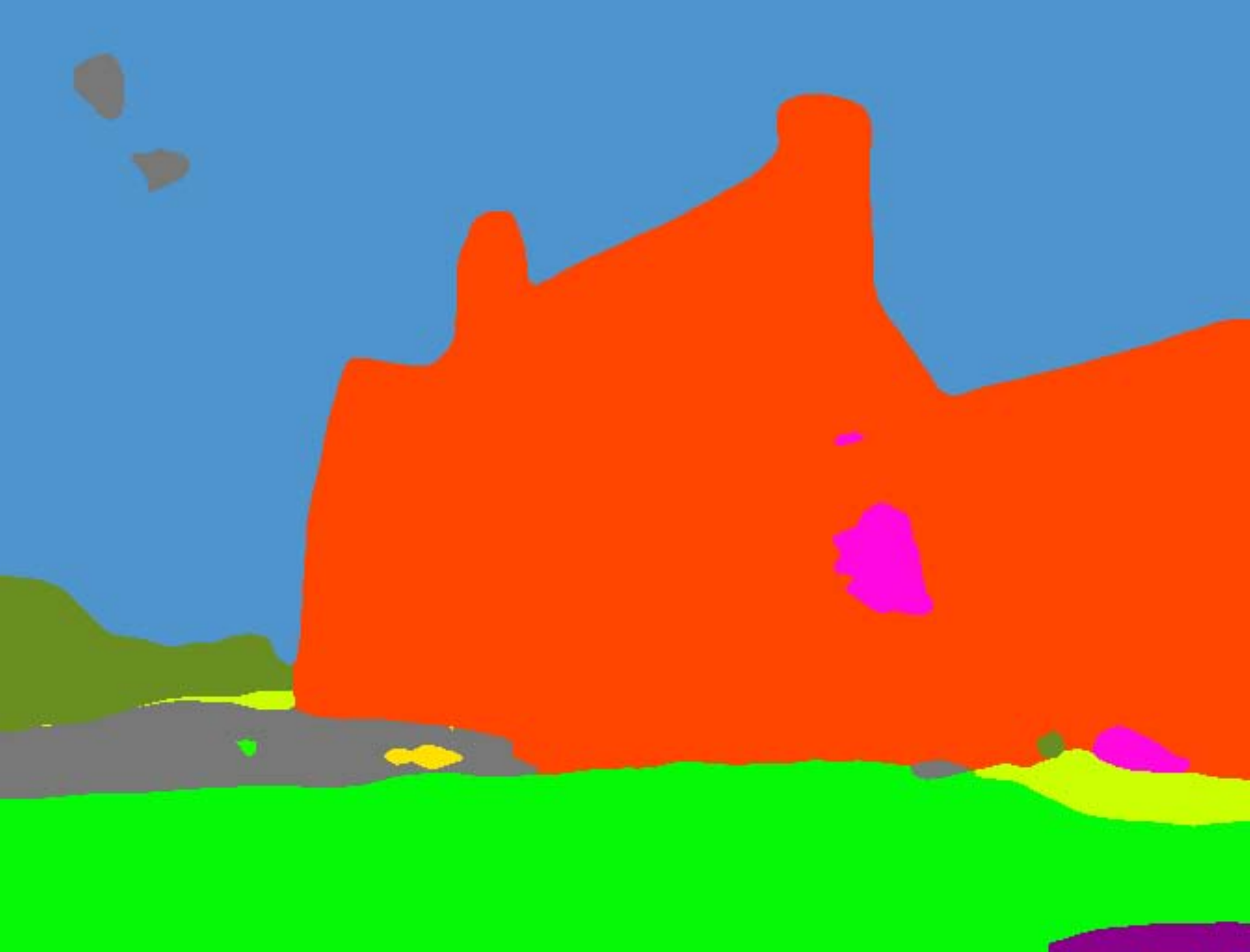}
		&\includegraphics[width=0.16\textwidth, height=0.075\textheight]{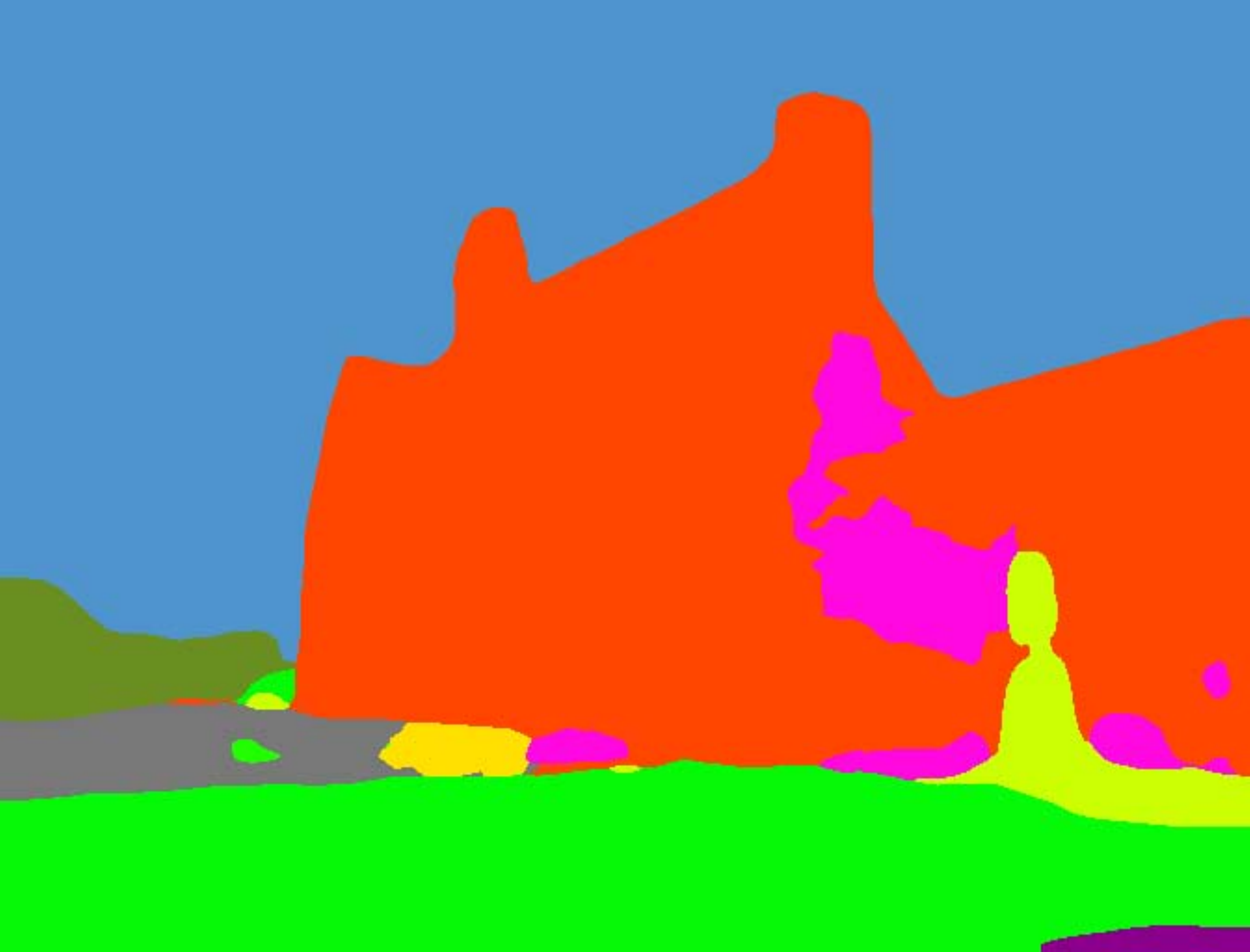}
		&\includegraphics[width=0.16\textwidth, height=0.075\textheight]{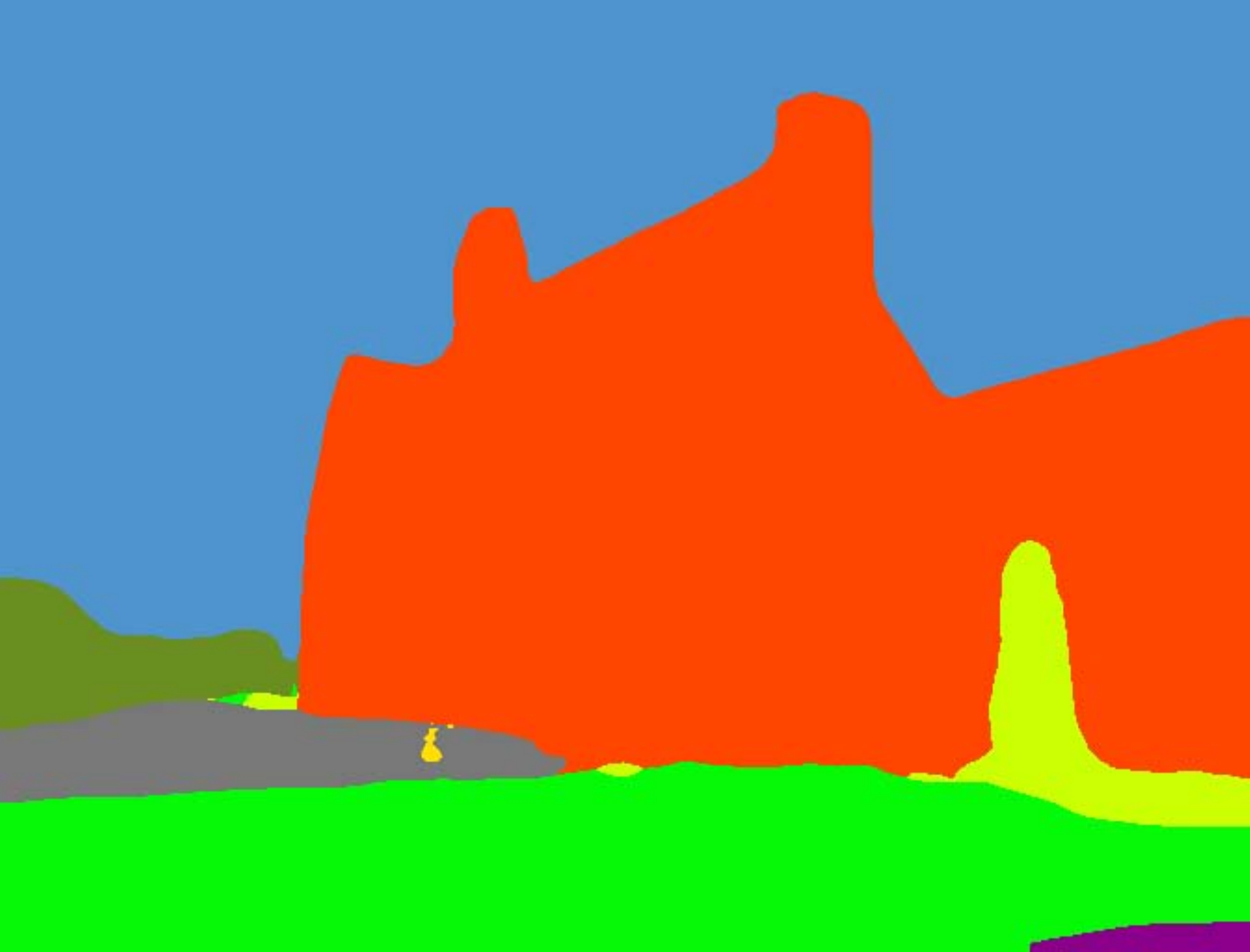}\\	
		SIRR&Syn2Real&MSPFN&DualGCN&MPRNet&Ours\\	
	\end{tabular}
	\caption{Examples of semantic segmentation results after the different deraining algorithms. It can be seen that the result preprocessed by our methods is the closest to the ground truth, proceeding the others in both correctness and accuracy.  }
	\label{fig:Derain_seg}
\end{figure*}
\begin{table*}[h]
	\centering
	\caption{Quantitative results of joint deraining with object detection and semantic segmentation on the VOC2007 and ADE20K datasets.}
	\begin{tabular}{>{\centering}p{1.3cm}|>{\centering}p{1cm}|>{\centering}p{1cm}|>{\centering}p{1cm}|>{\centering}p{1cm}|>{\centering}p{1cm}|>{\centering}p{1cm}|>{\centering}p{1cm}|>{\centering}p{1cm}|>{\centering}p{1cm}|>{\centering}p{1cm}|>{\centering}p{1cm}  }
		\hline
		Methods&\tabincell{c}{DSC\tabularnewline\cite{luo2015removing}} &\tabincell{c}{LP\tabularnewline\cite{li2016rain}}&\tabincell{c}{JORDER\tabularnewline\cite{yang2016joint}}&\tabincell{c}{DDN\tabularnewline\cite{fu2017removing}}&\tabincell{c}{DualRes\tabularnewline\cite{liu2019dual}}&\tabincell{c}{SIRR\tabularnewline\cite{wei2019semi}} &\tabincell{c}{Syn2Real\tabularnewline\cite{yasarla2020syn2real}}&\tabincell{c}{MSPFN\tabularnewline\cite{jiang2020multi}}&\tabincell{c}{DualGCN\tabularnewline\cite{fu2021rain}}&\tabincell{c}{MPRNet\tabularnewline\cite{zamir2021multi}}&Ours\tabularnewline \hline
		\multicolumn{12}{c}{Object detection: VOC2007}
		\tabularnewline \hline
		
		mAP	$\left( \% \right)$ &87.46 &89.04 &80.52 &87.03 &85.45 &90.77 &83.02 &79.37 &96.56 &87.22 &\textbf{97.26} \tabularnewline
		
		mIoU	$\left( \% \right)$ &76.97 &77.36 &81.61 &78.04 &82.64 &81.21 &71.30 &67.72 &82.80 &67.67 &\textbf{83.73} \tabularnewline \hline
		\multicolumn{12}{c}{Semantic segmentation: ADE20K}
		\tabularnewline\hline
		
		mPA	$\left( \% \right)$ &49.99  &40.72&65.92&61.45 &71.69 &57.36 &45.36 &63.50 &44.92 &63.31 &\textbf{74.12} \tabularnewline
		
		mIoU	$\left( \% \right)$& 53.33 & 42.29&70.28 &65.61 &72.92 &61.11 &50.80 &67.78 &46.78 &67.56 &\textbf{79.01} \tabularnewline \hline
		
	\end{tabular}
	\label{tab:app}
\end{table*}
\subsubsection{Effectiveness of Cross-level Recruitment Module}
Considering the effect of the proposed Dynamic Cross-level Recruitment~(DCR), we evaluate the performance with/without the DCR module. Fig.~\ref{fig:Derain_darts} illustrates the visual results. As expected, the low-level feature information is subtly incorporated into the high-level feature via DCR, advancing the network more sensitively to the degraded image. Therefore, for the challenging images where the rain streak cannot be identified from the chaotic background directly~(e.g., the first sample in Fig.~\ref{fig:Derain_darts}), the proposed DCR promotes the understanding of the captured scene and achieves a better performance on rain removal without the loss of detail texture.
\subsubsection{Performance Contributions of the Different Components}
We present ablation experiments on three individual components~(i.e., MFE, CP and DCR) to discuss their contributions. The results on the real-world datasets are shown in Table.~\ref{tab:ablation_components}. 
We treat Unet~\cite{ronneberger2015u} as our baseline and retrain it with rainy image pairs.
As expected, MFE yields a better result by aggregating the multi-scale feature information, and DCR takes advantage of the neural architecture search to achieve automatic cross-level information compensation and enhances the performance further. Additionally, CP implements the opposite constraint to ensure robust performance in the real world. Overall, all three components play a positive role in the entire RDMC method.

\subsection{Limitations}
	As our deraining network is developed for diverse rainy streaks, the recursive multi-scale network might be less effective when the rainy image contains a muddy and rain accumulation phenomenon. Fig.~\ref{fig:fail} shows two examples with ambiguous and accumulated rain. As expected, in the first case, the proposed method barely removes the undesirable rain in the red frame and shows a limited effect in the green frame because the hazy-like rain accumulation obscures the intensity. In the second case, the network cannot capture the imperceptible rain since there are intangible features, especially the region framed in red. A simple solution is to first apply a hazy restoration on the input image to stretch the rain intensity before performing our method. 
	In the future, we will bridge the gap between rain streaks and ambiguous rain, in principle, to fulfill the demands of practical implementation.

\section{Computer Vision Applications}
For outdoor vision systems, the stability and robustness of computer vision algorithms are easily affected by the quality of the images. Therefore, image deraining is an important preprocessing step for subsequent applications. To investigate the effect of the removal performance, we incorporate a series of single image deraining algorithms with object detection and semantic segmentation. To this end, we adopt SSD~\cite{liu2016ssd} and SUST~\cite{zhou2018semantic} for detection and segmentation respectively, and employ two datasets VOC2007~\cite{pascal-voc-2007} and ADE20K~\cite{zhou2017scene} to synthesize the corresponding rainy datasets. These synthetic rainy images contain diverse rain characteristics in direction, orientation, intensity, and brightness. We directly implement the proposed deraining method and the existing state-of-the-art deraining methods to restore the two datasets and then apply the preprocessed rain-free images to the object detection and semantic segmentation networks with their pretrained models. Fig.~\ref{fig:Derain_ssd} and Fig.~\ref{fig:Derain_seg} show the visual results, and the quantitative performance is reported in Table.~\ref{tab:app}.

In Fig.~\ref{fig:Derain_ssd}, it is obvious that the rainy image degrades the recognition of the target, and rain shading also decreases the confidence score. After the deraining procedure, the recognition rate and the detection precision remarkably improve. We can see that the mean Average Precision~(mAP) and Intersection over Union~(IoU) of the restored images by our method achieve the best score in Table.~\ref{tab:app}, preceding the conventional methods by 20\% and outperforming the other deep learning deraining methods.
For the semantic segmentation task shown in Fig.~\ref{fig:Derain_seg}, the segmentation results conducted on the nonrain images we restored are closer to the ground truth, which means that the derained images produced by our method are more applicable for segmentation. 
Our method ranks first in terms of both mean Pixel Accuracy~(mPA) and Intersection over Union~(IoU). It achieves a dramatic improvement where the proposed method is 30\% ahead of the conventional DSC, LP and 6\% higher than the top-performing DualRes in mIoU.

\section{Conclusion}
This paper proposes a contrastive learning based image deraining method. We exploit the interrelationship between the rain and no-rain images and propose a contrastive prior to regularize the reconstructed results. Given the intricate rain distribution and diverse appearance in the real-world scenarios, a recurrent multi-scale framework is established to progressively remove rain interference. In addition, a dynamic cross-layer module based on network structure search is introduced to achieve shallow feature complementarity for the deep layers. Evaluations on both synthetic and real-world images demonstrate that our method performs favorably against the state-of-the-art methods. Moreover, we also demonstrate the effect of the deraining methods on computer vision applications. The experimental results show that the proposed RDMC significantly advances the subsequent applications.

% or


%
%\appendices
%% you can choose not to have a title for an appendix
%% if you want by leaving the argument blank
%\section{}

%\section*{Acknowledgment}
%This work is partially supported by the National Key R\&D Program of China (2020YFB1313503), the National Natural Science Foundation of China under Grant(Nos. 61922019, 61733002, 61672125 and 61772105), LiaoNing Revitalization Talents Program (XLYC1807088), and the Fundamental Research Funds for the Central Universities.

\ifCLASSOPTIONcaptionsoff
  \newpage
\fi



%\bibliographystyle{IEEEtran}
\bibliographystyle{IEEEtran}
\bibliography{bib2020template}

\begin{thebibliography}{10}
\providecommand{\url}[1]{#1}
\csname url@samestyle\endcsname
\providecommand{\newblock}{\relax}
\providecommand{\bibinfo}[2]{#2}
\providecommand{\BIBentrySTDinterwordspacing}{\spaceskip=0pt\relax}
\providecommand{\BIBentryALTinterwordstretchfactor}{4}
\providecommand{\BIBentryALTinterwordspacing}{\spaceskip=\fontdimen2\font plus
\BIBentryALTinterwordstretchfactor\fontdimen3\font minus
  \fontdimen4\font\relax}
\providecommand{\BIBforeignlanguage}[2]{{%
\expandafter\ifx\csname l@#1\endcsname\relax
\typeout{** WARNING: IEEEtran.bst: No hyphenation pattern has been}%
\typeout{** loaded for the language `#1'. Using the pattern for}%
\typeout{** the default language instead.}%
\else
\language=\csname l@#1\endcsname
\fi
#2}}
\providecommand{\BIBdecl}{\relax}
\BIBdecl

\bibitem{liu2016ssd}
W.~Liu, D.~Anguelov, D.~Erhan, C.~Szegedy, S.~Reed, C.-Y. Fu, and A.~C. Berg,
  ``Ssd: Single shot multibox detector,'' in \emph{Proc. Eur. Conf. Comput.
  Vis.}\hskip 1em plus 0.5em minus 0.4em\relax Springer, 2016, pp. 21--37.

\bibitem{liu2022target}
J.~Liu, X.~Fan, Z.~Huang, G.~Wu, R.~Liu, W.~Zhong, and Z.~Luo, ``Target-aware
  dual adversarial learning and a multi-scenario multi-modality benchmark to
  fuse infrared and visible for object detection,'' in \emph{Proc. IEEE Conf.
  Comput. Vis. Pattern Recognit.}, 2022, pp. 5802--5811.

\bibitem{liu2022twin}
R.~Liu, Z.~Jiang, S.~Yang, and X.~Fan, ``Twin adversarial contrastive learning
  for underwater image enhancement and beyond,'' \emph{IEEE Trans. Image
  Process.}, vol.~31, pp. 4922--4936, 2022.

\bibitem{Chen_2019_CVPR}
Z.-M. Chen, X.-S. Wei, P.~Wang, and Y.~Guo, ``Multi-label image recognition
  with graph convolutional networks,'' in \emph{Proc. IEEE Conf. Comput. Vis.
  Pattern Recognit.}, Jun. 2019.

\bibitem{he2016deep}
K.~He, X.~Zhang, S.~Ren, and J.~Sun, ``Deep residual learning for image
  recognition,'' in \emph{Proc. IEEE Conf. Comput. Vis. Pattern Recognit.},
  Jun. 2016, pp. 770--778.

\bibitem{islam2020semantic}
M.~J. Islam, C.~Edge, Y.~Xiao, P.~Luo, M.~Mehtaz, C.~Morse, S.~S. Enan, and
  J.~Sattar, ``Semantic segmentation of underwater imagery: Dataset and
  benchmark,'' in \emph{IEEE/RSJ Int. Conf. on Intell. Robot. Syst.}, Oct.
  2020.

\bibitem{jiang2022target}
Z.~Jiang, Z.~Li, S.~Yang, X.~Fan, and R.~Liu, ``Target oriented perceptual
  adversarial fusion network for underwater image enhancement,'' \emph{IEEE
  Trans. Circuits Syst. Video Technol.}, 2022.

\bibitem{li2016rain}
Y.~Li, R.~T. Tan, X.~Guo, J.~Lu, and M.~S. Brown, ``Rain streak removal using
  layer priors,'' in \emph{Proc. IEEE Conf. Comput. Vis. Pattern Recognit.},
  Jun. 2016, pp. 2736--2744.

\bibitem{liu2020Knowledge}
R.~Liu, Z.~Jiang, X.~Fan, and Z.~Luo, ``Knowledge-driven deep unrolling for
  robust image layer separation,'' \emph{IEEE Trans. Neural Netw. Learn.
  Syst.}, vol.~31, no.~5, pp. 1653--1666, 2020.

\bibitem{risheng2018aggre}
R.~Liu, X.~Fan, M.~Hou, Z.~Jiang, Z.~Luo, and L.~Zhang, ``Learning aggregated
  transmission propagation networks for haze removal and beyond,'' \emph{IEEE
  Trans. Neural Netw. and Lear. Sys.}, 2019.

\bibitem{kang2012automatic}
L.-W. Kang, C.-W. Lin, and Y.-H. Fu, ``Automatic single-image-based rain
  streaks removal via image decomposition,'' \emph{IEEE Trans. Image Process.},
  vol.~21, no.~4, pp. 1742--1755, Apr. 2012.

\bibitem{luo2015removing}
Y.~Luo, Y.~Xu, and H.~Ji, ``Removing rain from a single image via
  discriminative sparse coding,'' in \emph{Proc. IEEE Int. Conf. Comput. Vis.},
  Dec. 2015, pp. 3397--3405.

\bibitem{jiang2020multi}
K.~Jiang, Z.~Wang, P.~Yi, C.~Chen, B.~Huang, Y.~Luo, J.~Ma, and J.~Jiang,
  ``Multi-scale progressive fusion network for single image deraining,'' in
  \emph{Proc. IEEE Conf. Comput. Vis. Pattern Recognit.}, 2020, pp. 8346--8355.

\bibitem{liu2020investigating}
R.~Liu, P.~Mu, J.~Chen, X.~Fan, and Z.~Luo, ``Investigating task-driven latent
  feasibility for nonconvex image modeling,'' \emph{IEEE Trans. Image
  Process.}, vol.~29, pp. 7629--7640, 2020.

\bibitem{zamir2021multi}
S.~W. Zamir, A.~Arora, S.~Khan, M.~Hayat, F.~S. Khan, M.-H. Yang, and L.~Shao,
  ``Multi-stage progressive image restoration,'' in \emph{Proc. IEEE Conf.
  Comput. Vis. Pattern Recognit.}, 2021.

\bibitem{chen2020simple}
T.~Chen, S.~Kornblith, M.~Norouzi, and G.~Hinton, ``A simple framework for
  contrastive learning of visual representations,'' in \emph{Proc. IEEE Int.
  Conf. Mach. Learn.}\hskip 1em plus 0.5em minus 0.4em\relax PMLR, 2020, pp.
  1597--1607.

\bibitem{liu2019auto}
C.~Liu, L.-C. Chen, F.~Schroff, H.~Adam, W.~Hua, A.~L. Yuille, and L.~Fei-Fei,
  ``Auto-deeplab: Hierarchical neural architecture search for semantic image
  segmentation,'' in \emph{Proc. IEEE Conf. Comput. Vis. Pattern Recognit.},
  2019, pp. 82--92.

\bibitem{liu2021learning}
J.~Liu, X.~Fan, J.~Jiang, R.~Liu, and Z.~Luo, ``Learning a deep multi-scale
  feature ensemble and an edge-attention guidance for image fusion,''
  \emph{IEEE Trans. Circuits Syst. Video Technol.}, vol.~32, no.~1, pp.
  105--119, 2021.

\bibitem{liu2021smoa}
J.~Liu, Y.~Wu, Z.~Huang, R.~Liu, and X.~Fan, ``Smoa: Searching a
  modality-oriented architecture for infrared and visible image fusion,''
  \emph{IEEE Signal. Proc. Let.}, vol.~28, pp. 1818--1822, 2021.

\bibitem{zhu2017unpaired}
J.-Y. Zhu, T.~Park, P.~Isola, and A.~A. Efros, ``Unpaired image-to-image
  translation using cycle-consistent adversarial networks,'' in \emph{Proc.
  IEEE Int. Conf. Comput. Vis.}, 2017, pp. 2223--2232.

\bibitem{fu2016weighted}
X.~Fu, D.~Zeng, Y.~Huang, X.-P. Zhang, and X.~Ding, ``A weighted variational
  model for simultaneous reflectance and illumination estimation,'' in
  \emph{Proc. IEEE Conf. Comput. Vis. Pattern Recognit.}, Jun. 2016, pp.
  2782--2790.

\bibitem{yang2016joint}
W.~Yang, R.~T. Tan, J.~Feng, J.~Liu, Z.~Guo, and S.~Yan, ``Deep joint rain
  detection and removal from a single image,'' in \emph{Proc. IEEE Conf.
  Comput. Vis. Pattern Recognit.}, Jul. 2017, pp. 1357--1366.

\bibitem{fu2017removing}
X.~Fu, J.~Huang, Y.~Huang, Delu~Zeng, X.~Ding, and J.~Paisley, ``Removing rain
  from single images via a deep detail network,'' in \emph{Proc. IEEE Conf.
  Comput. Vis. Pattern Recognit.}, Jul. 2017, pp. 1715--1723.

\bibitem{fu2021rain}
X.~Fu, Q.~Qi, Z.-J. Zha, Y.~Zhu, and X.~Ding, ``Rain streak removal via dual
  graph convolutional network,'' in \emph{Proc. AAAI Conf. Artif. Intell},
  2021.

\bibitem{li2017single}
R.~Li, L.-F. Cheong, and R.~T. Tan, ``Single image deraining using scale-aware
  multi-stage recurrent network,'' in \emph{Proc. IEEE Conf. Comput. Vis.
  Pattern Recognit.}, Jun. 2018.

\bibitem{zhang2018density}
H.~Zhang and V.~M. Patel, ``Density-aware single image de-raining using a
  multi-stream dense network,'' in \emph{Proc. IEEE Conf. Comput. Vis. Pattern
  Recognit.}, Jun. 2018, pp. 695--704.

\bibitem{hu2019depth}
X.~Hu, C.-W. Fu, L.~Zhu, and P.-A. Heng, ``Depth-attentional features for
  single-image rain removal,'' in \emph{Proc. IEEE Conf. Comput. Vis. Pattern
  Recognit.}, 2019, pp. 8022--8031.

\bibitem{liu2019dual}
X.~Liu, M.~Suganuma, Z.~Sun, and T.~Okatani, ``Dual residual networks
  leveraging the potential of paired operations for image restoration,'' in
  \emph{Proc. IEEE Conf. Comput. Vis. Pattern Recognit.}, 2019, pp. 7007--7016.

\bibitem{ren2020Progressive}
D.~Ren, W.~Zuo, Q.~Hu, P.~Zhu, and D.~Meng, ``Progressive image deraining
  networks: A better and simpler baseline,'' in \emph{Proc. IEEE Conf. Comput.
  Vis. Pattern Recognit.}, Jun. 2019, pp. 3937--3946.

\bibitem{wang2020spatial}
T.~Wang, X.~Yang, K.~Xu, S.~Chen, Q.~Zhang, and R.~W.~H. Lau, ``Spatial
  attentive single-image deraining with a high quality real rain dataset,'' in
  \emph{Proc. IEEE Conf. Comput. Vis. Pattern Recognit.}, Jun. 2019, pp.
  12\,271--12\,279.

\bibitem{wei2019semi}
W.~Wei, D.~Meng, Q.~Zhao, Z.~Xu, and Y.~Wu, ``Semi-supervised transfer learning
  for image rain removal,'' in \emph{Proc. IEEE Conf. Comput. Vis. Pattern
  Recognit.}, 2019, pp. 3877--3886.

\bibitem{yasarla2020syn2real}
R.~Yasarla, V.~A. Sindagi, and V.~M. Patel, ``Syn2real transfer learning for
  image deraining using gaussian processes,'' in \emph{Proc. IEEE Conf. Comput.
  Vis. Pattern Recognit.}, 2020, pp. 2726--2736.

\bibitem{dai2017contrastive}
B.~Dai and D.~Lin, ``Contrastive learning for image captioning,'' \emph{arXiv
  preprint arXiv:1710.02534}, 2017.

\bibitem{park2020contrastive}
T.~Park, A.~A. Efros, R.~Zhang, and J.-Y. Zhu, ``Contrastive learning for
  unpaired image-to-image translation,'' in \emph{Proc. Eur. Conf. Comput.
  Vis.}\hskip 1em plus 0.5em minus 0.4em\relax Springer, 2020, pp. 319--345.

\bibitem{wu2021contrastive}
H.~Wu, Y.~Qu, S.~Lin, J.~Zhou, R.~Qiao, Z.~Zhang, Y.~Xie, and L.~Ma,
  ``Contrastive learning for compact single image dehazing,'' in \emph{Proc.
  IEEE Conf. Comput. Vis. Pattern Recognit.}, 2021, pp. 10\,551--10\,560.

\bibitem{liu2022learn}
J.~Liu, Y.~Wu, G.~Wu, R.~Liu, and X.~Fan, ``Learn to search a lightweight
  architecture for target-aware infrared and visible image fusion,'' \emph{IEEE
  Signal Process. Lett.}, pp. 1--5, 2022.

\bibitem{zoph2018learning}
B.~Zoph, V.~Vasudevan, J.~Shlens, and Q.~V. Le, ``Learning transferable
  architectures for scalable image recognition,'' in \emph{Proc. IEEE Conf.
  Comput. Vis. Pattern Recognit.}, 2018, pp. 8697--8710.

\bibitem{cai2018path}
H.~Cai, J.~Yang, W.~Zhang, S.~Han, and Y.~Yu, ``Path-level network
  transformation for efficient architecture search,'' in \emph{Proc. IEEE Int.
  Conf. Mach. Learn.}\hskip 1em plus 0.5em minus 0.4em\relax PMLR, 2018, pp.
  678--687.

\bibitem{liu2018darts}
H.~Liu, K.~Simonyan, and Y.~Yang, ``Darts: Differentiable architecture
  search,'' in \emph{Proc. IEEE Int. Conf. Learn. Rep.}, 2019.

\bibitem{zela2018towards}
A.~Zela, A.~Klein, S.~Falkner, and F.~Hutter, ``Towards automated deep
  learning: Efficient joint neural architecture and hyperparameter search,''
  \emph{arXiv preprint arXiv:1807.06906}, 2018.

\bibitem{rawal2018nodes}
A.~Rawal and R.~Miikkulainen, ``From nodes to networks: Evolving recurrent
  neural networks,'' \emph{arXiv preprint arXiv:1803.04439}, 2018.

\bibitem{chrabaszcz2017downsampled}
P.~Chrabaszcz, I.~Loshchilov, and F.~Hutter, ``A downsampled variant of
  imagenet as an alternative to the cifar datasets,'' \emph{arXiv preprint
  arXiv:1707.08819}, 2017.

\bibitem{xie2018snas}
S.~Xie, H.~Zheng, C.~Liu, and L.~Lin, ``Snas: stochastic neural architecture
  search,'' in \emph{Proc. IEEE Conf. Learn. Rep.}, 2018.

\bibitem{liu2022attention}
J.~Liu, J.~Shang, R.~Liu, and X.~Fan, ``Attention-guided global-local
  adversarial learning for detail-preserving multi-exposure image fusion,''
  \emph{IEEE Trans. Circuits Syst. Video Technol.}, 2022.

\bibitem{ronneberger2015u}
O.~Ronneberger, P.~Fischer, and T.~Brox, ``U-net: Convolutional networks for
  biomedical image segmentation,'' in \emph{Int. Conf. on Med. Image Computing
  and comput.-assist. Interv.}\hskip 1em plus 0.5em minus 0.4em\relax Springer,
  2015, pp. 234--241.

\bibitem{Dong2020Multi}
H.~Dong, J.~Pan, L.~Xiang, Z.~Hu, X.~Zhang, F.~Wang, and M.~H. Yang,
  ``Multi-scale boosted dehazing network with dense feature fusion,'' in
  \emph{Proc. IEEE Conf. Comput. Vis. Pattern Recognit.}, 2020.

\bibitem{liu2021bilevel}
R.~Liu, J.~Liu, Z.~Jiang, X.~Fan, and Z.~Luo, ``A bilevel integrated model with
  data-driven layer ensemble for multi-modality image fusion,'' \emph{IEEE
  Trans. Image Process.}, vol.~30, pp. 1261--1274, 2021.

\bibitem{li2020single}
C.~Li, Y.~Yang, K.~He, S.~Lin, and J.~E. Hopcroft, ``Single image reflection
  removal through cascaded refinement,'' in \emph{Proc. IEEE Conf. Comput. Vis.
  Pattern Recognit.}, 2020, pp. 3565--3574.

\bibitem{wei2019single}
K.~Wei, J.~Yang, Y.~Fu, D.~Wipf, and H.~Huang, ``Single image reflection
  removal exploiting misaligned training data and network enhancements,'' in
  \emph{Proc. IEEE Conf. Comput. Vis. Pattern Recognit.}, 2019, pp. 8178--8187.

\bibitem{simonyan2014very}
K.~Simonyan and A.~Zisserman, ``Very deep convolutional networks for
  large-scale image recognition,'' \emph{Comput. Sci.}, 2014.

\bibitem{Kingma2014Adam}
D.~Kingma and J.~Ba, ``Adam: A method for stochastic optimization,''
  \emph{Comput. Sci.}, 2014.

\bibitem{zhang2019image}
H.~Zhang, V.~Sindagi, and V.~M. Patel, ``Image de-raining using a conditional
  generative adversarial network,'' \emph{IEEE Trans. Circuits Syst. Video
  Technol.}, vol.~30, no.~11, pp. 3943--3956, 2019.

\bibitem{zhou2018semantic}
B.~Zhou, H.~Zhao, X.~Puig, T.~Xiao, S.~Fidler, A.~Barriuso, and A.~Torralba,
  ``Semantic understanding of scenes through the ade20k dataset,'' \emph{Int.
  J. Comput. Vis.}, 2018.

\bibitem{pascal-voc-2007}
M.~Everingham, L.~Van~Gool, C.~K.~I. Williams, J.~Winn, and A.~Zisserman, ``The
  pascal visual object classes challenge 2007(voc2007) results,''
  http://www.pascal-network.org/challenges/VOC/voc2007/workshop/index.html.

\bibitem{zhou2017scene}
B.~Zhou, H.~Zhao, X.~Puig, S.~Fidler, A.~Barriuso, and A.~Torralba, ``Scene
  parsing through ade20k dataset,'' in \emph{Proc. IEEE Conf. Comput. Vis.
  Pattern Recognit.}, 2017.

\end{thebibliography}


%\end{IEEEbiography}
%% insert where needed to balance the two columns on the last page with
%%\newpage


%\vfill


\end{document}